\documentclass[conference, twocolumn]{IEEEtran}
\IEEEoverridecommandlockouts

\usepackage{cite}
\usepackage{amsmath,amssymb,amsfonts}
\usepackage{algorithmic}
\usepackage{graphicx}
\usepackage{textcomp}
\usepackage{xcolor}

\usepackage[utf8]{inputenc}
\usepackage[T1]{fontenc}
\usepackage{url}
\usepackage{booktabs}
\usepackage{nicefrac}
\usepackage{multirow}
\usepackage{microtype}
\usepackage{graphbox}
\usepackage{subcaption}
\usepackage{relsize}
\usepackage[percent]{overpic}
\usepackage{mathtools}
\usepackage{centernot}
\newcommand\fy[1][\partial]{\centernot{#1}}
\usepackage{dsfont}
\usepackage{makecell}
\usepackage{lipsum}
\usepackage[most]{tcolorbox}

\usepackage{soul}
\usepackage{enumitem}
\usepackage{afterpage}
\usepackage{placeins}
\newcommand{\approach}{\textsc{illume}}

\def\BibTeX{{\rm B\kern-.05em{\sc i\kern-.025em b}\kern-.08em
    T\kern-.1667em\lower.7ex\hbox{E}\kern-.125emX}}
    
\begin{document}

\title{Explanations Go Linear: Post-hoc Explainability for Tabular Data with Interpretable Meta-Encoding 
\thanks{This work has been partially supported by the European Community Horizon~2020 programme under the funding scheme 
ERC-2018-ADG G.A. 834756 \textit{XAI: Science and technology for the eXplanation of AI decision making}, and the NextGenerationEU programme under the funding schemes: PNRR-PE-AI (M4C2, investment 1.3, line on AI) \textit{FAIR} (Future Artificial Intelligence Research), and ``\textit{SoBigData.it} - Strengthening the Italian RI for Social Mining and Big Data Analytics'' - Prot. IR0000013, and by the Italian Project Fondo Italiano per la Scienza FIS00001966 MIMOSA.
}
}

\author{Simone Piaggesi$^1$, Riccardo Guidotti$^{1,2}$, Fosca Giannotti$^3$, Dino Pedreschi$^1$
\\
$^1$University of Pisa, $^2$ISTI-CNR, $^3$Scuola Normale Superiore - Pisa, Italy
\\
\textit{simone.piaggesi@di.unipi.it}, \{\textit{riccardo.guidotti}, \textit{dino.pedreschi}\}\textit{@unipi.it}, \textit{fosca.giannotti@sns.it} 
}

\maketitle

\begin{abstract}
Post-hoc explainability is essential for understanding black-box machine learning models. Surrogate-based techniques are widely used for local and global model-agnostic explanations but have significant limitations. Local surrogates capture non-linearities but are computationally expensive and sensitive to parameters, while global surrogates are more efficient but struggle with complex local behaviors.
In this paper, we present \approach{}, a flexible and interpretable framework grounded in representation learning, that can be integrated with various surrogate models to provide explanations for any black-box classifier. 
Specifically, our approach combines a globally trained surrogate with instance-specific linear transformations learned 
with a meta-encoder
to generate both local and global explanations.
Through extensive empirical evaluations, we demonstrate the effectiveness of \approach{} in producing feature attributions and decision rules that are not only accurate but also robust and 
computationally efficient, 
thus providing a unified explanation framework that effectively addresses the limitations of traditional surrogate methods.
\end{abstract}

\section{Introduction}

In eXplainable AI (XAI), post-hoc explanations seek to clarify how machine learning (ML) black-box models make decisions. Commonly studied explanations are feature attribution, decision rules, and counterfactuals~\cite{bodria2023benchmarking, guidotti2022counterfactual}. Among post-hoc techniques, surrogate explainers are largely adopted given their effectiveness in performing complex model distillation~\cite{herbinger2023leveraging}. 
Global surrogates, like \textsc{trepan}~\cite{craven1996extracting} 
and related approaches~\cite{frosst2017distilling}, 
involve training a single interpretable model (e.g., linear regression or decision tree) to replicate the behavior of the target black-box
across the entire dataset, providing broad insight but often missing complex non-linear patterns.
Instead, local surrogates, such as \textsc{lime}~\cite{ribeiro2016should} and \textsc{lore}~\cite{guidotti2024stable}, 
build an interpretable model within a (typically synthetically generated) neighborhood of each specific instance, thus better approximating local non-linearities of decision boundaries.
Despite being pivotal 
for explainability~\cite{bodria2023benchmarking, burkart2021survey}, existing local explainers face several limitations~\cite{kindermans2019reliability, zhou2022feature}, including instability~\cite{alvarez2018robustness}, sensitivity to hyperparameters~\cite{bansal2020sam}, computational inefficiencies~\cite{lundberg2020local} and misleading behaviors~\cite{amparore2021trust}.
These drawbacks are primarily attributed to the biases and sampling variability introduced by the neighborhood generation~\cite{dhurandhar2022right, laugel2018defining}. 
In this work, we bridge the gap between local and global surrogate methods by providing a comprehensive post-hoc explanation framework that:
\emph{(i)} overcomes the restrictions of global surrogates regarding their generalization capabilities, \emph{(ii)} addresses the methodological issues of local surrogates going beyond neighborhood sampling, and \emph{(iii)} supports any type of explanation format.

Our proposed framework redefines the post-hoc explanation problem by operating in a feature space distinct from the input. Specifically, it transforms the data into a latent representation~\cite{bengio2013representation} that enhances the expressiveness of global surrogate explanations. 
\begin{figure}[t]
    \centering
    \includegraphics[trim={5mm 15mm 5mm 15mm},clip,width=0.75\linewidth]
    {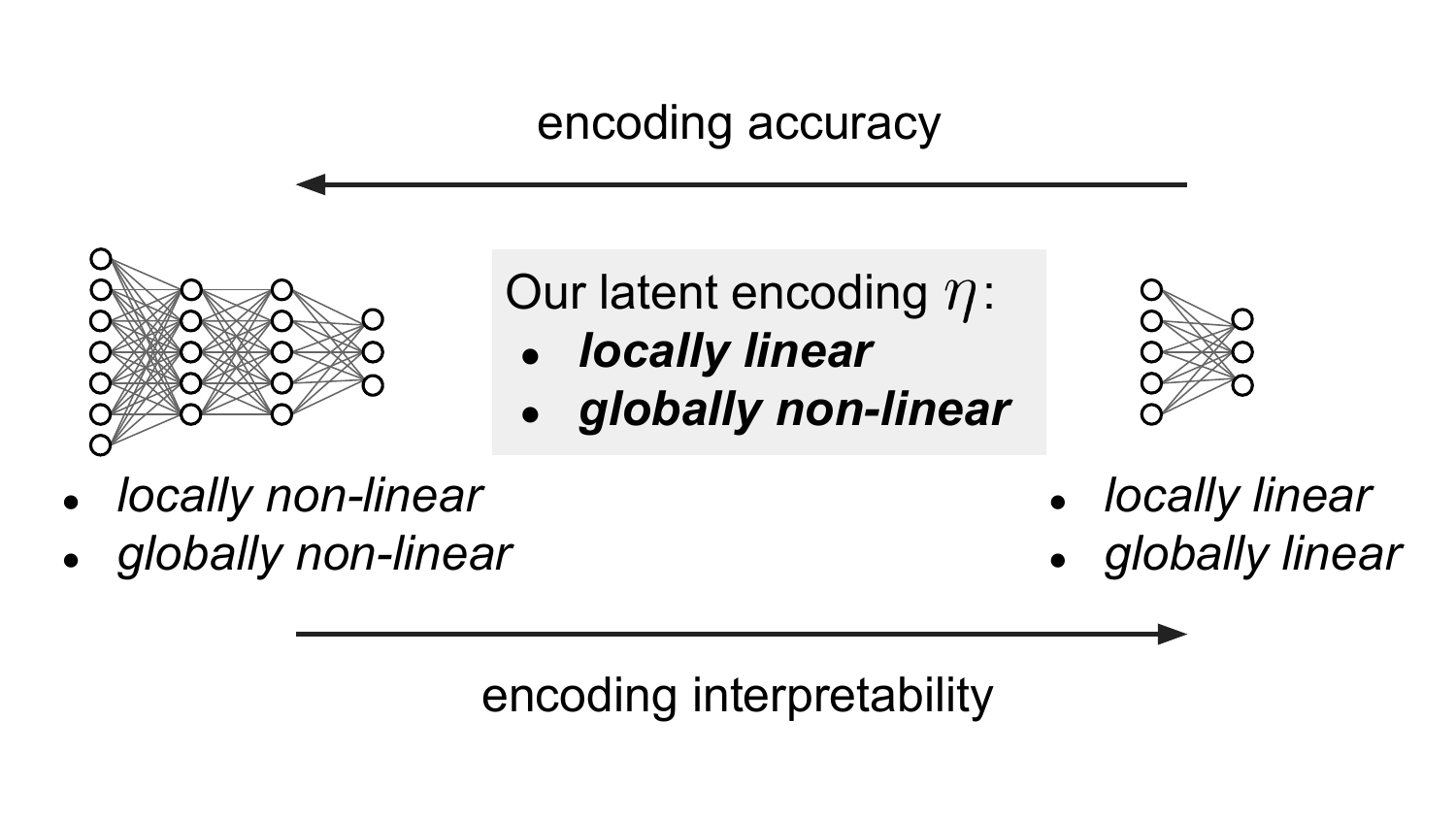}
    \caption{Accuracy-interpretability trade-off for neural encoding.}
    \label{fig:scheme}
\end{figure}
Typically, latent mappings~\cite{sainburg2021parametric} are produced by highly non-linear operations, such as neural networks (NNs), and present 
uninterpretable 
features, rendering any surrogate explanations that rely on these features ineffective. 
Conversely, creating interpretable embeddings is inherently challenging~\cite{koh2020concept} and requires balancing accuracy with transparency~\cite{espinosa2022concept}.
Most straightforward interpretable encodings are achieved through linear transformations, e.g., implemented as single-layer NNs~\cite{piaggesi2024counterfactual, bodria2022transparent} without 
non-linear activations. 
As in Figure~\ref{fig:scheme}, linear NNs (on the right) can be seen as interpretable considering the single-layer structure and the additive nature of output neurons when activation functions are not used.
However, at the same time, they are also too restrictive due to their overly simplified architecture~\cite{takai2021number}.
Specifically, they provide interpretable transformations both globally, since the single-layer network defines a linear function with a unique weights matrix $W$, and locally, since it applies linearly on each specific instance $\mathbf{x}$, resulting in the latent representation $\mathbf{z} = W\mathbf{x}$. 
In contrast, deep NNs achieve greater accuracy but are not interpretable, because applying $\mathit{NN}(\mathbf{x})$ involves obscure complex and stacked (global and local) nonlinear operations.
Neural Additive Models (NAMs)~\cite{agarwal2021neural} are possible candidates for balancing this trade-off. 
NAMs define the mapping as a sum of non-linear functions $\mathit{NN}_j(x_j)$, for each feature $j$, possibly including feature interactions $\mathit{NN}_{jj'}(x_j, x_{j'})$~\cite{lou2013accurate}. 
NAMs are human-readable by looking at visual plots of $NN_j$s and $NN_{jj'}$s~\cite{agarwal2021neural}. 
However, assuming only pairwise interactions, NAMs are restrictive and higher-order extensions lose interpretability and scalability.

As an alternative solution to balance the trade-off, we introduce \textbf{locally linear transformations} as non-linear maps 
$\eta$, such that 
$\eta(\mathbf{x}) = W(\mathbf{x})\mathbf{x}$, where $W(\mathbf{x})$ 
is a neural function returning a specific weight matrix  
once applied to any instance $\mathbf{x}$. 
We see that the output of the function $\eta$ is linear when applied on instance $\mathbf{x}$, without being a linear model as a whole, i.e., globally non-linear. 
Intuitively, $\eta$ acts as a \emph{meta}-encoder, returning for each instance a proper linear encoding used to transform the instance itself.
In combination with a global surrogate model, such meta-encoding approach can be remarkably effective. Specifically, we found that \textit{global explanations deriving from latent space surrogates can be turned into local explanations in the original space if the latent mapping is done with locally linear transformations}.
From a certain point of view, our proposal is inspired by \textit{HyperNetworks}~\cite{ha2017hyper}, a class of NNs used to generate the weights for other NNs.
However, in our case, the generated NN is single-layer perceptron without biases.

To give an intuition, 
we examine the case of additive feature importance methods like \textsc{lime}~\cite{ribeiro2016should} and \textsc{shap}~\cite{lundberg2020local}. 
Let $\mathbf{x} = \{ x_1, \dots, x_m \}$ be one input to a black‐box classifier $b(\cdot)$, where $x_j$ is the value of $j$-th feature in $\mathbf{x}$. 
Local additive explainers fit an instance-specific linear model
$b(\mathbf{x}) \approx  \sum_j \psi_j(\mathbf{x}) x_j$, where 
each learned weight $\psi_j(\mathbf{x})$
is the contribution of feature $j$ to the opaque prediction for instance $\mathbf{x}$. 
On the other hand, analogous global surrogates enable an additive expression of the log-odds as $\log \frac{b(\mathbf{x})}{1-b(\mathbf{x})} \approx \sum_j \beta_j x_j$~\cite{rahnama2024can}. 
Contrarily to local methods,
here the weights $\boldsymbol{\beta} = \{\beta_1, \dots, \beta_m\}$ do not depend on the specific $\mathbf{x}$, failing to describe black-box behavior locally.
In our approach, 
we fit a latent interpretable surrogate, e.g., the linear regressor $\sum_r\beta^L_r z_r$, whose variables are  obtained through locally linear maps, i.e.,
$z_r = \sum_jW_{j,r}(\mathbf{x})x_j$. 
Because $W$ is a linear operation depending on $\mathbf{x}$, the latent global weights can be pulled back to the original features $\log \frac{b(\mathbf{x})}{1-b(\mathbf{x})} \approx \sum_j (W(\mathbf{x})^\top\cdot \boldsymbol{\beta}^L)_jx_j$, yielding local attribution scores valid in the input space. 
Similarly, the illustrated approach can be applied to any other surrogate explanations, 
including factual and counterfactual rules returned by \textsc{lore}~\cite{guidotti2024stable} or decision trees adopted as global surrogates like 
in 
\textsc{trepan}~\cite{craven1996extracting}.

Based on the idea of locally linear transformations, we introduce \approach{}, 
an \textsc{i}nterpretable 
individua\textsc{l}
\textsc{l}atent 
ne\textsc{u}ral
\textsc{m}apping for 
\textsc{e}xplainability.
\approach{} is based on latent space meta-encoding designed to guarantee desirable explanation properties.  
\approach{} trains a regularized meta-encoder that maps inputs into a latent space via locally linear transformations. 
Then, with the resulting embeddings, it fits a post-hoc surrogate model to globally approximate any target black-box system. 
This design enables to generate local explanations from global surrogate logic, 
achieving the precision of  local explainers, while maintaining the efficiency of global surrogates. 
Also, the latent encoding is agnostic with respect to the surrogate model, supporting different types of explanations. 
With a wide range of experiments on tabular data, 
we show that \approach{} is able to generate local explanations as feature importance and decision rules, leading to more accurate, robust, faithful, and efficient explanations than state-of-the-art methods.
The rest of the paper is organized as follows. 
After reviewing related literature in Section~\ref{sec:related}, we describe \approach{} in Section~\ref{sec:method}. 
In Section~\ref{sec:experiments}, we present the experimental results. 
Finally, Section~\ref{sec:conclusion} summarizes our contributions and outlines potential directions for future research.

\begin{figure*}[t]
    \centering
    \includegraphics[trim={20mm 40mm 20mm 45mm},clip,width=\linewidth]
    {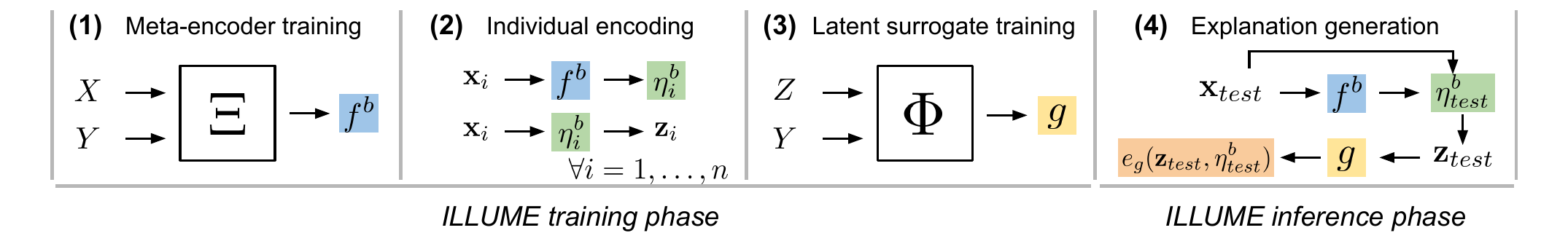}
    \caption{\approach{} steps: 
        \textbf{(1)} the meta-encoder $f^b$ is trained using input instances $X$ and black-box decisions $Y$; 
        \textbf{(2)} $f^b$ generates specific encoding functions $\eta^b_i$ to individually map instances $\mathbf{x}_i$ into latent representations $\mathbf{z}_i$; 
        \textbf{(3)} the set of latent vectors $Z$ is used to train a surrogate model $g$ for imitating the black-box $Y=b(X)$; 
        \textbf{(4)} given a test instance $\mathbf{x}_{test}$, it is mapped into $\mathbf{z}_{test}$ with $\eta^b_{test}$ obtained by $f^b$, then the explanation is obtained with $e_g(\mathbf{z}_{test}, \eta^b_{test})$ by combining the surrogate  logic $g$ with local mapping $\eta^b_{test}$. Training algorithms are marked with white squares. Learned functions are marked with different colored boxes.
        }
    \label{fig:illume}
\end{figure*}

\section{Related Works}
\label{sec:related}

Surrogate explainability methods approximate black-box model predictions using simple interpretable models~\cite{burkart2021survey}. 
Global surrogates aim to capture the overall decision logic of the black-box through model distillation~\cite{craven1996extracting,frosst2017distilling}.
On the other hand, local surrogates focus on specific model decisions.  
\textsc{lime}~\cite{ribeiro2016should} 
computes feature importance using linear models on locally sampled neighborhoods, while \textsc{lore}~\cite{guidotti2024stable} extracts logic rules with locally trained decision trees, enhancing neighborhood generation with genetic algorithms. 
Other works connecting game theory with local explanations, such as \textsc{shap}~\cite{lundberg2017unified}, enable model-agnostic estimation of Shapley values with local surrogate models. 
Global surrogates can be more efficient than local ones but they might miss complex non-linear relationships of the black-box models~\cite{craven1996extracting, frosst2017distilling}. 
In contrast, local surrogates better capture non-linear decision boundaries but can be unstable~\cite{alvarez2018robustness}, sensitive to hyperparameters~\cite{bansal2020sam}, and computationally demanding~\cite{lundberg2020local}.

Various XAI methods train surrogate models by leveraging feature projection techniques~\cite{wang2021understanding}. 
These approaches are typically used to visualize high-dimensional data by optimizing low-dimensional representations.
Basic methods such as 
\textsc{pca} and \textsc{mds}~\cite{cunningham2015linear}
were designed to preserve the overall structure of the data. 
Later, more sophisticated techniques such as 
\textsc{isomap} and \textsc{lle}~\cite{izenman2012introduction} 
were developed to capture local relationships often overlooked by global techniques.
Recently, algorithms like 
\textsc{t-sne} and \textsc{umap}~\cite{wang2021understanding} 
have gained widespread popularity across scientific disciplines due to their ability to efficiently maintain local complexities and non-linear patterns. 
However, embeddings produced by these techniques are opaque, preventing their direct usage in XAI, as they rely on complex, non-linear transformations. 
Consequently, there is growing research interest in  understanding latent representations~\cite{senel2018semantic}, or learning inherently interpretable ones~\cite{bodria2022transparent, piaggesi2024counterfactual}.

Representation learning involves techniques for automatically deriving feature encoding functions from data~\cite{bengio2013representation}. 
It has become a fundamental component of NN-based models, enabling a wide range of tasks including generative methods~\cite{harshvardhan2020comprehensive}, classification~\cite{le2018supervised}, and regression~\cite{agarwal2021neural}. 
In XAI, latent space methods usually employ deep architectures like auto-encoders, to generate explanations
such as exemplars~\cite{crabbe2021explaining} and counterfactuals~\cite{crupi2024counterfactual}, without looking at the transparency of the entire process. 
Alternatively, interpretable latent spaces have been proposed~\cite{piaggesi2024counterfactual, bodria2022transparent} to facilitate counterfactual search, though their expressive power is limited by global linear mappings. 
Moreover, approaches that jointly train neural encoders with explainers have been developed to create self-interpretable models~\cite{ji2025self}, offering an alternative to post-hoc, model-agnostic explainability methods. While these self-explaining models~\cite{ kadra2024interpretable} provide robust explanatory capabilities, they are not designed for post-hoc explainability, which limits their usability.

\section{Methodology}
\label{sec:method}

We define here \approach{}, a procedure to train an interpretable latent space model that enable learning global post-hoc surrogates to approximate black-box systems in a transparent way.
We first outline the problem formulation and the design principles. 
Then we provide the details of the meta-encoding model and the required regularization terms.
Finally, we describe the procedure that leverages the interpretable encoding to realize the explanation generator.
In Table~\ref{tab:notation} we summarize the notation adopted in the rest of the paper.

\begin{table}[t]
\caption{Common symbols and functions.}
\label{tab:notation}
\setlength{\tabcolsep}{1mm}
        \scriptsize
        \centering
        \begin{tabular}{cl}
        \midrule
        \textbf{Symbol} & \textbf{Description} \\
        \midrule
        \makecell[c]{$\mathcal{X}, \mathcal{Z}, \mathcal{W}, \mathcal{Y}, \mathcal{E}$\\$X, Z, W, Y, E$} & \makecell[l]{continuous spaces and discrete subsets, respectively for:\\ input instances, latent embeddings, input-to-latent \\transformations, black-box decisions and explanations.} \\ 
        \midrule
        \makecell[c]{$\mathbf{x}_{i}$, $\mathbf{x}_{:,j}$, $x_{i,j}$\\ $\mathbf{z}_{i}$, $\mathbf{z}_{:,j}$, $z_{i,j}$} & \makecell[l]{original and latent vectors of $i$-th instance,\\ 
         vectors with $j$-th feature for all instances,\\ original and latent $j$-th feature of $i$-th instance.}\\
         \midrule
         $W_{i}$, $W_{i,:,r}$, $W_{i,j,r}$ & \makecell[l]{$i$-th transformation matrix,\\ 
         $r$-th column of $i$-th  matrix,\\  $(j,r)$-th entry of $i$-th  matrix.}\\
         \midrule
         $b:\mathcal{X}\rightarrow \mathcal{Y}$ & black-box classifier $b(\mathbf{x}_i)$\\
         \midrule
         $g:\mathcal{Z}\rightarrow \mathcal{Y}$ & latent surrogate classifier $g(\mathbf{z}_i)$\\
         \midrule
         $e^b:\mathcal{X}\rightarrow \mathcal{E}$ & black-box local explainer $e^b(\mathbf{x}_i)$\\
         \midrule
         $f^b:\mathcal{X}\rightarrow \mathcal{W}$& meta-encoder function $f^b(\mathbf{x}_i) = W^b_i$\\
         \midrule
         $\eta_i^b:\mathcal{X}\rightarrow\mathcal{Z}$ & \makecell[l]{embedding function $\mathbf{z}_i = \eta_i^b(\mathbf{x}_i)= W^b_i \mathbf{x}_i$ \\ (linear application of $f^b$ output on instance $\mathbf{x}_i$)}\\
         \midrule
          $e_g:\mathcal{Z}\times\mathcal{W}\rightarrow \mathcal{E}$& explanation generator function $e_g(\mathbf{z}_i, \eta^b_i)$\\
         \midrule
         $\Xi:(\mathcal{X}{\times} \mathcal{Y}) {\rightarrow}(\mathcal{X} {\rightarrow} \mathcal{W})$ &  \makecell[l]{meta-encoder training function on $(\mathbf{x}_i, b(\mathbf{x}_i)) {\in} \mathcal{X} {\times} \mathcal{Y}$}\\
         \midrule
         $\Phi:(\mathcal{Z} {\times} \mathcal{Y}) {\rightarrow}(\mathcal{Z} {\rightarrow} \mathcal{Y})$ & \makecell[l]{surrogate training function on $(\mathbf{z}_i, b(\mathbf{x}_i)) {\in} \mathcal{Z} {\times} \mathcal{Y}$}\\
         \midrule
         \end{tabular}
\end{table}

\subsection{Problem Setting and Proposed Approach}
\label{sec:proposed}
Given an input space $\mathcal{X} \subset \mathbb{R}^m$ where $m$ is the number of features, let $X = \{ \mathbf{x}_1, \dots,  \mathbf{x}_n\}$ denote a dataset of $n$ instances in $\mathcal{X}$.
Each instance $\mathbf{x}_i = \{ x_{i,1}, \dots, x_{i,m} \}$ consists of $m$ feature values, where $x_{i,j}$ represents the value of the $j$-th feature in $\mathbf{x}_i$. 
We define a black-box $b$ predictor trained on $X$, $b: \mathcal{X} \rightarrow \mathcal{Y}$,  where $\mathcal{Y}$ is the codomain of the black-box.
We focus here on classification tasks, where $\mathcal{Y} \subset [0,1]^c$ and $c$ denotes the number of classes. 
Typically, the black-box outputs probability estimates for each class, i.e., $c=2$ corresponds to binary classification problems, while $c>2$ applies to multi-class problems\footnote{Although here we focus on binary classification, our approach can be easily extended to multi-class problems using one-vs-rest classifiers.}.  
Local explanations of black-box $b$ are the output of functions $e^b: \mathcal{X} \rightarrow \mathcal{E}$, where $\mathcal{E}$ is the space of explanations. 
Given an instance $\mathbf{x}_i$, a local explainer optimizes a function $e_i^b$ that returns $e_i^b(\mathbf{x}_i)$ as explanation, to highlight the factors that activate black-box decision $b(\mathbf{x}_i)$.
Given the pair $(X, Y)$, where $Y = \{ b(\mathbf{x}_1), \dots,$ $b(\mathbf{x}_n)\}$ denotes the predictions of the black-box $b$ on $X$,
the objective of \approach{} is to 
fit a single \textit{explanation generator} $e_g$ that, given any instance $\mathbf{x}_i$, is able to explain the decision $b(\mathbf{x}_i)$.
Thus, instead of returning an independent explainer $e_i^b$ for each instance like for local surrogates, \approach{} produces a single ML function $e_g$ that individually adapts to the instances analyzed. Such a post-hoc explanation generator offers superior advantages over inherently local explainers. As an inductive function, the generator has more generalization capabilities over local surrogates, which require training an independent model for each instance. 
By retraining for every single explanation, local surrogates are impractical for extensive interpretation tasks.

At the core of \approach{} is the \textit{meta-encoder}. 
Departing from traditional latent space approaches, such as autoencoders or conditioned autoencoders \cite{harvey2022conditional, le2018supervised}, which directly learn functions $\mathcal{X} \rightarrow \mathcal{Z}$ or $(\mathcal{X} \times \mathcal{Y}) \rightarrow \mathcal{Z}$, we focus on optimizing a more sophisticated class of  functions $f : \mathcal{X} \rightarrow (\mathcal{X} \rightarrow \mathcal{Z})$.
While mimicking classical encoders in the mapping result, proposed meta-encoder 
optimizes an intermediate space $\mathcal{W}$, i.e., the continuous space of linear applications from $\mathcal{X}$ to $\mathcal{Z}$, returning an \textit{individual, locally-linear transformation} for each instance.
Individual and linear maps enhance interpretability through linear combinations of input variables and, being instance-specific, offer more expressive power than usual linear maps.
\approach{}'s steps illustrated in Figure~\ref{fig:illume} are as follows:
\begin{description}
    
    \item[Step \textbf{(1)}] The meta-encoder training function
    $\Xi: (\mathcal{X} \times \mathcal{Y}) \rightarrow (\mathcal{X} {\rightarrow} \mathcal{W})$
    is trained to return the meta-encoder
    $f: \mathcal{X} \rightarrow \mathcal{W}$
    over the conditioned space $\mathcal{X} \times \mathcal{Y}$.
    Being conditioned to $b$, we denote the learned function as $f^b$, which returns locally-linear transformations $\eta_i^b$ for any given instance $\mathbf{x}_i$. 
    Each $\eta_i^b: \mathcal{X} \rightarrow \mathcal{Z}$ maps input records into a $k$-D latent space, while preserving feature and decision proximities.
    
    \item[Step \textbf{(2)}]
    Then, for each instance $\mathbf{x}_i \in X$, the trained meta-encoder derives its local-linear transformations $\eta_i^b = f^b(\mathbf{x}_i)$ that consists in a matrix $W_i^b \in \mathbb{R}^{m \times k}$.
    Such transformations are applied to each instance $\mathbf{x}_i \in X$, obtaining the latent embeddings $\mathbf{z}_i = \eta_i^b(\mathbf{x}_i) = W_i^b\mathbf{x}_i$.

    \item[Step \textbf{(3)}] Given the dataset of latent instances $Z=\{\mathbf{z}_1,\dots, \mathbf{z}_n\}$ and the observed black-box decisions $Y$, a second optimization $\Phi: (\mathcal{Z} \times \mathcal{Y}) \rightarrow (\mathcal{Z} \rightarrow \mathcal{Y})$ is performed to train the \emph{surrogate model} $g:\mathcal{Z} \rightarrow \mathcal{Y}$ that globally approximate the black-box in the latent space $\mathcal{Z}$.

    \item[Step \textbf{(4)}] At inference time, given any unseen instance $\mathbf{x}_{\mathit{test}}$, the locally-linear transformation and the latent representation are obtained as $W_{\mathit{test}}^b = f^b(\mathbf{x}_{\mathit{test}})$ and $\mathbf{z}_{\mathit{test}} = W_{\mathit{test}}^b \mathbf{x}_{\mathit{test}} = \eta_{\mathit{test}}^b (\mathbf{x}_{\mathit{test}})$.
    The decision of the surrogate is obtained as $y_{\mathit{test}} = g(\mathbf{z}_{\mathit{test}})$, while the explanation is extracted from the \textit{explanation generator} $e_g(\mathbf{z}_{\mathit{test}}, \eta_{\mathit{test}}^b)$ depending on the type of the surrogate. 
\end{description}
\approach{} constructs explanations according to the chosen surrogate model, e.g., providing feature importance with linear models, or factual/counterfactual rules with decision trees.
The encoder layer is trained separately from the surrogate, which allows capturing the black-box behavior in a general-purpose meta-model.
This modular design enables seamless integration with any surrogate predictor without redesigning the architecture.
\approach{} is designed for input data with interpretable features, such as semantically labeled or concept-based attributes typical of tabular datasets. 
While this may seem restrictive, 
interpretable feature representations are the backbone of many explainers that target black-box predictive systems~\cite{sokol2024interpretable}. 
Hence, we focus on explanations for tabular data, which are easier to analyze and understand without conversions.

\subsection{Principled Design of \approach{}}
\label{sec:design}
As 
depicted 
in Figure~\ref{fig:illume}, \textbf{(1)} consists of training 
the meta-encoder
$f^b: \mathcal{X} \rightarrow (\mathcal{X} \rightarrow \mathcal{Z})$,  
capable of returning 
encoding 
functions $\eta^b_i : \mathcal{X} \rightarrow \mathcal{Z}$ for any input $\mathbf{x}_i$. In \textbf{(2)}, individual encodings map input instances into latent representations, that are used in \textbf{(3)} as predictor variables for surrogate model fitting. We outline here the key properties that encoding functions, obtained in \textbf{(1)},
should satisfy to ensure that 
any interpretable surrogate used in \textbf{(3)} can produce meaningful 
explanations.

\noindent\textbf{(P1) Decision Conditioning.}
The encoding captures the relationships among the input features $\mathbf{x}_i$ and the black-box outcomes for each instance $b(\mathbf{x}_i)$, thereby aligning the learned representations with the black-box decision boundary ~\cite{piaggesi2024counterfactual,bodria2022transparent}. 
In other words, the encoding $\eta$ also depends on the local black-box prediction $b$, such that $\mathbf{z}_i = \eta^b(\mathbf{x}_i)$.  
This design choice ensures that the surrogate model trained on the interpretable latent encoding $\mathcal{Z}$ will capture the behavior of the black-box model it aims to explain.

\noindent\textbf{(P2) Local Linearity.}
The encoding maps linearly the input,
by using matrix transformation $W^b \in \mathbb{R}^{m \times k}$, 
such that relationship between the original 
and the latent features is human-interpretable~\cite{yang2023local,piaggesi2024counterfactual}. 
Moreover, instead of using a 
global 
transformation 
$W^b$ 
for all instances, 
we allow each instance to have its own \emph{individual} linear map.
Indeed, we aim to derive a set of 
matrices $W =\{W^b_1, \dots,  W^b_n\}$, each of which linearly maps an instance to its corresponding latent representation, i.e., $\mathbf{z}_i = \eta_i^b(\mathbf{x}_i) = W^b_i \mathbf{x}_i$. 
Specifically, $z_{i,r} = \sum_{j=1}^m W^b_{i,j,r} x_{i,j}$ where $W_{i,j,r}$ is the value that models the linear relationship between the $j$-th input feature and the $r$-th latent feature for the $i$-th instance.
In this way $\eta_i^b$ retains the flexibility of a deep architecture, avoiding the loss of expressiveness that arises when using a single 
globally
linear transformation~\cite{piaggesi2024counterfactual,bodria2022transparent}. 

\noindent\textbf{(P3) Explanation Consistency.} 
Individual transformations inform how to map specific instances locally, influencing surrogate predictions and thus the explanation generator.  
To ensure \emph{consistent} explanations, we must guarantee that similar instances $\mathbf{x}_1$ and $\mathbf{x}_2$ receive similar encodings, i.e., $W_1$ and $W_2$ should be close.
Specifically, if $\eta_i^b(\mathbf{x}_i) = W^b_i\mathbf{x}_i$ represents the mapping for $\mathbf{x}_i$, then the same transformation should hold for a sufficiently small perturbation $\mathbf{x}_j=\mathbf{x}_i+\delta$ applied to the input~\cite{alvarez2018towards}, 
such that $\eta_j^b(\mathbf{x}_j)\approx \eta_i^b(\mathbf{x}_i + \delta) = W^b_i\mathbf{x}_i + W^b_i\delta$.

\smallskip
Building on the above principles, 
\approach{} is inspired by dimensionality reduction~\cite{wang2021understanding}, learning an encoding to the space $\mathcal{Z}$. 
First, \approach{} incorporates both feature-wise and prediction-wise similarity into the latent space model \textbf{(P1)}. 
Inspired by conditioned training in autoencoders and other recent conditioned approaches in the XAI literature~\cite{guyomard2022vcnet,piaggesi2024counterfactual,bodria2022transparent}, 
through $\Xi$, we optimize $f^b$ on 
an augmented
space $\mathcal{X}^b \subset\mathcal{X} \times \mathcal{Y}$, using the enriched dataset $X^b = \{(\mathbf{x}_i, b(\mathbf{x}_i))~|~\forall i \in [1,n] \}$ with black-box predictions paired to each instance. 
Second, \approach{} permits training linear transformations without constraining the encoding 
architecture to be strictly linear, thereby enhancing its expressive power \textbf{(P2)}.
In particular, \approach{} supports the use of a shared model to compute individual linear transformations while maintaining the ability to generalize beyond the training instances.
We express $W^b_i = f^b(\mathbf{x}_i)$ to emphasize that the $i$-th transformation is an explicit learnable function 
applied to $\mathbf{x}_i$.
Third, transformations $W^b_i$ are enforced to be stable w.r.t. infinitesimal displacements of $\mathbf{x}_i$ \textbf{(P3)}. Intuitively, we demand local Lipschitz continuity~\cite{alvarez2018robustness} to bound $\vert\vert f^b(\mathbf{x}) - W^b_i\vert\vert_\text{F}$ with $\Lambda\vert\vert \mathbf{x} - \mathbf{x}_i\vert\vert$, for some constant $\Lambda \in \mathbb{R}$ and  perturbation $\delta \in \mathbb{R}^m$, such that $\vert\vert \mathbf{x} - \mathbf{x}_i\vert\vert<\vert\vert\delta\vert\vert$.
Hence, given $\mathbf{x}_i$, we require that $f^b$ applied to a perturbed instance $(\mathbf{x}_i + \delta)$ remains 
similar
to the transformation applied to $\mathbf{x}_i$. 
This can obtained by minimizing
$L^{\mathit{st}} =  \frac{1}{n}\sum_i\vert\vert ~J_i - W^b_i~\vert\vert ^2_\text{F}$,
where $J_i \in \mathbb{R}^{m \times k}$ is the Jacobian matrix of the transformation for the data-point $\mathbf{x}_i$, with entries
$J_{i,j,r} = \frac{\partial z_{i,r}}{\partial x_j} = W^b_{i,j,r} +  \sum_{v=1}^m \frac{\partial W^b_{i,v,r}}{\partial x_j}x_{i,v}$.

Since the surrogate aims to replicate black-box decisions $b(\mathbf{x}_i)$, the linear functions $\eta_i^b$ are designed not to directly accept $b(\mathbf{x}_i)$ as input argument. 
This prevents information from leaking from the features to the classification label. 
Instead, the influence of the black-box 
is incorporated indirectly through the loss function, as
detailed in the following.
Conceptually, the training optimization 
$\Xi$
conditioned on $\mathcal{Y}$ 
takes as input the training instances 
along with
black-box decisions, and returns the 
trained 
meta-encoder $f^b$ that does not require $\mathcal{Y}$ as input.
We underline that this is markedly different than using a function $f^b$ on the explicit domain $\mathcal{X} \times \mathcal{Y}$.
Indeed, in our proposal, the dependences from $\mathcal{Y}$'s observation are captured implicitly in the optimized weights of the neural function $f^b$ and not requested at inference time.

\subsection{Training the Interpretable Meta-Encoder}
\label{sec:train}

Considering the step \textbf{(1)} of Figure~\ref{fig:illume}, \approach{} procedure starts by training 
the meta-encoder function
$f^b: \mathcal{X} \rightarrow (\mathcal{X} \rightarrow \mathcal{Z})$. 
For any input $\mathbf{x}_i$, 
the learned model $f^b$ returns a 
mapping $\eta^b_i : \mathcal{X} \rightarrow \mathcal{Z}$ in the form of a matrix $W^{b}_i \in \mathbb{R}^{m \times k}$, i.e., $W^{b}_i = f^b(\mathbf{x}_i)$. The scope of this matrix is to linearly embed the original instance $\mathbf{x}_i$ into $\mathbf{z}_i = \eta_i^b(\mathbf{x}_i) = W^b_i \mathbf{x}_i$.
In this section, we describe the training details of \approach{} to obtain the model $f^b$, capable of generating linear transformations and satisfying the design principles discussed.

\smallskip
\noindent\textbf{Learning Objective.}
To learn the meta-encoder $f^b$, we enforce that the input pairwise distance distributions, $P_\mathcal{X}$ and $P_\mathcal{Y}$, mirror the distribution $P_\mathcal{Z}$ over the corresponding latent representations. This requirement forces $\mathcal{Z}$ to capture the structure of data and black-box decisions in $\mathcal{X}^b$, thus yielding feature embeddings that 
faithfully incorporate the black-box behavior.
Moreover, the introduced latent transformations, $\{W^b_i\}_{i=1\dots n} \subset \mathcal{W}$, modulating the mapping of inputs into the encoding space $\mathcal{Z}$, are optimized
to prevent their distribution from deviating arbitrarily, by requiring $P_\mathcal{W}$ to stay close to $P_\mathcal{Z}$. 
This ensures that also $P_\mathcal{W}$ as well reflects the black-box behavior and feature distribution. The learning objective for the model returning $f^b$, then, consists of minimizing the following superposition of Kullback-Leibler divergences:
\begin{equation*}
    L^{\mathit{kl}} {=} \frac{1}{n}\sum_i  \underbrace{\mathit{KL}_i(P_\mathcal{X}||P_\mathcal{Z}){+}\mathit{KL}_i(P_\mathcal{Z}||P_\mathcal{W})}_{\text{aligns~} P_\mathcal{Z} \text{~and~} P_\mathcal{W} \text{~with~} P_\mathcal{X}}\!+\!\underbrace{\mathit{KL}_i(P_\mathcal{Y}||P_\mathcal{Z})}_{\text{aligns~} P_\mathcal{Z} \text{~with~} P_\mathcal{Y}}
\end{equation*}

where $KL_i(P_{\Omega}||P_{\Omega'}) = \sum_{j=1}^n S_{i,j}(\Omega) \log \frac{S_{i,j}(\Omega)}{S_{i,j}(\Omega')}$. 
Probability distributions 
are calculated with pairwise similarity $S_{i,j}$ between instances $\mathbf{x}_i, \mathbf{x}_j \in \mathcal{X}$, black-box predictions $b(\mathbf{x}_i), b(\mathbf{x}_j) \in \mathcal{Y}$, latent representations $\mathbf{z}_i, \mathbf{z}_j \in \mathcal{Z}$, or individual mappings $W^b_i, W^b_j \in \mathcal{W}$, as:
$S_{i,j}(\Omega) = e^{-d_\Omega(i,j)^2}/\sum_{v \neq i}e^{-d_\Omega(i,v)^2}$,
where $d_\Omega(\cdot,\cdot)$ denotes a specified distance metric over the space $\Omega$. 
In practice, the similarity $S_{i,j}$ represents the probability of $j$ being a neighbor of $i$ according to a Gaussian distribution centered on $i$.
This objective encourages distributions $P_\mathcal{W}$ and $P_\mathcal{Z}$ to align with $P_{\mathcal{X}^b}$.
Instead of directly aligning spaces $\mathcal{W}$ and $\mathcal{X}^b$, we minimize the divergence between distributions $P_\mathcal{W}$ and $P_\mathcal{Z}$. The concordance with space $\mathcal{Z}$, which is optimized to mirror $\mathcal{X}$ and $\mathcal{Y}$, indirectly aligns $\mathcal{W}$ and $\mathcal{X}^b$, with $Z$ serving as a denoised and compact abstraction of $X^b$. 
Also, we express the loss as $L^{\mathit{kl}} = L_{x}^{\mathit{kl}} + L_{y}^{\mathit{kl}}$, emphasizing the term $L_{y}^{\mathit{kl}} = \frac{1}{n} \sum_i \mathit{KL}_i(P_\mathcal{Y} \| P_\mathcal{Z})$  responsible for conditioning on $\mathcal{Y}$.

\smallskip
\noindent\textbf{Model Regularizations.}
We employ latent space regularization to impose additional constraints, ensuring \textit{sparsity}, \textit{orthogonality}, and \textit{non-collinearity}.
Inspired by previous work on $\alpha$-sparse autoencoders~\cite{makhzani2013k}, we \textit{sparsify} the transformation matrices $W^b_i$ into $\mathit{sp}_{\alpha}(W^b_i)$ by identifying the $\alpha$ largest weights for each column and setting the other to zero. 
This mechanism ensure that latent space mapping maintains a linear relationship with a limited number of input features. 
Moreover, it enables the user to choose the preferred sparsity level $\alpha$.
Also, to minimize redundancy and ensure that diverse input features contribute to distinct latent dimensions, we apply \textit{soft-orthogonality} constraints~\cite{massart2022orthogonal} between column pairs of the transformation matrices, i.e., we impose to minimize the loss
$L^{\mathit{so}} {=} \frac{1}{n} \sum_i
    \vert\vert \mathit{sp}_{\alpha}(W^b_i) \mathit{sp}_{\alpha}(W^b_i)^\top {-} \mathds{1}_k \vert\vert^2_\text{F}$,
with $\mathds{1}_k$ the unitary matrix.
Finally, to ensure the resulting latent space is composed of minimally correlated
variables~\cite{aas2021explaining}, we optimize the correlation matrix of latent data-points $C(Z)$, with entries as the empirical Pearson scores\footnote{Non-linear rank-based correlation measures, such as Spearman or Kendall scores, are harder to optimize requiring differentiable sorting algorithms~\cite{blondel2020fast}.}
$C_{r,s} = \frac{1}{n}\sum_i 
\Big(\frac{z_{i,r} - \mu(\mathbf{z}_{:,r})}{\sigma(\mathbf{z}_{:,r})}\Big)\Big(\frac{z_{i,s} - \mu(\mathbf{z}_{:,s})}{\sigma(\mathbf{z}_{:,s})}\Big)$, 
denoting with $\mathbf{z}_{:,r} = \{ z_{1,r}, \dots, z_{n,r}\}$ the realizations of the $r$-th feature,
and with $\mu(\cdot)$ and $\sigma(\cdot)$ the empirical average and standard deviation functions. 
\textit{Non-collinearity} is reached by imposing that correlation matrix is nearly identical to the unitary matrix:
$L^{\mathit{co}} = \vert\vert C(Z) - \mathds{1}_k \vert\vert^2_\text{F}$.

\smallskip
\noindent\textbf{Optimization.}
The training of $\Xi$ to obtain $f^b$ is done using mini-batch gradient minimization, where the objective function a single batch of training instances is:
\begin{equation*}
    \mathcal{L}(X,Y,\alpha) = L_{x}^{\mathit{kl}} + \lambda^{\mathit{y}}L_{y}^{\mathit{kl}} + \lambda^{\mathit{st}}L^{\mathit{st}} +
    \lambda^{\mathit{so}}  L^{\mathit{so}} + \lambda^{\mathit{co}}L^{\mathit{co}}
\end{equation*}
Similar to $\alpha$-sparse autoencoders ~\cite{makhzani2013k}, we introduce a sparsity scheduling approach over training epochs to avoid ``dead'' hidden units within the deep neural network architecture of $\Xi$.
Namely, we begin by pre-training the linear model without sparsity constraints, then gradually increase sparsity linearly to the desired level, and finally fine-tune the resulting sparse model until convergence.

\subsection{Generating the Explanations}
\label{sec:explain}
In Figure~\ref{fig:illume}, after training the meta-encoder in \textbf{(1)}, individual encodings map input instances into latent representations in \textbf{(2)}, that serve as predictor variables for fitting the surrogate model $g$ in \textbf{(3)}. 
At inference time \textbf{(4)}, for any instance $\mathbf{x}_i$, \approach{} produces an explanation through the generator $e_g: \mathcal{Z} \times \mathcal{W} \rightarrow \mathcal{E}$. 
This function receives both the meta-encoded local transformation $W^b_i = f^b(\mathbf{x}_i)$ and the resulting linear embedding $\mathbf{z}_i = W^b_i\mathbf{x}_i$, to translate the latent decision logic of $g$ into an explanation.
Here, we describe how $e_g$ generates local explanations by exploiting the inner logic of the surrogate and the locally linear structure of the latent space.

\smallskip
\noindent\textbf{Feature Importance Explanations.}
Given an instance $\mathbf{x}_i$, feature importance explainers assign a real-valued vector $\psi_i=\{\psi_{i,1}, \dots, \psi_{i,m}\}$, in which every $\psi_{i,j}$ is the relevance of the $j$-th feature for the prediction $b(\mathbf{x}_i)$. 
For example, the vector 
$\psi^\text{(Alice)}=\{\psi_{\mathit{age}}\!=\!-0.2, \psi_{\mathit{inc}}\!=\!0.8, \psi_{\mathit{edu}}\!=\!0.5\}$ that we can obtain through \approach{}, 
illustrates the feature importance for the decision to reject the loan application for 
$\mathbf{x}^\text{(Alice)} = \{\mathit{age}\!\!=\!\!25, \mathit{income}\!\!=\!\!15k, \mathit{education}\!\!=\!\!high\}$.
Indeed, by learning a logistic classifier over $Z$, in \approach{} we first derive explanations expressed on latent encodings in the form of additive feature attributions $g(\mathbf{z}_i)= \beta_0 + \sum_{r=1}^k \beta_r z_{i,r}$. 
Then, by exploiting the linearity of $\eta^b_i$, through $e_g$, the projected space attributions\footnote{
This follows from applying the encoding $z_{i,r} = \sum_{j=1}^m W^b_{i,j,r} x_{i,j}$ in the expression for $g(\mathbf{z}_i)$.
For simplicity, we neglected the intercept $\beta_0$.} are converted into input space attributions $g(\eta^b_i(\mathbf{x}_i))= \sum_{j=1}^m \psi_{i,j} x_{i,j}$, with feature importance defined as $\psi_{i,j} = \sum_{r=1}^k \beta_r W^b_{i,j,r}$. 
Hence, in \approach{} the local importance of input feature $j$ is determined by summing up the global relevances of the latent features $\{\beta_r\}$ weighted by the magnitudes of the mapping $W^b_{i,j,r}$. 
This weighting reflects how strongly feature $j$ contributes to each latent feature $r$ based on the logistic surrogate coefficients.
Recalling the example above, suppose that instance 
$\mathbf{x}^\text{(Alice)}$
is mapped into a 2-D vector 
$\mathbf{z}^\text{(Alice)}=\{z_1\!\!=\!\!W^\text{(Alice)}_{\mathit{age},1}x_{\mathit{age}}\!+\! W^\text{(Alice)}_{\mathit{inc},1}x_{\mathit{inc}}, z_2\!\!=\!\! W^\text{(Alice)}_{\mathit{age},2}x_{\mathit{age}}\!+\! W^\text{(Alice)}_{\mathit{edu},2}x_{\mathit{edu}}\}$
given by a sparse transformation valid for Alice. 
Hence, after learning the global logistic explanation $\psi^z=\{\beta_1, \beta_2\}$, the local feature importance vector $
\psi^\text{(Alice)}$ is generated as:
$
\psi^\text{(Alice)}\!\!=\!\! 
\{\psi_{\mathit{age}}\!\!=\!\! \beta_1 W^\text{(Alice)}_{\mathit{age},1}\!+\!\beta_2 W^\text{(Alice)}_{\mathit{age},2}, 
\psi_{\mathit{inc}}\!\!=\!\! \beta_1 W^\text{(Alice)}_{\mathit{inc},1}, \psi_{\mathit{edu}}\!\!=\!\! \beta_2 W^\text{(Alice)}_{\mathit{edu},2}\}
$.

\smallskip
\noindent\textbf{Decision Rule Explanations.}
Given a record $\mathbf{x}_i$, a set of decision rules $\rho^x_i$ explains the black-box decision $b(\mathbf{x}_i)$ with the logical premises that lead to the decision~\cite{guidotti2024stable}. 
$\rho^x_i$ is composed by axis-parallel Boolean conditions on feature values in the form $x_{i,j} \in [l^x_{i,j}, u^x_{i,j}]$, where $l^x_{i,j}, u^x_{i,j}$ are lower and upper bound values in the domain of $x_{i,j}$, extended with $\pm \infty$. 
For example, the rule 
$\rho^\text{(Bob)}=\{\mathit{age}\!\leq\!20, \mathit{income}\!\leq\!30k, \mathit{education}\!\leq\!\mathit{bachelor}\}$ that we can obtain through \approach{},
explains the rejection of loan application for $\mathbf{x}^\text{(Bob)} = \{\mathit{age}\!\!=\!\!18, \mathit{income}\!\!=\!\!25k, \mathit{education}\!\!=\!\!low\}$.
In \approach{}, when $g$ is a decision tree, we first derive global decision rules, determined as root-leaf paths in the decision tree trained over feature space $Z$, i.e., $\rho^z_i=\{z_{i,r} \in [l^z_r, u^z_r]\}_{r=1\dots k}$ (lower and upper bound are in the domain of $z_{i,r}$, extended with $\pm \infty$).  
Then, by exploiting the linearity of $\eta^b_i$, these rules are converted into input space local oblique rules 
$\Tilde{\rho}^x_i=\{\sum_{j=1}^m W^b_{i,j,r}x_{i,j} \in [l^z_r, u^z_r]\}_{r=1\dots k}$. 
Also, for more readability, we convert oblique rules into the axis-parallel format: $\rho^x_i=\{x_{i,j} \in [l^x_{i,j}, u^x_{i,j}]\}_{j=1\dots m}$. 
The upper and lower bounds for these rules satisfy the following constraints: 
$ \quad
l^x_{i,j} - x_{i,j} = \underset{r}{\max}\frac{l^z_r - z_{i,r}}{W^b_{i,j,r}}\quad  \text{and}\quad 
u^x_{i,j} - x_{i,j} = \underset{r}{\min}\frac{u^z_r - z_{i,r}}{W^b_{i,j,r}}, 
$
where the max/min operations ensure taking the most restrictive inequality among the $k$ oblique latent conditions. 
Essentially, with \approach{} the global rules with bounds $[l^z_r, u^z_r]$ are locally rescaled using the individual weights $W^b_{i,j,r}$.
This rescaling makes the explanations more 
adaptive, tailoring them to the 
local contributions 
of each input feature.
For instance, w.r.t. the previous example, we fit a surrogate tree $g$ on the 2-D embedding space, 
and obtain global latent rules $\rho^z=\{-\infty \leq z_1 \leq \xi, \lambda \leq z_2 \leq +\infty \}$. 
Due to the mapping linearity, we obtain the local axis-parallel rules valid for Bob as:
$
\rho^\text{(Bob)}\!\!=\!\! \big\{ x_{\mathit{age}} -\frac{\xi - z_1}{W^\text{(Bob)}_{\mathit{age},1}} \!\leq\! x_{\mathit{age}} \!\leq\! x_{\mathit{age}}+\frac{\lambda - z_2}{W^\text{(Bob)}_{\mathit{age},2}}, -\infty\!\leq\! x_{\mathit{inc}}\!\leq\! x_{\mathit{inc}}+ \frac{\xi - z_1}{W^\text{(Bob)}_{\mathit{inc},1}}, x_{\mathit{edu}} - \frac{\lambda - z_2}{W^\text{(Bob)}_{\mathit{edu},2}} \!\leq\! x_{\mathit{edu}} \!\leq\! +\infty \big\}.
$

\smallskip
\noindent\textbf{Perfect Fidelity via Similarity Search.} 
Relying on surrogate logic, in \approach{} the explanations $e_g(\mathbf{z}_i, \eta_i^b)$ are valid iff $g(\eta_i^b(\mathbf{x}_i)) = b(\mathbf{x}_i)$, i.e., when the surrogate 
correctly predicts the black-box.
To ensure producing valid explanations for \emph{every} instance, when $g(\eta_i^b(\mathbf{x}_i)) \neq b(\mathbf{x}_i)$, we perform a vector search to find in $Z$ the nearest 
latent 
instance $\mathbf{z}_j$ to $\mathbf{z}_i$ s.t. 
$g(\eta_j^b(\mathbf{x}_j)) = b(\mathbf{x}_j)$ and $b(\mathbf{x}_i)= b(\mathbf{x}_j)$, using its explanation $e_g(\mathbf{z}_j, \eta_j^b)$ instead. 
This approach leverages \approach{}'s design, which ensures that nearby points in latent space have similar transformations. 
Hence, explanations for these points remain closely aligned and reliable.

\section{Experiments}
\label{sec:experiments}
We run large-scale experiments to answer these questions:

\begin{description}[leftmargin=*]
    \item[RQ1 -] Is the latent space from \approach{} effective in preserving the original structure concerning both features and decisions?   
    \item[RQ2 -] Are black-box explanations generated with \approach{} accurate and aligned with truthful explanations?    
    \item[RQ3 -] Is \approach{} reliable in the absence of truthful explanations?
    \item[RQ4 -] Is \approach{} computationally efficient respect to conventional local explainers?
\end{description}

\subsection{Experimental Setting}
\label{sec:settings}
In this section we detail the experimental setup adopted\footnote{
Experiments were performed with CPU 3.0 GHz × 36 Intel Core i9, 252 GB RAM, and GPU Nvidia RTX 6000 24GB. Source code of \approach{} and Appendix
are available at: \url{https://github.com/simonepiaggesi/illume/}.}.

\begin{table}[t]
    \caption{Real-world data characteristics and black-box performances. In order: number of instances $n$, number of total features $m$, categorical features $h$, number of classes $c$, majority and minority class percentages, macro-F1 classification scores for XGB, LGB and CTB on test sets.}
    \centering
    \setlength{\tabcolsep}{1.mm}
    \footnotesize
    \begin{tabular}{lrrrrrrrrrr}
    \midrule
    \textbf{Dataset} & $\boldsymbol{n}$ & $\boldsymbol{m}$ &
    $\boldsymbol{h}$ & $\boldsymbol{c}$ & \textbf{maj(\%)} & \textbf{min(\%)} & \textbf{XGB} & \textbf{LGB} & \textbf{CTB} \\
    \midrule
    \texttt{aids}       & 2,139  & 36 & 26  & 2  & .756 & .244 & .893 & .895 & .890\\
    \texttt{austr}  & 690   & 46 & 40  & 2  & .555 & .445 & .912 & .905 & .920 \\
    \texttt{bank}        & 4,119  & 63 & 53  & 2  & .891 & .109 & .775 & .786 & .771 \\
    \texttt{breast}      & 569   & 30 & 0   & 2  & .627 & .373 & .972 & .991 & .981 \\
    \texttt{churn}     & 3,333  & 71 & 55  & 2  & .855 & .145 & .946 & .930 & .946 \\
    \texttt{compas}      & 7,214  & 20 & 13  & 2  & .723 & .277 & .714 & .711 & .713 \\
    \texttt{ctg}         & 2,126  & 56 & 33  & 3  & .778 & .083 & .979 & .982 & .979 \\
    \texttt{diabetes}    & 768   & 8 & 0   & 2  & .651 & .349 & .803 & .805 & .824 \\
    \texttt{ecoli}      & 336   & 7 & 0   & 8  & .426 & .006 & .859 & .876 & .892 \\
    \texttt{fico}        & 10,459 & 23 & 0   & 2  & .522 & .478 & .733 & .733 & .737 \\
    \texttt{german}      & 1,000  & 61 & 54  & 2  & .700   & .300   & .706 & .706 & .716\\
    \texttt{home}        & 492   & 7 & 0   & 2  & .545 & .455 & .949 & .959 & .949\\
    \texttt{ionos}  & 351   & 34 & 0   & 2  & .641 & .359 & .938 & .953 & .953\\
    \texttt{sonar}       & 208   & 60 & 0   & 2  & .534 & .466 & .881 & .881 & .905\\
    \texttt{spam}        & 4,601  & 57 & 0   & 2  & .606 & .394 & .959 & .960 & .964\\
    \texttt{titanic}     & 891   & 9 & 5   & 2  & .616 & .384 & .841 & .830 & .830\\
    \texttt{wine}        & 6,497  & 11 & 0   & 7  & .437 & .001 & .426 & .433 & .418\\
    \texttt{yeast}       & 1,484  & 8 & 0   & 10 & .312 & .003 & .592 & .527 & .606\\
    \midrule
    \end{tabular}
    \label{tab:datasets}
\end{table}

\smallskip
\noindent\textbf{Datasets.}
We present our results on a variety of synthetic and real-world datasets used in prior works. 
In line with~\cite{rahnama2023blame,mollas2023truthful}, for synthetic datasets, we utilize the \textsc{seneca} framework for implementing  transparent classifiers, as proposed in~\cite{guidotti2021evaluating}. 
Specifically, we generate synthetic datasets with $t$ informative features and $u$ uninformative features. 
The total number of features, $m=t+u$, is set in the list $\{4,8,16,32,64\}$, and we set $t = \mathrm{min}\{16, t+u\}$. For a fixed $m$, we generate five rule-based classifiers and five linear classifiers. 
Hence, we explain 2,048 instances for each of them.
Furthermore, we employ 18 real-world datasets from UCI ML Repository.
As black-box classifiers, 
we consider the ensemble methods~\cite{ibomoiye2022survey} XGBoost (XGB), LightGBM (LGB), and CatBoost (CTB)
as they are among the most effective techniques for tabular data~\cite{grinsztajn2022tree}. For binary classification datasets, each explanation method is asked to generate explanations for the class 1.
For multi-class datasets, explanations are referred for the majority class.

\smallskip
\noindent\textbf{Model Setup.}
In line with~\cite{guidotti2024stable, foss2019distance}, we divide the input distance into continuous and categorical (one-hot encoded) features distances. 
With $h$ categorical features after one-hot encoding, we use cosine and Hamming metrics to express the input distance\footnote{$d_{\mathit{cos}}(\mathbf{u},\mathbf{v}) = 1-\frac{\mathbf{u}\cdot\mathbf{v}}{\rvert\rvert\mathbf{u}\rvert\rvert ~\rvert\rvert\mathbf{v}\rvert\rvert}$,~~~~ $d_{\mathit{hmm}}(\mathbf{u},\mathbf{v})=\frac{1}{len(\mathbf{u})}\sum_i \mathds{1}[u_i \neq v_i]$} $d_\mathcal{X}(i,j) = \frac{h-m}{m}d_{\mathit{cos}}(\mathbf{x}^{\mathit{con}}_i, \mathbf{x}^{\mathit{con}}_j) + \frac{h}{m}d_{\mathit{hmm}}(\mathbf{x}^{\mathit{cat}}_i, \mathbf{x}^{\mathit{cat}}_j)$. 
For the latent space and black-box distances ($d_\mathcal{Z}$ and $d_\mathcal{Y}$) we employ cosine distance as well.
For latent space transformation matrices, we consider the average per-column cosine distance $d_\mathcal{W}(i,j) = \frac{1}{k} \sum_r d_{cos}(W^b_{i,:,r},W^b_{j,:,r})$.
We consider \approach{}-\textsc{lr} and \approach{}-\textsc{dt}, where as surrogate models $g$ are used \textsc{l}ogistic \textsc{r}egression to generate feature importance, and \textsc{d}ecision \textsc{t}ree to derive rules.
After hyper-parameter tuning on a validation set, we employ the best combination of regularizations, i.e., \textit{s}oft-\textit{o}rthogonality, and non-\textit{co}llinearity, abbreviated with \textit{so}, and \textit{co}. 
For sparsity, we test two opposite situations: $\alpha=m$ (no sparsity), and $\alpha=2$ (max non-trivial sparsity). 
In each case, every single loss term is optimized with $\lambda=1$ or $\lambda=0$. 
Also, in order to evaluate the impact of decision conditioning and consistency principles, we consider the \textsc{unc}onditioned (\approach{}-\textsc{uc}) and the \textsc{uns}table (\approach{}-\textsc{us}) variations, setting to zero respectively $\lambda^y$ and $\lambda^{st}$.
Finally, we evaluate the impact of the local linearity assumption, by training a global linear encoding \textsc{lin} where a single NN layer represents the function $\eta^b$, without non-linear activations.

\smallskip
\noindent\textbf{Competitors.}
We compare \approach{} against feature reduction methods to evaluate the neighborhood structure preservation in the latent space: 
\textsc{pca},
\textsc{isomap},
\textsc{lle}, 
(parametric-)\textsc{umap}~\cite{sainburg2021parametric}. 
In addition, we compare \approach{} with local explainers to evaluate the quality of explanations: \textsc{lime},
\textsc{shap},
\textsc{lore}, 
\textsc{anchor}~\cite{ribeiro2018anchors}. 
We use kernel\textsc{shap} method for synthetic black-boxes and  tree\textsc{shap} for ensemble black-boxes.
Finally, we compare \approach{} with global surrogate classifiers trained on the input space (\textsc{inp-lr} and \textsc{inp-dt}). 
For logistic-based surrogates (\textsc{inp-lr} and \textsc{lin-lr}), as local explanation the -- global -- coefficients $\{\beta_1, \beta_2, \dots\}$ are weighted by the feature values of the specific instance (see model-intrinsic additive scores~\cite{rahnama2024can}), namely 
$\psi^{\textsc{inp}}_{i,j} =  \beta_j x_{i,j}$ and $\psi^{\textsc{lin}}_{i,j} = \sum_{r=1}^k \beta_r W^b_{j,r}x_{i,j}$.
For a fair comparison, we apply the latent search for maximizing surrogate fidelity also for \textsc{lin}-\textsc{lr} and \textsc{lin}-\textsc{dt} using distance $d_{\mathcal{Z}}$. 
Instead, for \textsc{lr} and \textsc{dt} we search the nearest neighbor in the input space according to $d_{\mathcal{X}}$.
For remaining methods, i.e., \textsc{lime}, \textsc{shap}, \textsc{lore} and \textsc{anchor},  we trust each explanation as valid one, since all these methods ensure guarantees for maximal fidelity.
Each black-box, explainer or embedding is trained and evaluated on 80/20\% splits of every dataset.

\begin{table}[t]
    \caption{RQ1 - Latent space quality metrics.}
    \label{tab:space_quality}
    \setlength{\tabcolsep}{0.8mm}
    \centering
    \begin{tabular}{lccccccc}
        &&\textsc{ill-uc} &\textsc{lin-uc} & \textsc{pca} & \textsc{ismp} & \textsc{lle} & \textsc{ump}\\
        \midrule
        \multirow{2}{*}{\textbf{KNN Gain}}&best&\underline{1.015}&1.002&\textbf{1.019}&.980&.943&.968\\
        &mad&\underline{.023}&.014&\textbf{.010}&.012&.017&.011\\
        \midrule
        \multirow{2}{*}{\textbf{Feat. Pres.}}&best&\underline{.937}&.933&\textbf{.986}&.819&.640&.739\\
        &mad&\textbf{.043}&.042&\textbf{.043}&.014&.012&.006\\
        \midrule
        & & \textsc{ill} &\textsc{lin} & \textsc{s-ump} &\textsc{ill-uc} &\textsc{lin-uc} &\textsc{ump}  \\
        \midrule
        \multirow{2}{*}{\textbf{Dec. Pres.}}&best&\textbf{.700}&\underline{.670}& .666&.635&.634&.597\\
        &mad&\underline{.012}&\textbf{.011}&.014&.013&.012&.005\\
        \midrule
    \end{tabular}
    
    \makebox[\linewidth]{
    \begin{overpic}[trim={3mm 5mm 3mm 9mm},clip,height=0.26\linewidth]{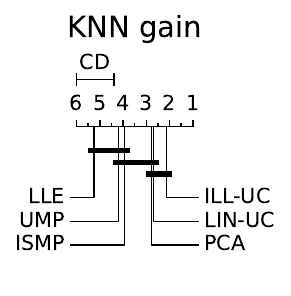}
    \put (22,-5) {\textbf{KNN Gain}}
    \end{overpic}
    \begin{overpic}[trim={8mm 5mm 4mm 9mm},clip,height=0.26\linewidth]{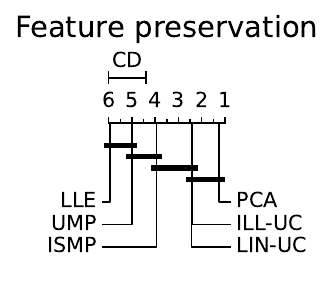}
    \put (5,-5) {\textbf{Feature Preservation}}
    \end{overpic}
    \begin{overpic}[trim={6mm 5mm 6mm 9mm},clip,height=0.26\linewidth]{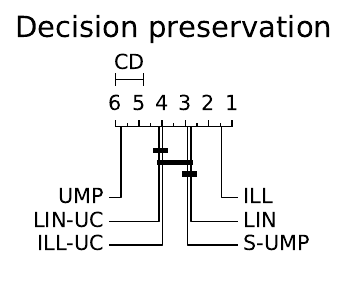}
    \put (6,-5) {\textbf{Decision Preservation}}
    \end{overpic}
    }
\end{table}

\subsection{Results and Discussion}
In the following, we describe and discuss the main findings from the experiments. Within real-world data,  
tables display the average of best metrics across all datasets,
along with average sensitivity measured by the median absolute deviation across all hyperparameters and black-boxes.
Top methods are highlighted in bold for every metric, while the second-highest results are underlined. Among those, we highlight the less sensitive ones as well. Additionally, we provide Critical Difference (CD) plots to compare statistically significant average ranks (with the null hypothesis rejected at $p\text{-}value$<$.001$), determined using the non-parametric Friedman test,  across multiple methods based on a single evaluation measure~\cite{demsar2006statistical}.
Two methods are tied if the null hypothesis that their performance is the same cannot be rejected using the Nemenyi test at 90\% confidence level. 

\smallskip
\noindent\textbf{RQ1 - Latent space quality.}
Using real-world datasets, we compared \approach{} against 
dimensionality reduction frameworks to assess the quality of the latent spaces in preserving neighborhoods information. 
In Table~\ref{tab:space_quality}, is reported the \textit{KNN Gain}~\cite{bodria2022transparent}, defined as the ratio of a KNN classifier's accuracy in the latent space to its accuracy in the original one, $\frac{acc_{KNN}(Z)}{acc_{KNN}(X)}$.
This evaluation is based on the principle of homophily, assuming that instances within the same ground-truth class are closely clustered together, an effective latent encoding should reinforce these similarities, enabling gains when the latent configuration of instances is better organized.
At this stage, for a fair comparison with 
unsupervised 
reduction
methods, we remove the effects of label conditioning by studying \approach{}-\textsc{uc}. Moreover, we did not observe significant improvement with regularizations. 
Our analysis reveals that \approach{}-\textsc{uc} ranks as the top-performing method, while its average performance is comparable to that of \textsc{pca}, but with slightly higher variability with respect the latent dimensionality. 

In Table~\ref{tab:space_quality} we also assess whether the global arrangement of the neighborhoods is preserved in terms of \emph{Feature Preservation} and \emph{Decision Preservation} calculated 
with \textit{triplet accuracy}.
The \textit{triplet accuracy}~\cite{wang2021understanding} measures the percentage of triplets for which the relative ordering of pairwise distances remains consistent between the original and projected spaces.
Thus, \emph{Feature Preservation} compares the pairwise latent distance orderings $||\mathbf{z}_i - \mathbf{z}_j||_2$ with the original feature-based distances $||\mathbf{x}_i - \mathbf{x}_j||_2$, while the \emph{Decision Preservation} compares 
$||\mathbf{z}_i - \mathbf{z}_j||_2$ with relative distances of black-box decisions $||b(\mathbf{x})_i - b(\mathbf{x})_j||_2$.
In line with the \textit{KNN Gain}, when comparing feature preservation across 
unsupervised
methods, \approach{}\textsc{-uc} ranks second only to \textsc{pca} and performs similarly to \textsc{lin-uc}, having all these methods similar sensitivity to dimension size. 
On the other hand, 
when comparing label-aware methods in terms of decision preservation, 
label conditioning in \approach{} is determinant for significantly enhancing triplet accuracy, even outperforming \textsc{lin} and \textsc{s-umap}, i.e., \textsc{umap} trained in supervised setting. 
These results highlight \approach{}'s ability to capture black-box decision logic and feature proximity in latent representations.

\begin{table}[t]
    \centering
    \setlength{\tabcolsep}{0.8mm}
    \caption{RQ2 - Correctness of synthetic explanations. 
    Prediction accuracy of surrogate models inside parentheses.
    }
    \label{tab:seneca}

    \begin{tabular}{l@{\hspace{5pt}}c@{\hspace{5pt}}c@{\hspace{5pt}}c@{\hspace{5pt}}c@{\hspace{5pt}}c@{\hspace{5pt}}}
        \textbf{\textsc{seneca-rc}}&\multicolumn{5}{c}{\textbf{Feature Importance Correctness}}\\
        \cline{2-6}
        $t+u$& $4+0$ & $8+0$ & $16+0$ & $16+16$ & $16+48$ \\
        \midrule
        \textsc{ill-lr}{\scriptsize(co)} & \underline{.588} \scriptsize{(87.6)} & \underline{.476} \scriptsize{(79.9)} & \textbf{.181} \scriptsize{(78.6)} & \textbf{.133} \scriptsize{(77.5)} & .077 \scriptsize{(73.3)} \\
        \textsc{ill-lr-uc}{\scriptsize(co)} & \textbf{.603} \scriptsize{(87.2)} & \textbf{.479} \scriptsize{(79.6)} & \underline{.170} \scriptsize{(79.1)} & \underline{.124} \scriptsize{(77.0)} & \textbf{.085} \scriptsize{(72.7)} \\
        \textsc{ill-lr-us}{\scriptsize(co)} & .503 \scriptsize{(85.1)} & .393 \scriptsize{(77.6)} & \textbf{.181} \scriptsize{(78.8)} & .110 \scriptsize{(73.1)} & \underline{.081} \scriptsize{(72.7)} \\
        \midrule
        \textsc{lin-lr}{\scriptsize(co)} & .289 \scriptsize{(79.1)} & .226 \scriptsize{(73.7)} & .163 \scriptsize{(77.3)} & .094 \scriptsize{(74.8)} & .022 \scriptsize{(66.0)} \\
        \textsc{inp-lr} & .275 \scriptsize{(78.3)} & .217 \scriptsize{(73.2)} & .107 \scriptsize{(75.3)} & .092 \scriptsize{(74.9)} & \underline{.080} \scriptsize{(75.8)} \\
        \textsc{lime} & .420 & .267 & .102 & .076 & .054 \\
        \textsc{shap} & .303 & .350 & .030 & .031 & .030 \\
        \midrule
    \end{tabular}
   
    \begin{tabular}{l@{\hspace{4pt}}c@{\hspace{4pt}}c@{\hspace{4pt}}c@{\hspace{4pt}}c@{\hspace{4pt}}c@{\hspace{4pt}}}
        \textbf{\textsc{seneca-rb}}&\multicolumn{5}{c}{\textbf{Decision Rule Correctness}}\\
        \cline{2-6}
        $t+u$& $4+0$ & $8+0$ & $16+0$ & $16+16$ & $16+48$ \\
        \midrule
        \textsc{ill-dt}{\scriptsize(so,$\alpha$=2)} & .531 {\scriptsize (71.6)} & .339 {\scriptsize (74.4)} & \underline{.240} {\scriptsize (73.3)} & \underline{.200} {\scriptsize (71.9)} & \underline{.166} {\scriptsize (70.3)} \\
        \textsc{ill-dt-uc}{\scriptsize(so,$\alpha$=2)} & .523 {\scriptsize (72.1)} & .318 {\scriptsize (73.9)} & .209 {\scriptsize (71.3)} & .165 {\scriptsize (69.3)} & .128 {\scriptsize (66.6)} \\
        \textsc{ill-dt-us}{\scriptsize(so,$\alpha$=2)} & .504 {\scriptsize (70.9)} & .308 {\scriptsize (74.4)} & .218 {\scriptsize (73.1)} & .199 {\scriptsize (73.1)} & .143 {\scriptsize (68.8)} \\
        \midrule
        \textsc{lin-dt}{\scriptsize(so,$\alpha$=2)} & \underline{.545} {\scriptsize (71.2)} & .344 {\scriptsize (74.8)} & .226 {\scriptsize (73.8)} & .165 {\scriptsize (71.3)} & .116 {\scriptsize (65.4)} \\
        \textsc{inp-dt} & \underline{.545} {\scriptsize (70.5)} & \textbf{.356} {\scriptsize (72.8)} & \textbf{.266} {\scriptsize (71.2)} & \textbf{.244} {\scriptsize (69.8)} & \textbf{.227} {\scriptsize (69.6)} \\
        \textsc{lore} & \textbf{.557} & \underline{.346} & .202 & .150 & .123 \\
        \textsc{anchor} & .402 & .326 & .204 & .156 & .139 \\
        \midrule
    \end{tabular}
    
\end{table}

\smallskip
\noindent\textbf{RQ2 - Correctness of explanations.} 
We employed \textsc{seneca-rb} and \textsc{seneca-rc}~\cite{guidotti2021evaluating} to study the \emph{explanation correctness} of local model-agnostic explainers on tabular data\footnote{In the repository accompanying the paper, we extend the evaluation with additional synthetic benchmarks~\cite{cortez2013using}, obtaining comparable results.}. 
By exploiting synthetic transparent classifiers and treating them as black-boxes, we can directly compare the explanations provided by an explainer with the ground-truth decision logic of the synthetic black-box\footnote{Please refer to~\cite{guidotti2021evaluating} for the definitions of ground-truth explanations.}. 
We evaluate the correctness of local explanations by measuring the closeness between the extracted explanations $\epsilon$ and the ground-truth $\hat{\epsilon}$ provided by the synthetic classifiers of \textsc{seneca}. 
Following~\cite{guidotti2021evaluating,rahnama2023blame}, for feature importance we measure the proximity of two explanations with the \textit{cosine similarity score}
of attribution vectors: $\mathit{cs}\text{-}\mathit{score}(\psi, \hat{\psi}) = \frac{\psi\cdot\hat{\psi}}{||\psi||~||\hat{\psi}||}$. 
Besides, for decision rules we measure the similarity of the bounded regions of the instance space described by two rules, calculating the closeness between upper and lower bounds when both are different from $\pm \infty$ (\textit{complete rule score}\footnote{ $N_{\fy[\infty]}$ denotes the number of finite lower/upper bounds in a decision rule. When an infinite
bound is 
subtracted from 
a finite one, the 
fraction
counts zero ($\frac{1}{1\pm\infty}=0$).
Our metric spans between $0$ and $1$ and captures  differences in a continuous spectrum, while in \cite{guidotti2021evaluating} a binarized metric is used.
}): 
$\mathit{cplt}\text{-}\mathit{score}(\rho, \hat{\rho}) = \frac{1}{N_{\fy[\infty]}}(\sum_j \frac{1}{1+|l_j-\hat{l}_j|^2} + \sum_j\frac{1}{1+|u_j-\hat{u}_j|^2})$.

\begin{table}[t]
    \caption{RQ3 - Quality for feature importance.}
    \label{tab:fimp}
    \centering
    \setlength{\tabcolsep}{0.9mm}

    \begin{tabular}{l@{\hspace{2pt}}c@{\hspace{5pt}}c@{\hspace{5pt}}c@{\hspace{5pt}}c@{\hspace{5pt}}c@{\hspace{5pt}}c@{\hspace{5pt}}}
        
        \midrule
        \multirow{2}{*}{\textbf{Robustness}}&\textsc{ill-lr} & \textsc{ill-lr-us} & \textsc{lin-lr} & \textsc{inp-lr} & \textsc{lime} & \textsc{shap} \\
        &(co) & (co) & (co) & & & \\
        \midrule
         best&\underline{.959}&.863&.494&.414&\textbf{.978}&.373\\
         mad&\textbf{.062}&.112&.042&.016&\underline{.103}&.037\\
    \end{tabular}

    \begin{tabular}{l@{\hspace{2pt}}c@{\hspace{5pt}}c@{\hspace{5pt}}c@{\hspace{5pt}}c@{\hspace{5pt}}c@{\hspace{5pt}}c@{\hspace{5pt}}}
        \midrule
        
        \multirow{2}{*}{\textbf{Faithfulness}}&\textsc{ill-lr} & \textsc{ill-lr-uc} & \textsc{lin-lr} & \textsc{inp-lr} & \textsc{lime} & \textsc{shap} \\
        &(.) & (.) & (so,$\alpha$=2) &  & &  \\
        \midrule
         best&\underline{.656}&.397&.496&.485&.230&\textbf{.663}\\
         mad&\underline{.097}&.050&.039&.026&.049&\textbf{.026}\\
        \midrule
    \end{tabular}
    
    \makebox[\linewidth]{
    \begin{overpic}[trim={2mm 4mm 2mm 8mm},clip,height=0.3\linewidth]{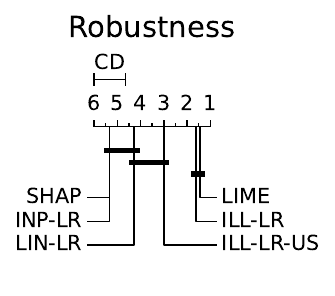}
    \put (26,0) {\textbf{Robustness}}
    \end{overpic}
    \begin{overpic}[trim={2mm 4mm 2mm 8mm},clip,height=0.3\linewidth]{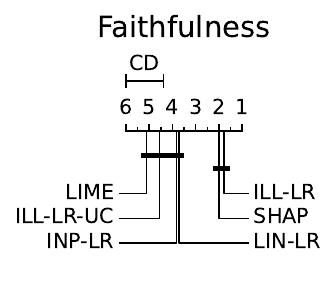}
    \put (36,0) {\textbf{Faithfulness}}
    \end{overpic}
    }

\end{table}

Table~\ref{tab:seneca} reports correctness synthetic explanations from \textsc{seneca-rc} and \textsc{seneca-rb}, using the best regularizer for \approach{} and \textsc{lin}. 
Correctness generally decreases with noisy input dimensions denoted with $u$ (while $t$ refers to informative dimensions). \textsc{ill-lr} consistently outperforms \textsc{lin-lr} and \textsc{inp-lr} in surrogate performance.  
For feature importance explanations, \textsc{ill-lr} and its variants (\textsc{ill-lr-uc}, \textsc{ill-lr-us}) significantly surpass competitors, with label conditioning and consistency having minimal impact. 
For decision rule explanations, \textsc{ill-dt} ranks second-best for input dimensions $\geq 16$, always outperforming \textsc{ill-dt-uc}/\textsc{ill-dt-us} and surpassing \textsc{lin-dt} for dimensions $\geq 8$. 
\textsc{inp-dt} performs best, as expected, since \textsc{seneca-rb} is based on decision trees trained in the input space, aligning its structure with the ground-truth logic.

\smallskip
\noindent\textbf{RQ3 - Robustness and faithfulness of explanations.} 
Resorting real-world data\-sets, we compared \approach{} against local model-agnostic explainers to assess the quality of the resulting explanations.
Since ground-truth for individual explanations is unavailable in real-world data~\cite{guidotti2021evaluating}, we focus on other criteria that are critical for explanations' evaluation, i.e., \emph{(i)} their sensitivity to feature modifications (\emph{Robustness})~\cite{alvarez2018robustness}, and \emph{(ii)} their ability to accurately reflect the reasoning of the black-box model (\emph{Faithfulness} or fidelity)~\cite{dasgupta2022framework}. 
In line with~\cite{yeh2019fidelity}, we evaluate 
robustness for each instance as the maximum change in the explanation with small input perturbations. 
For each test instance $\mathbf{x}_i$, we compute the average \textit{max-sensitivity} over multiple nearest neighborhoods ($K_{max}=20$)
$\frac{1}{K_{max}} \sum_{K=1}^{K_{max}} \mathrm{max}_{j \in \mathcal{N}_K^=(i)} d_\mathcal{E}(i, j)$, where $\mathcal{N}^=_K(i)$ denotes the set of $K$ nearest neighbors of $\mathbf{x}_i$ with same predicted black-box label, and $d_\mathcal{E}(i, j)$ is a suitable pairwise explanations distance. 
In line with~\cite{dasgupta2022framework}, we evaluate fidelity by assessing whether similar black-box predictions result in similar explanations.  
Thus, faithfulness is evaluated by measuring the rank correlation between the pairwise distances of explanations, $\{d_\mathcal{E}(i, j)\}_{i<j}$, and the corresponding pairwise differences in black-box predictions, $\{\rvert \rvert b(\mathbf{x}_i) -  b(\mathbf{x}_j)\rvert \rvert_2\}_{i<j}$.
As for \textsc{seneca}, we use 
proximity
metrics $\mathit{cs}\text{-}\mathit{score}$ and $\mathit{cplt}\text{-}\mathit{score}$ 
to evaluate explanation
(dis)similarity.
For robustness, we select the most dissimilar explanations within each fixed-size neighborhood. 

Tables~\ref{tab:fimp} and~\ref{tab:drule} present the results, with CD plots summarizing overall performance across datasets and black-boxes. 
As generally expected, latent methods without stability optimization show weaker robustness, and removing decision conditioning compromises faithfulness. 
Regarding importance-based explainers in Table~\ref{tab:fimp}, \textsc{lime} demonstrates the highest robustness, while \textsc{shap} is the most faithful, even though they do not  excel in both metric. Conversely, the CD plots highlight that \textsc{ill-lr} is statistically comparable to the best of them in both metrics. Moreover, in reference to robustness, it appears 
also less sensitive to hyperparameters than \textsc{lime}. 
Regarding rule-based explainers in Table~\ref{tab:drule}, latent methods \textsc{ill-dt} and \textsc{lin}-\textsc{dt} exhibit strong robustness, while \textsc{lin}-\textsc{dt} being more sensitive to hyperparameters. 
For faithfulness, \textsc{ill-dt} is the most effective, being less sensitive than its runner-up \textsc{lin}-\textsc{dt}.

\begin{table}[t]
    \caption{RQ3 - Quality for decision rules.}
    \label{tab:drule}
    \centering
    \setlength{\tabcolsep}{1mm}

    \begin{tabular}{l@{\hspace{2pt}}c@{\hspace{3pt}}c@{\hspace{3pt}}c@{\hspace{3pt}}c@{\hspace{3pt}}c@{\hspace{3pt}}c@{\hspace{2pt}}}

        \midrule
        \multirow{2}{*}{\textbf{Robustness}}&\textsc{ill-dt} & \textsc{ill-dt-us} & \textsc{lin-dt} & \textsc{inp-dt} & \textsc{lore} & \textsc{anchor} \\
        &(so,$\alpha$=2) & (so,$\alpha$=2) & (so,$\alpha$=2) & & &  \\
        
        \midrule
         best&\underline{.505}&.440&\textbf{.672}&.439&.225&.221\\
         mad&\textbf{.058}&.053&\underline{.084}&.018&.021&.018\\
    \end{tabular}
    
    \begin{tabular}{l@{\hspace{2pt}}c@{\hspace{3pt}}c@{\hspace{3pt}}c@{\hspace{3pt}}c@{\hspace{3pt}}c@{\hspace{3pt}}c@{\hspace{2pt}}}
        \midrule

        \multirow{2}{*}{\textbf{Faithfulness}}&\textsc{ill-dt} & \textsc{ill-dt-uc} & \textsc{lin-dt} & \textsc{inp-dt} & \textsc{lore} & \textsc{anchor} \\
        &(so,$\alpha$=2) & (so,$\alpha$=2) & (so,$\alpha$=2) & & &  \\
        
        \midrule
         best&\textbf{.658}&.578&\underline{.626}&.605&.500&.600\\
         mad&\textbf{.058}&.077&\underline{.069}&.021&.034&.020\\
        \midrule
    \end{tabular}
    
    \makebox[\linewidth]{
    \begin{overpic}[trim={2mm 4mm 2mm 8mm},clip,height=0.3\linewidth]{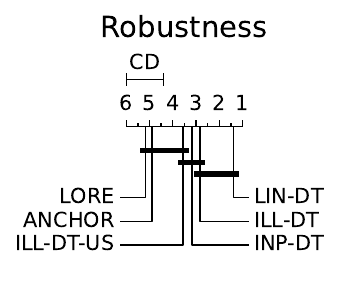}
    \put (36,0) {\textbf{Robustness}}
    \end{overpic}
    \begin{overpic}[trim={2mm 4mm 2mm 8mm},clip,height=0.3\linewidth]{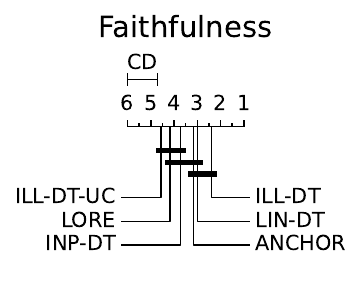}
    \put (34,0) {\textbf{Faithfulness}}
    \end{overpic}
    }
\end{table}

\smallskip
\noindent\textbf{RQ4 - Efficiency of explanations.}
We measured the inference time required to generate feature importance on synthetic datasets with ground-truth explanations (see \textbf{RQ2}). 
In \approach{}, the explanation generator is shared across all instances. Thus, once the model has been trained, it can produce explanations for unseen instances with a single forward pass through the meta-encoder and the surrogate model. Because conventional local surrogates have to retrain the model for every instance, the inductive capabilities of our explainer make it substantially more efficient~\cite{luo2020parameterized}. In fact, training costs in \approach{} incur only once, and involve learning the parameterized function $f^b$ via mini-batches, scaling efficiently to large datasets without significant memory constraints.
Importantly, the meta-encoder is memory-saving because dynamically computes instance-specific matrices $\{W^b_i\}$ during inference, without the need of storing them.  To quantify computational benefits, 
we denote with explanation time as the time required to explain a new instance with an explainer that has already been trained. Because local surrogates train an independent model for every instance, their reported times include also the training cost. For \approach{}, the training cost is listed separately in parentheses.

In Table~\ref{tab:runtime}, we show average per-instance explanation times for producing the most accurate explanations (whose correctness scores are shown in Table~\ref{tab:seneca}), revealing that
\approach{} delivers explanations orders of magnitude faster than \textsc{lime} and \textsc{shap}. Its training time is comparable or lower than the per-instance training costs of \textsc{lime} and \textsc{shap}, an expense that those methods incur for every instance to explain. The figure below illustrates how \approach{}'s performance scales with the number of training instances: the test \textit{cs-score} improves as additional instances are used, confirming the effectiveness of our approach. Notably, the method achieves competitive explanations even when trained on small fractions of training data, making it especially practical for large-scale datasets.

\begin{table}[t]
    \centering
    \setlength{\tabcolsep}{0.8mm}
    \caption{RQ4 - Efficiency of explanations.
    }
    \label{tab:runtime}

    \begin{tabular}{l@{\hspace{5pt}}c@{\hspace{5pt}}c@{\hspace{5pt}}c@{\hspace{5pt}}c@{\hspace{5pt}}c@{\hspace{5pt}}}
        \textbf{\textsc{seneca-rc}}&\multicolumn{5}{c}{\textbf{Per-instance Explanation Time (s)}}\\
        \cline{2-6}
        $t+u$& $4+0$ & $8+0$ & $16+0$ & $16+16$ & $16+48$ \\
        \midrule
        \multirow{2}{*}{\textsc{ill-lr}} & 4.4$\cdot10^{-3}$ & 5.3 $\cdot10^{-3}$ & 4.5 $\cdot10^{-3}$ & 6.6 $\cdot10^{-3}$ & 4.7 $\cdot10^{-3}$ \\
         & (13.0$\pm$.2) & (12.2$\pm$.1) & (8.5$\pm$.08) & (15.1$\pm$.2) & (8.2$\pm$.1) \\
        \midrule
        \textsc{lime} & 3.0$\pm$.4 & 8.3$\pm$1.3 & 26.9$\pm$7.5 & 27.1$\pm$6.8 & 28.2$\pm$6.6 \\
        \textsc{shap} & .1$\pm$.1 & .5$\pm$.1 & 20.9$\pm$7.6 & 22.2$\pm$8.0 & 25.7$\pm$9.5 \\
        \midrule
    \end{tabular}
\includegraphics[width=\linewidth]{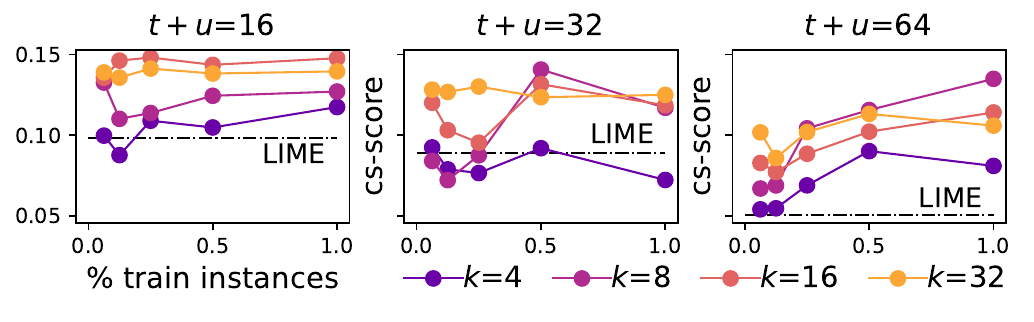}
\end{table}

\section{Conclusions}
\label{sec:conclusion}
We have introduced \approach{}, a 
generative
framework for local explanations compatible with any interpretable surrogate model. 
Unlike traditional 
surrogate-based explainers, which often fail to satisfy desirable explanation properties, \approach{} represents a paradigm shift by combining global reasoning capabilities of interpretable surrogates with local linear encodings of input features.
Through extensive experiments on tabular datasets, we have empirically demonstrated that \approach{} produces explanations that are accurate, robust, and faithful, achieving performance comparable to or surpassing state-of-the-art attribution-based and rule-based explainers in most cases. 

As future work, \approach{} can be extended to end-to-end training a meta-surrogate model, where, like self-interpretable models~\cite{ji2025self}, explainability is built-in architecturally and enforced through task-specific constraints.
Moreover, by leveraging recent advances in tabular foundation models~\cite{breugel2024tabular}, the presented methodology can be further generalized to develop cross-dataset explanation inference frameworks.

\bibliographystyle{IEEEtran}
\bibliography{IEEEabrv,biblio_short}

\appendix

\setcounter{table}{0}
\renewcommand{\thetable}{A\arabic{table}}

\setcounter{figure}{0}
\renewcommand{\thefigure}{A\arabic{figure}}

\subsection{Detailed Experimental Settings}
\label{app:settings}

\textbf{Datasets.}
We implement synthetic classifiers from \textsc{seneca} framework~\cite{guidotti2021evaluating} by varying the dimensions of the synthetic instances. 
In particular, we adjust the number of informative features ($t$), and uninformative features ($u$). 
The total number of features, $m=t+u$, is set in the list $\{4,8,16,32,64\}$ and, for a fixed $t+u$, we define $t = \mathrm{min}\{16, t+u\}$. 
For each dataset dimension, we generate five rule-based classifiers and five linear classifiers. 
Hence, we explain 2,048 instances for each of them. 
Besides synthetic data, we employ 18 real-world datasets from UCI ML Repo\footnote{\url{https://archive.ics.uci.edu/ml/index.php}}.
In every dataset, we apply standard scaling, with zero mean and unitary variance to continuous features, and one-hot encoding to categorical features. 

\textbf{Black-boxes.}
As black-box classifiers, 
we consider the ensemble methods XGBoost
(XGB)~\cite{chen2016xgboost}, LightGBM (LGB)~\cite{ke2017lightgbm} and CatBoost (CTB)~\cite{prokhorenkova2018catboost},
because they are among the most effective techniques for tabular data, even outperforming deep learning models~\cite{grinsztajn2022tree, shwartz2022tabular}. Anyway, there is no specified limitation in using other black-boxes, such as NNs or SVMs.
Each black-box 
is trained using 80\% of the dataset, with the remaining 20\% reserved for testing and evaluation. The black-box models are trained on each dataset's classification tasks, whether binary or multi-class. 
For binary datasets, each explanation method is asked to generate explanations for the class 1.
For multi-class datasets, each explanation method is asked to generate explanations for the majority class.

\textbf{Experimental setup for ILLUME.}
The function $f^b$ is implemented as a 3-layer fully-connected neural network. We did not tune the architecture with respect to the best number of layers and/or hidden sizes. Meta-encoder training is done with Adam optimizer, fixing learning rate to $10^{-3}$, with early stopping technique to prevent overfitting.
To compute the loss functions, for the latent space vectors we use the cosine distance\footnote{$d_{\mathit{cos}}(\mathbf{u},\mathbf{v}) = 1-\frac{\mathbf{u}\cdot\mathbf{v}}{\rvert\rvert\mathbf{u}\rvert\rvert ~\rvert\rvert\mathbf{v}\rvert\rvert}$}, $d_\mathcal{Z}(i,j) = d_{\mathit{cos}}(\mathbf{z}_i,\mathbf{z}_j)$. 
In line with~\cite{guidotti2024stable, mccane2008distance, foss2019distance}, we divide the input distance into contributions from continuous and categorical (one-hot encoded) features.
Assuming 
$h$ categorical features after one-hot encoding, we express the input distance\footnote{$d_{\mathit{hmm}}(\mathbf{u},\mathbf{v})=\frac{1}{len(\mathbf{u})}\sum_i \mathds{1}[u_i \neq v_i]$} as $d_\mathcal{X}(i,j) = \frac{h-m}{m}d_{\mathit{cos}}(\mathbf{x}^{\mathit{con}}_i, \mathbf{x}^{\mathit{con}}_j) + \frac{h}{m}d_{\mathit{hmm}}(\mathbf{x}^{\mathit{cat}}_i, \mathbf{x}^{\mathit{cat}}_j)$. 
For the black-box score vectors, we employ cosine distance as well.
For latent space transformation matrices, we consider the  average per-column cosine distance $d_\mathcal{W}(i,j) = \frac{1}{k} \sum_r d_{cos}(W^b_{i,:,r},W^b_{j,:,r})$. 
All the encodings are tested tuning the latent space dimension as hyperparameter from the list $\{2,4,8,16,32\}$. 

\smallskip
\textbf{Experimental Setup for Dimensionality Reduction.}
We compare \approach{} against latent embedding methods to evaluate the neighborhood structure preservation in the latent space. Considered methods for dimensionality reduction are \textsc{pca}\footnote{\url{https://scikit-learn.org/stable/modules/generated/sklearn.decomposition.PCA.html}}, 
\textsc{isomap}\footnote{\url{https://scikit-learn.org/stable/modules/generated/sklearn.manifold.Isomap.html}}, 
\textsc{lle}\footnote{\url{https://scikit-learn.org/stable/modules/generated/sklearn.manifold.LocallyLinearEmbedding.html}} and p-\textsc{umap}\footnote{\url{https://umap-learn.readthedocs.io/en/latest/parametric_umap.html}}. 
As for \approach{}, the reduction methods are tested tuning the latent space dimension as hyperparameter from the list $\{2,4,8,16,32\}$. 
For p-\textsc{umap}, we train a 3-layer fully-connected neural network for 50 epochs. For the other methods, we used standard parameters.

\smallskip
\textbf{Experimental Setup for Local Explainers.}
We compare \approach{}
with  well-known local explainers to evaluate the quality of explanations: \textsc{lime}\footnote{\url{https://lime-ml.readthedocs.io/en/latest/lime.html\#module-lime.lime\_tabular}}, \textsc{shap}\footnote{\url{https://shap.readthedocs.io/en/latest/generated/shap.Explainer.html}}
for feature importance; \textsc{lore}\footnote{\url{https://kdd-lab.github.io/LORE_sa/html/index.html}},  \textsc{anchor}\footnote{\url{https://github.com/marcotcr/anchor}}
for decision rules. 
\textsc{lime} is tuned with neighborhood sizes $\{100, 300, 1000, 5000\}$; \textsc{lore} with sizes $\{300, 1000\}$ and $\{1, 5, 10\}$ decision trees; \textsc{anchor} with batch sizes $\{100, 300\}$ and beam sizes $\{4, 10\}$.
With \textsc{shap} we use the kernel\textsc{shap} method for synthetic black-boxes and  tree\textsc{shap} for ensemble-tree black-boxes.

\smallskip
\textbf{Experimental Setup for Global Surrogates.}
Global surrogates models \textsc{lr}\footnote{\url{https://scikit-learn.org/stable/modules/generated/sklearn.linear\_model.LogisticRegression.html}} and \textsc{dt}\footnote{\url{https://scikit-learn.org/stable/modules/generated/sklearn.tree.DecisionTreeClassifier.html}} are trained with latent or input space variables by tuning their main hyperparameters to maximize test set prediction accuracy. 
For surrogates based on \textsc{lr} (\textsc{inp-lr} and \textsc{lin-lr}), as local explanation the --global-- logistic coefficients $\{\beta_1, \beta_2, \dots\}$ are weighted by the feature values of the specific instance (see model-intrinsic additive scores~\cite{rahnama2024can}), namely 
$\psi^{\textsc{inp}}_{i,j} =  \beta_j x_{i,j}$ and $\psi^{\textsc{lin}}_{i,j} = \sum_{r=1}^k \beta_r W^b_{j,r}x_{i,j}$.
For a fair comparison, we apply the latent search for maximizing surrogate fidelity also for \textsc{lin}-\textsc{lr} and \textsc{lin}-\textsc{dt} using distance $d_{\mathcal{Z}}$. 
Instead, for \textsc{lr} and \textsc{dt} we search the nearest neighbor in the input space according to $d_{\mathcal{X}}$. 
For remaining methods --\textsc{lime}, \textsc{shap}, \textsc{lore} and \textsc{anchor}-- we trust each explanation as valid one, since all these methods ensure guarantees for maximal fidelity.

\begin{figure*}[t]
    \centering
    \includegraphics[height=0.25\linewidth]
    {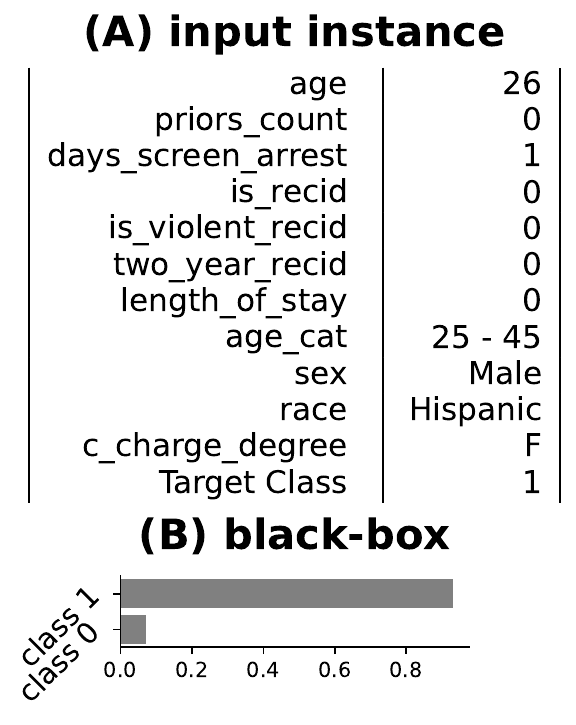}
    \includegraphics[height=0.25\linewidth]{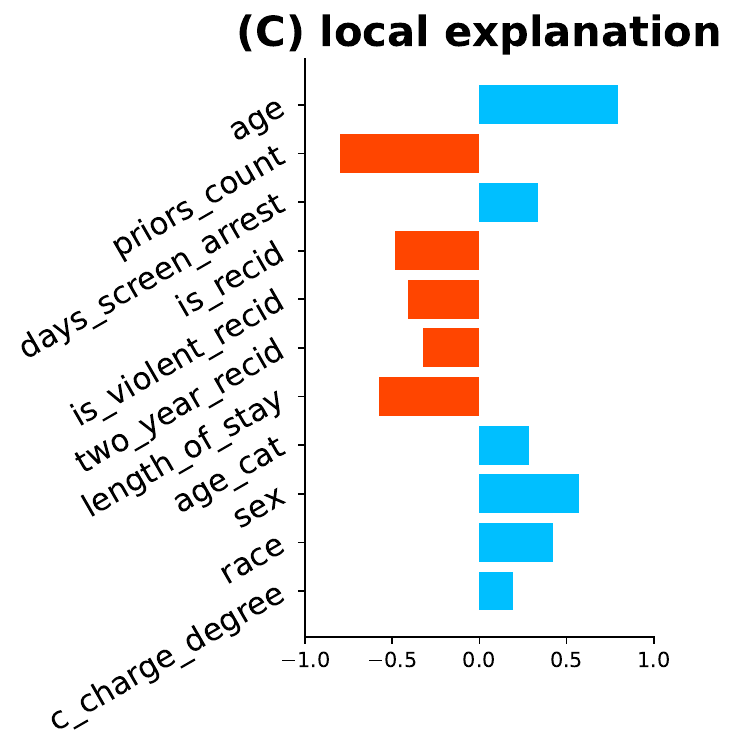}\\
    \begin{minipage}[t]{\linewidth}
    \vspace{-35mm}
    \includegraphics[height=0.12\linewidth]
    {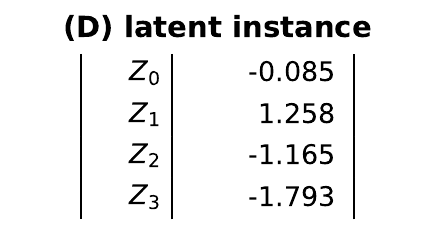}
    \includegraphics[height=0.12\linewidth]{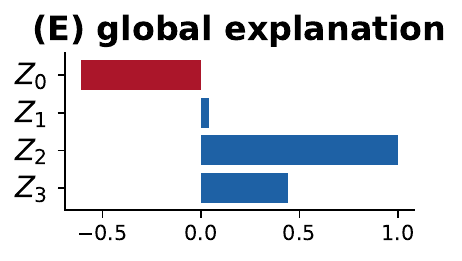}
    \end{minipage}
    \hspace{-105mm}
    \includegraphics[height=0.195\linewidth]{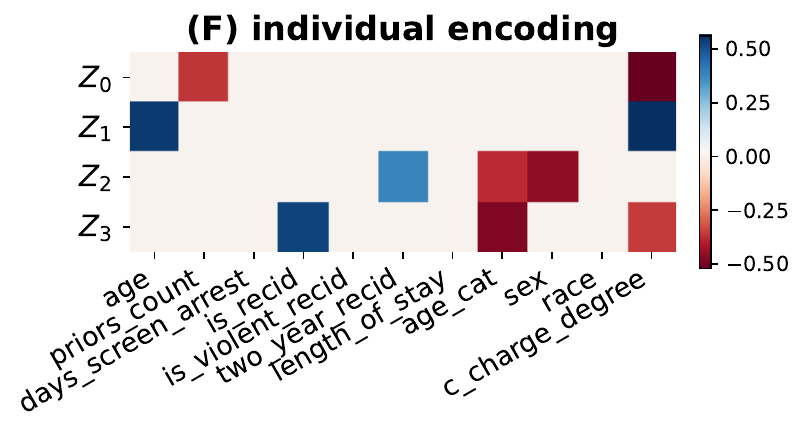}
    \caption{Exemplification of \approach{}'s inference phase on \texttt{compas} dataset when generating feature importance local explanations. Given an input data-point $\mathbf{x}_{test}$ (A) and its corresponding black-box prediction $b(\mathbf{x}_{test})$ (B), the method outputs instance-specific explanation $e_g(\mathbf{z}_{test}, \eta^b_{test})$ (C). This explanation is derived: \emph{(i)} by encoding the instance into a latent representation $\mathbf{z}_{test}$ (D), \emph{(ii)} extracting the logic of the global surrogate $g$ (E), and \emph{(iii)} combining it with local interpretable mapping $\eta^b_{test}$ (F), represented by a sparse and linear transformation returned by the meta-encoder.}
    \label{fig:example1}
\end{figure*}

\begin{figure}[h]
\makebox[\linewidth]{
    \includegraphics[height=0.35\linewidth]{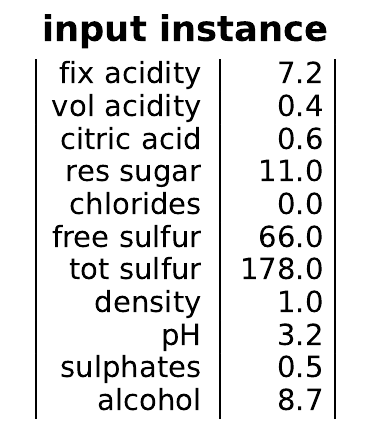}
    \hspace{-4mm}
    \includegraphics[height=0.35\linewidth]{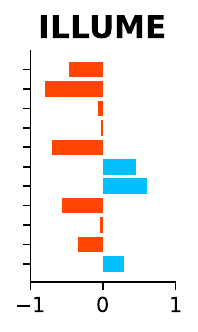}
    \hspace{-4mm}
    \includegraphics[height=0.35\linewidth]{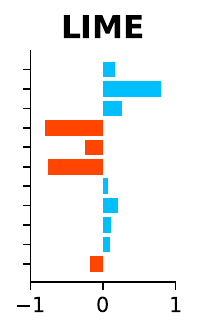}
    \hspace{-4mm}
    \includegraphics[height=0.35\linewidth]{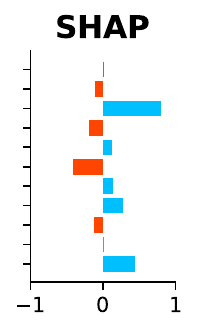}
    \hspace{-6mm}
    \includegraphics[height=0.35\linewidth]{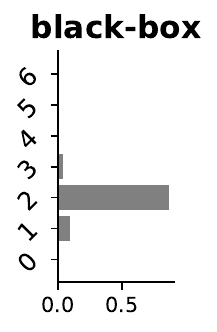}
    }
    \\
    \makebox[\linewidth]{
    \includegraphics[height=0.35\linewidth]{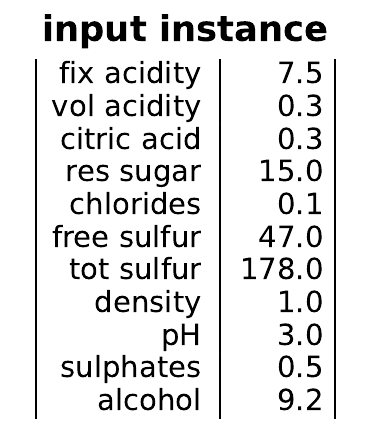}
    \hspace{-4mm}
    \begin{overpic}[height=0.35\linewidth]
    {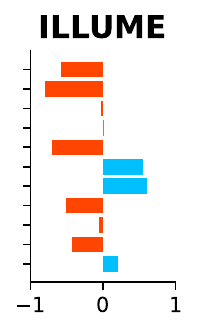}
    \put (10,-10) {corr=0.98}
    \end{overpic}
    \hspace{-4.6mm}
    \begin{overpic}[height=0.35\linewidth]
    {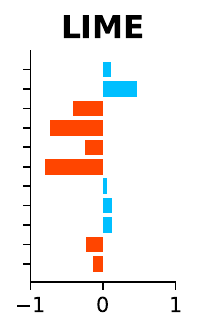}
    \put (10,-10) {corr=0.68}
    \end{overpic}
    \hspace{-4.6mm}
    \begin{overpic}[height=0.35\linewidth]
    {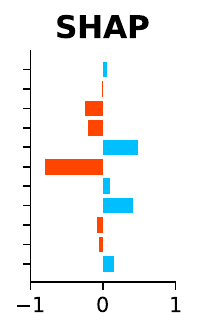}
    \put (10,-10) {corr=0.50}
    \end{overpic}
    \hspace{-6mm}
    \includegraphics[height=0.35\linewidth]{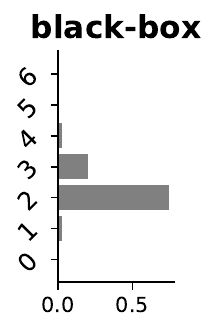}
    }
    \vspace{2mm}
    \caption{Feature importance explanations for two similar records of the \texttt{wine} dataset. Feature values are reported on the left. In the center, explanations are derived with \approach{}, \textsc{lime} and \textsc{shap}. On the right, the prediction probability returned by an XGB classifier.}
    \label{fig:example2}
\end{figure}

\begin{figure}[h]
\makebox[\linewidth]{
    \includegraphics[height=0.35\linewidth]{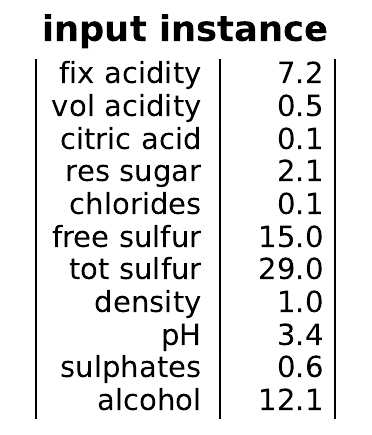}
    \hspace{-4mm}
    \begin{minipage}[t]{0.2\linewidth}
    \vspace{-30mm}
    \centering
    \renewcommand{\arraystretch}{1.}
    \begin{tabular}{@{}c@{}}
        \midrule
        \textbf{\approach{}} \\
        \scriptsize{0.48$<$vol.aci.$\leq$0.98} \\
        \scriptsize{res.sug.$>$1.15} \\
        \scriptsize{tot.sul.$\leq$20.6} \\
        \scriptsize{dens.$\leq$1.01} \\
        \scriptsize{pH$>$3.20} \\
        \scriptsize{sulp.$>$0.55} \\
    \end{tabular}
    \end{minipage}
    \hspace{-2mm}
    \begin{minipage}[t]{0.2\linewidth}
    \vspace{-30mm}
    \centering
    \renewcommand{\arraystretch}{0.8}
    \begin{tabular}{@{}c@{}}
        \midrule
        \textbf{\textsc{lore}} \\
        \scriptsize{fix.aci.$\leq$9.40} \\
        \scriptsize{vol.aci.$\leq$0.71} \\
        \scriptsize{cit.aci.$\leq$0.21} \\
        \scriptsize{res.sug.$>$1.25} \\
        \scriptsize{chlo.$>$0.03} \\
        \scriptsize{tot.sul.$\leq$146.} \\
        \scriptsize{dens.$\leq$1.00} \\
        \scriptsize{sulp.$\leq$1.03} \\
        \scriptsize{11.1$<$alco.$\leq$12.7} \\
    \end{tabular}
    \end{minipage}
    \hspace{-2mm}
    \begin{minipage}[t]{0.2\linewidth}
    \vspace{-30mm}
    \centering
    \renewcommand{\arraystretch}{0.89}
    \begin{tabular}{@{}c@{}}
        \midrule
        \textbf{\textsc{anchor}} \\
        \scriptsize{fix.aci.$\leq$7.70} \\
        \scriptsize{cit.aci.$\leq$0.25} \\
        \scriptsize{1.8$<$res.sug.$\leq$8.0} \\\
        \scriptsize{chlo.$>$0.05} \\
        \scriptsize{fre.sul.$\leq$40.9} \\
        \scriptsize{0.99$<$dens.$\leq$1.0} \\
        \scriptsize{sulp.$>$0.60} \\
        \scriptsize{alco.$>$11.3} \\
    \end{tabular}
    \end{minipage}
    \hspace{-2mm}
    \includegraphics[trim={3mm 0 12mm 0},clip,height=0.35\linewidth]{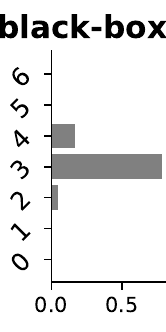}
    }
    \\\hspace{8mm}
    \makebox[\linewidth]{
    \includegraphics[height=0.35\linewidth]{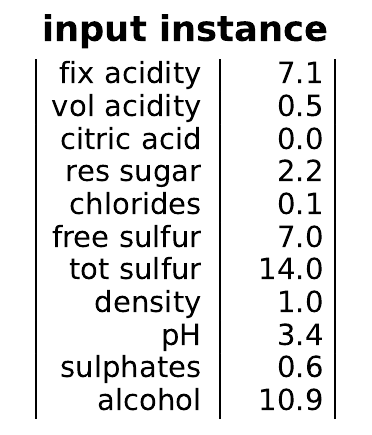}
    \hspace{-4mm}
    \begin{minipage}[t]{0.2\linewidth}
    \vspace{-30mm}
    \centering
    \begin{tabular}{@{}c@{}}
        \midrule
        \textbf{\approach{}} \\
        \scriptsize{0.55$<$vol.aci.$\leq$0.64} \\
        \scriptsize{res.sug.$>$0.55} \\
        \scriptsize{tot.sul.$\leq$23.0} \\
        \scriptsize{pH$>$3.35} \\
        \scriptsize{sulp.$>$0.54} \\
        \\
        \\
        \midrule
        cplt=0.69 \\
    \end{tabular}
    \end{minipage}
    \hspace{-2mm}
    \begin{minipage}[t]{0.2\linewidth}
    \vspace{-30mm}
    \centering
    \renewcommand{\arraystretch}{1.}
    \begin{tabular}{@{}c@{}}
        \midrule
        \textbf{\textsc{lore}} \\
        \scriptsize{vol.aci.$>$0.29} \\
        \scriptsize{res.sug.$\leq$10.3} \\
        \scriptsize{tot.sul.$\leq$47.8} \\
        \scriptsize{0.99$<$dens.$\leq$1.00} \\
        \scriptsize{pH$>$3.03} \\
        \scriptsize{sulp.$>$0.58} \\
        \scriptsize{9.9$<$alco.$\leq$11.4} \\
        \midrule
        cplt=0.12 \\
    \end{tabular}
    \end{minipage}
    \hspace{-2mm}
    \begin{minipage}[t]{0.2\linewidth}
    \vspace{-30mm}
    \centering
    \renewcommand{\arraystretch}{0.800}
    \begin{tabular}{@{}c@{}}
        \midrule
        \textbf{\textsc{anchor}} \\
        \scriptsize{6.4$<$fix.aci.$\leq$7.7} \\
        \scriptsize{cit.aci.$\leq$0.25} \\
        \scriptsize{res.sug.$>$1.82} \\\
        \scriptsize{chlo.$>$0.07} \\
        \scriptsize{fre.sul.$\leq$40.9} \\
        \scriptsize{tot.sul.$\leq$154.} \\
        \scriptsize{0.99$<$dens.$\leq$1.0} \\
        \scriptsize{0.51$<$sulp.$\leq$0.60} \\
        \scriptsize{alco.$>$10.3} \\
        \midrule
        cplt=0.59 \\
    \end{tabular}
    \end{minipage}
    \hspace{-2mm}
    \includegraphics[trim={3mm 0 12mm 0},clip,height=0.35\linewidth]{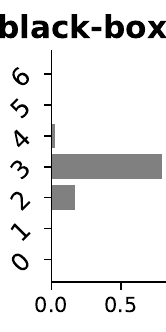}
    }
    \caption{Decision rule explanations for two similar records of the \texttt{wine} dataset. Feature values are reported on the left. In the center, explanations derived with \approach{}, \textsc{lore} and \textsc{anchor}. On the right, the prediction probability returned by an LGB classifier.}
    \label{fig:example3}
\end{figure}

\subsection{Qualitative Results}
\label{app:qualitative}
In Figure~\ref{fig:example1} it is illustrated how our approach works in the explanation inference phase with an example on \texttt{compas} dataset.
At first glance, with \approach{}, instances and their local black-box decisions (A-B) are employed to generate local explanations (C). 
These explanations are computed with the information deriving from an auxiliary embedding space where we project the input (D). 
In this space we have extracted global latent explanations (E) during the surrogate learning process. 
The key component of this space is a set of instance-specific linear transformations (F) learned during the meta-encoder training process. Local explanations are produced by combining the latent explanation with the local projection of the data. 
Figure~\ref{fig:example2} showcase feature importance explanations for two neighboring instances from the \texttt{wine} dataset. 
Being closely located in the feature space and yielding similar predictions from the black-box model for the multi-class task, the explanations for these instances are expected to align,
ranking important features in comparable ways. \approach{} demonstrates its ability to generate robust and consistent explanations, producing rank-preserving feature importance profiles for neighboring instances. 
In contrast, competing methods such as \textsc{lime} and \textsc{shap} fail to achieve similar consistency. 
Additionally, Figure~\ref{fig:example3} presents a comparable example for decision rules using the same dataset, illustrating that \approach{} is capable of generating robust and consinstent rules. 
The similarity of these rules is evaluated using the \textit{cplt-score}, like in the main experimental result of the paper. 
Furthermore, we demonstrate that \approach{} can produce more concise decision rules with respect to the analyzed competitors, a benefit achieved through sparsity regularization.

\subsection{Detailed Formulas for Decision Rules}
\label{app:rules}
In \approach{}, we first derive global explanations in the form of axis-parallel decision rules on latent features, determined by root-leaf paths in a trained decision tree, i.e., $\rho^z_i=\{z_{i,r} \in [l^z_r, u^z_r]\}_{r=1\dots k}$ (lower and upper bound are in the domain of $z_{i,r}$, extended with $\pm \infty$).  
By exploiting the linearity of the mapping $\eta^b_i$, these rules are converted into input space local oblique rules 
$\Tilde{\rho}^x_i=\{\sum_{j=1}^m W^b_{i,j,r}x_{i,j} \in [l^z_r, u^z_r]\}_{r=1\dots k}$. 
While oblique rules offer a valid and expressive form of explanation, for consistency with most methods we perform an additional step to convert the oblique rules into the standard axis-parallel format: $\rho^x_i=\{x_{i,j} \in [l^x_{i,j}, u^x_{i,j}]\}_{j=1\dots m}$. 
The upper and lower bounds for these axis-parallel rules are derived by 
isolating $x_{i,j}$ to express the rule in terms of individual input features.
For example, from the oblique rules $\sum_{j=1}^m W^b_{i,j,r}x_{i,j} \geq l^z_r$ and $\sum_{j=1}^m W^b_{i,j,r}x_{i,j} \leq u^z_r$, it follows: 
$$l^x_{i,j} = \underset{r}{\max}\frac{l^z_r - \sum_{v \neq j} W^b_{i,v,r}x_{i,v}}{W^b_{i,j,r}} $$ 
$$
u^x_{i,j} = \underset{r}{\min}\frac{u^z_r - \sum_{v \neq j} W^b_{i,v,r}x_{i,v}}{W^b_{i,j,r}}.$$ 
The max/min operations ensure taking the most restrictive inequality among the $k$ latent conditions, i.e., the largest lower bound and the smallest upper bound. Finally, exploiting the equality $z_{i,r} = \sum_{v \neq j} W^b_{i,v,r}x_{i,v} + W^b_{i,j,r}x_{i,j}$, we find the relations reported in the main paper  that link upper/lower bounds in the embedding and in the original space:
\begin{equation*}
l^x_{i,j} - x_{i,j} = \underset{r}{\max}\frac{l^z_r - z_{i,r}}{W^b_{i,j,r}}\quad \quad
u^x_{i,j} - x_{i,j} = \underset{r}{\min}\frac{u^z_r - z_{i,r}}{W^b_{i,j,r}}. 
\end{equation*}

\subsection{Perfect Fidelity via Similarity Search} 
\label{app:fidelity}
Relying on surrogate logic, in \approach{} the explanations $e_g(\mathbf{z}_i, \eta_i^b)$ are valid iff $g(\mathbf{z}_i) = b(\mathbf{x}_i)$, i.e., when
surrogate model agrees with the black-box on the corresponding instance. 
To ensure producing valid explanations for \emph{every} instance, we refine the latent representation of those samples for which $g(\mathbf{z}_i) \neq b(\mathbf{x}_i)$.
The goal of this refinement is to realign misclassified latent points with nearby correctly predicted ones that share the same black-box label.
This approach leverages \approach{}'s design, which ensures that nearby points in latent space have similar transformations. Hence, explanations for these points remain closely aligned and reliable.
Thus, for each $\mathbf{z}_i \in Z$ such that $g(\mathbf{z}_i) \neq b(\mathbf{x}_i)$,
we proceed as follows:
\begin{description}

\item[\textbf{(1)}]
\textbf{ Closest valid neighbor search.}
We perform a cosine-based nearest-neighbor search in latent space to identify $\mathbf{z}_{nn} \in Z$ such that 
$g(\mathbf{z}_{nn}) = b(\mathbf{x}_{nn})$ and $b(\mathbf{x}_i)= b(\mathbf{x}_{nn})$. This provides a nearby latent point that is locally consistent with the black-box prediction and serves as a reference.

\item[\textbf{(2)}] \textbf{First-order approximation.} Since $\mathbf{z}_{nn}$ is the closest latent vector to $\mathbf{z}_i$, we assume their corresponding inputs differ by a small displacement (i.e., $\mathbf{x}_{nn} \approx \mathbf{x}_{i} + \delta$).
Similarly, we assume the associated local transformations satisfy $W^b_{nn} \approx W^b_{i} +\Delta W_i$. We use a first-order approximation from the formula $W^b_{nn} \mathbf{x}_{nn} = (W^b_{i} +\Delta W_i)(\mathbf{x}_{i}+\delta)$ to obtain: $$\mathbf{z}_{nn}\approx\mathbf{z}_i + J_i \delta = W^b_i\mathbf{x}_i + \Delta W_i \mathbf{x}_i + W^b_{nn}\delta  + \mathcal{O}(\rvert\rvert\Delta W_i \delta\rvert\rvert).$$ This decomposition separates the effect of input displacement $\delta$ with the projection variation $\Delta W_i$. Here, $W^b_{nn}\delta \approx W^b_{i}\delta$ because we neglected second-order terms.

\item[\textbf{(3)}] \textbf{Interpolated latent representation.}
Guided by the previous approximation, we define the latent perturbation: $$\mathbf{z}^\gamma_i = W^b_i\mathbf{x}_i + \gamma_W(W^b_{nn}-W^b_i)\mathbf{x}_i +\gamma_x W^b_{nn}(\mathbf{x}_{nn}-\mathbf{x}_{i}),$$
where $(W^b_{nn}-W^b_i)$ and $(\mathbf{x}_{nn}-\mathbf{x}_i)$, both scaled by small factors $\gamma_W$ and $\gamma_x$, identify the directions of the perturbations $\delta$ and $\Delta W_i$. This expression can be interpreted as an interpolation toward the nearest valid neighbor:
$$\mathbf{z}^\gamma_i = (1-\gamma_W)\mathbf{z}_i + \gamma_W W^b_{nn}\mathbf{x}_i +\gamma_x W^b_{nn}(\mathbf{x}_{nn}-\mathbf{x}_{i}).$$

\item[\textbf{(4)}] \textbf{Minimal constrained perturbation.} We determine the optimal interpolation parameters by solving
$$
\gamma^*_W, \gamma^*_x = \underset{(\gamma_W, \gamma_x) \in (0,1]^2}{\mathrm{arg~min}} \rvert\rvert \mathbf{z}^\gamma_i - W^b_i\mathbf{x}_i\rvert\rvert^2 ~~\text{s.t.}~~g(\mathbf{z}_i^\gamma) = b(\mathbf{x}_i).$$
This ensures that the refined latent vector remains as close as possible to the original one,
while restoring the agreement between surrogate and black-box predictions. Notably, at least the solution $(\gamma^*_W=1, \gamma^*_x=1)$ exists because $\mathbf{z}^{(\gamma_W=1, \gamma_x=1)}_i\equiv \mathbf{z}_{nn}$ and $g(\mathbf{z}_{nn}) \equiv b(\mathbf{x}_{i})$.
\end{description}
The refined latent instance is then given by
$\mathbf{z}_i^* = W^*_i\mathbf{x}_i + \boldsymbol{\varepsilon}^*_i$, where $W^*_i = W^b_i+\gamma^*_W(W^b_{nn}-W^b_i)\equiv W^b_i + \Delta W_i$ is a updated transformation matrix and $\boldsymbol{\varepsilon}^*_i = \gamma^*_x W^b_{nn}(\mathbf{x}_{nn}-\mathbf{x}_{i}) \equiv W^b_{nn}\delta$ is an induced offset.
In practice, we approximate the solution via grid-search over $(0,1]^2$. Among feasible solutions, we select the one minimizing $\gamma_W + \gamma_x$, favoring minimal corrections in both transformation and input space.
Overall, the refinement corresponds to the smallest local perturbation of the latent mapping that restores surrogate–black-box agreement.

This approach enables the construction of a valid explanation even for those instances where the surrogate is not capable to replicate black-box label. Specifically, for feature importance explanations, the attribution vectors will be linearly updated: $$\psi^*_{i,j} = \sum_{r=1}^k \beta_r (W^b_{i,j,r} +\Delta W_{i,j,r})\equiv \psi_{i,j} + \Delta\psi_{i,j}$$
Furthermore, for decision rule explanations, we will have corrected oblique rules with translated upper/lower bounds as: $$\Tilde{\rho}^*_i=\Big\{\sum_{j=1}^m  (W^b_{i,j,r}+\Delta W_{i,j,r})x_{i,j}\in [l^z_r-\varepsilon^*_{i,r}, u^z_r-\varepsilon^*_{i,r}]\Big\}_{r=1\dots k}$$ 
Once converted to axis-aligned rules $\rho^*_i=\{x_{i,j} \in [l^*_{i,j}, u^*_{i,j}]\}_{j=1\dots m}$, they will satisfy the following: 
$$
l^*_{i,j} - x_{i,j} = \underset{r}{\max}\frac{l^z_r - z^*_{i,r}}{W^b_{i,j,r}+\Delta W_{i,j,r}}$$
$$
u^*_{i,j} - x_{i,j} = \underset{r}{\min}\frac{u^z_r - z^*_{i,r}}{W^b_{i,j,r}+\Delta W_{i,j,r}}. 
$$

\subsection{Counterfactual Rules and Counter-Examples}
\label{app:counterfactuals}

While decision rules are directly obtained from the root-to-leaf paths of a decision tree, counterfactual rules are derived through symbolic reasoning applied to the same tree. Following the methodology proposed in~\cite{guidotti2024stable}, we analyze the latent decision tree (already used to generate the decision rules) to identify paths leading to a prediction opposite to that of the input instance.

For each test instance, we first rank training instances whose surrogate tree predictions differ from that of the test instance based on the cosine similarity between their latent representations. We then examine the decision rules associated with these instances and select those requiring the minimal number of split condition changes with respect to the rule satisfied by the test instance. This strategy identifies the closest instances that achieve the desired prediction change with the smallest possible modifications, thereby optimizing both proximity and sparsity, which are key properties of high-quality counterfactual explanations~\cite{piaggesi2024counterfactual}. 

Finally, we also return the corresponding training instances from which the counterfactual rules are derived, providing them as counterfactual examples.

\subsection{Analysis of Stability Loss}
\label{app:stability}

Here, we provide an intuition behind the stability loss computation, which guarantees that small changes in the original data space lead to minimal perturbations in the latent space, thus ensuring a consistent and reliable mapping across data points. 
We enforce each local transformation to remain valid when applied to slightly perturbed input. 
Specifically, omitting $b$ for simplicity, if the encoding is given by $\mathbf{z} = \eta(\mathbf{x}) = f(\mathbf{x})\cdot\mathbf{x}$, where $f(\mathbf{x})$ represents the individual transformation for $\mathbf{x}$, then the same transformation must hold for a
small perturbation $\mathbf{x} + \delta$. 
In other words, the encoding operates locally around the instance $\mathbf{x}$ as a linear transformation characterized by a matrix $f(\mathbf{x})$, which is approximately constant within the neighborhood of  $\mathbf{x}$. 
While the coefficients of the matrix dynamically adapt to the input, their rate of variation is slower than that of the input $\mathbf{x}$, ensuring stability and consistency of the transformation. This implies the following constraint:
$$
\eta(\mathbf{x} + \delta) = f(\mathbf{x}+ \delta) \cdot(\mathbf{x} + \delta)\approx f(\mathbf{x})\cdot (\mathbf{x} + \delta).
$$

We enforce this property by minimizing specific quantities which ensure the validity of the equation above. 
First, we show the first-order Taylor expansions of matrix entries of the local transformation $f$ under small input perturbations:
$$
f_{j,r}(\mathbf{x}+\delta) = f_{j,r}(\mathbf{x}) + \sum_{v=1}^m \frac{\partial f_{j,r}(\mathbf{x})}{\partial x_v} \delta_v
+ \mathcal{O}(\rvert\rvert\delta\rvert\rvert^2)$$
Substituting these approximations into the expression for the encoding of the perturbation, $\eta(\mathbf{x} + \delta) = f(\mathbf{x} + \delta)\cdot (\mathbf{x} + \delta)$, for each entry of the vector we get (neglecting second-order terms):
\begin{align*}
\eta_r(\mathbf{x} + \delta) \approx & \sum_{j=1}^m \Big( f_{j,r}(\mathbf{x}) + \sum_{v=1}^m \frac{\partial f_{j,r}(\mathbf{x})}{\partial x_v} \delta_v \Big)(x_j + \delta_j)\\
 =& \sum_{j=1}^m f_{j,r}(\mathbf{x}) (x_j+\delta_j)+ \\+ &\sum_{j=1}^m \Big( \sum_{v=1}^m \frac{\partial f_{v,r}(\mathbf{x})}{\partial x_j} x_v \Big) \delta_j + \mathcal{O}(\rvert\rvert\delta\rvert\rvert^2).
\end{align*}
Last equation tells us that describing the variation of the latent encoding $\eta$, when applied to a minimal perturbation of $\mathbf{x}$, requires the sum of two quantities (up to second-order corrections):
$$ \quad 
\eta(\mathbf{x} + \delta) \approx f(\mathbf{x}) (\mathbf{x} + \delta) + D(\mathbf{x}) \cdot \delta \quad \quad  \Big[D_{j,r} = \sum_v \frac{\partial f_{v,r}}{\partial x_j} x_v \Big]
$$
The first term, $f(\mathbf{x})\cdot(\mathbf{x} + \delta)$, represents the mapping of the perturbation $\delta$ applied to the input $\mathbf{x}$, with the linear transformation $f$ held constant. The second term, $D(\mathbf{x}) \cdot \delta$,  captures the change in the transformation due to the perturbation $\delta$ and its interaction with the input $\mathbf{x}$.
Therefore, we enforce stability by minimizing $||D(\mathbf{x})||^2_\text{F}$ during training. Reordering the terms $
\eta(\mathbf{x} + \delta) \approx f(\mathbf{x}) \cdot\mathbf{x}  + [ f(\mathbf{x}) + D(\mathbf{x})] \cdot \delta$, we obtain the Taylor expansion of the encoding $\eta$ under small input perturbations:
$$ \quad 
\eta(\mathbf{x} + \delta) \approx \eta(\mathbf{x}) + J(\mathbf{x})\cdot\delta \quad \quad \quad \Big[J(\mathbf{x}) = f(\mathbf{x})+D(\mathbf{x})  \Big]$$
where we highlight the Jacobian matrix $J$ around the data-point $\mathbf{x}$, with entries $J_{j,r} = \frac{\partial \eta_r}{\partial x_j}$.  To minimize $\vert\vert D(\mathbf{x})\vert\vert^2_\text{F}$, we optimize the reported loss involving Jacobian matrix
$L^{\mathit{st}}(\mathbf{x}) =  
    \vert\vert J(\mathbf{x}) - f(\mathbf{x}) \vert\vert ^2_\text{F}$ for each instance.

\begin{figure*}[h!]
\centering
\includegraphics[width=0.32\linewidth]{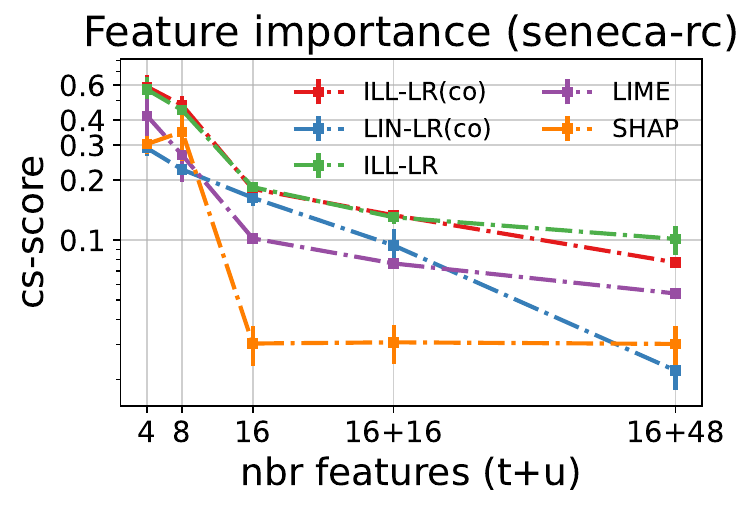}
\includegraphics[width=0.32\linewidth]{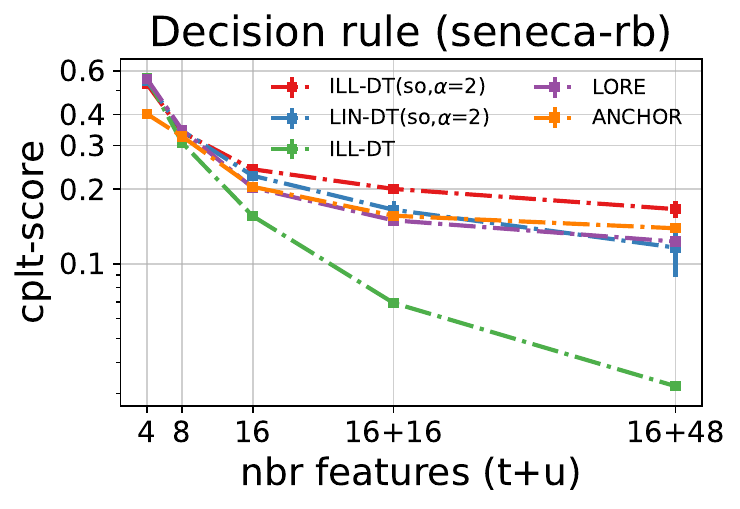}
\includegraphics[width=0.32\linewidth]{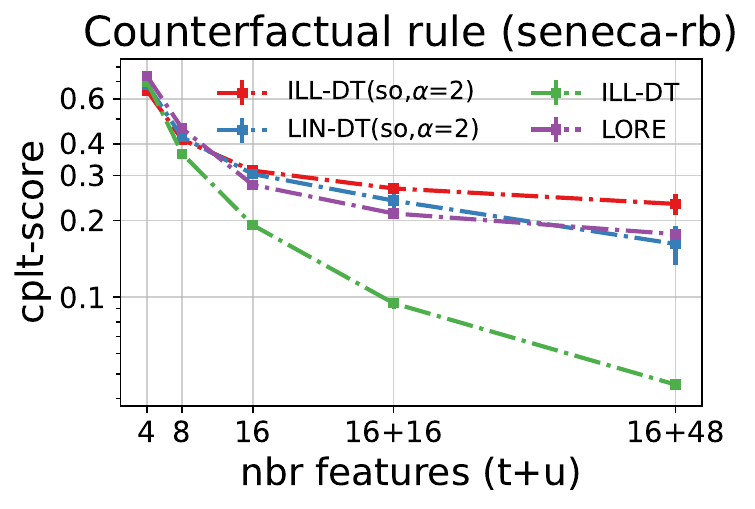}

\includegraphics[height=0.18\linewidth]{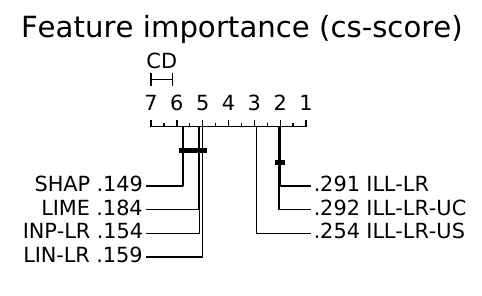}
\includegraphics[height=0.18\linewidth]{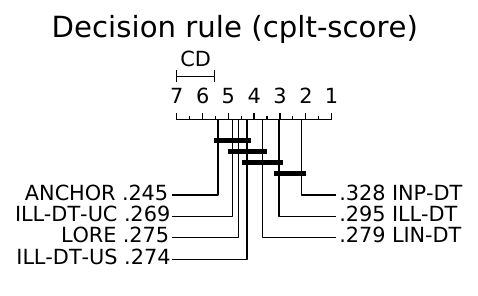}
\includegraphics[height=0.18\linewidth]{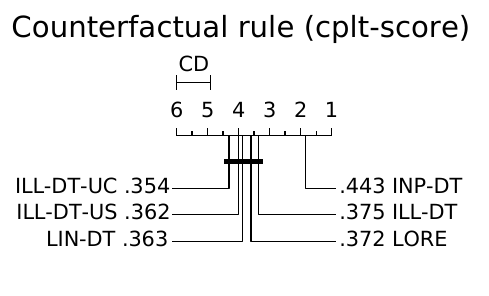}
\caption{Explanation correctness for synthetic classifiers. In the Figures, average metrics for correctness in feature importance and decision rules are reported varying the input feature size. In the CD plots, average rankings for individual metrics are reported for different methods and datasets.}
\label{fig:synth_suppl}
\end{figure*}

\begin{table*}[t]
    \centering
    \caption{Feature-based latent space quality metrics for all datasets and methods.}
    \footnotesize
    
    \begin{tabular}{lcccccc|cccccc}
    \midrule
    &\multicolumn{6}{c}{\textbf{Feature Preservation}}&\multicolumn{6}{c}{\textbf{KNN Gain}}\\
    \cmidrule(lr){2-7}  \cmidrule(lr){8-13}
     & \textsc{ill}\textsc{-uc} &\textsc{lin-uc} & \textsc{pca} & \textsc{isomap} & \textsc{lle} & \textsc{umap} & \textsc{ill}\textsc{-uc} & \textsc{lin-uc} & \textsc{pca} & \textsc{isomap} & \textsc{lle} & \textsc{umap} \\
    \hline
    \texttt{aids}  & .842 &  \underline{.851} & \textbf{1.000} & .706 & .527 & .683 & .946 &  \underline{.960} & \textbf{1.008} & .896 & .791 & .895 \\
    \texttt{austr} & .852 &  \underline{.855} &  \textbf{.993} & .790 & .612 & .732 & \textbf{1.009} & \underline{1.001} & 1.000 & .984 & .960 & .984 \\
    \texttt{bank} &  \underline{.845} & .843 &  \textbf{.952} & .786 & .594 & .666 & .963 & .963 & .959 & .938 &  \underline{.990} &  \textbf{.998} \\
    \texttt{breast} & \textbf{1.000} &  \underline{.996} & \textbf{1.000} & .928 & .746 & .849 & 1.000 & \textbf{1.010} & 1.000 & \underline{1.001} & \underline{1.001} & \underline{1.001} \\
    \texttt{churn} &  \underline{.908} & .861 &  \textbf{.919} & .668 & .521 & .679 & \underline{1.078} & 1.026 & \textbf{1.234} & .977 & .960 & 1.000 \\
    \texttt{compas} & .859 &  \underline{.870} & \textbf{1.000} & .724 & .563 & .643 & .985 & .994 & 1.002 & \underline{1.019} & .976 & \textbf{1.020} \\
    \texttt{ctg}&  \underline{.939} & .927 & \textbf{1.000} & .716 & .635 & .678 & \textbf{1.033} & \underline{1.023} & 1.000 & 1.000 & .996 & .996 \\
    \texttt{diabetes}& \textbf{1.000} & \textbf{1.000} & \textbf{1.000} &  \underline{.845} & .700 & .792 & \textbf{1.019} & \underline{1.000} & \underline{1.000} & .964 & .848 & .894 \\
    \texttt{ecoli}& \textbf{1.000} & \textbf{1.000} & \textbf{1.000} &  \underline{.871} & .806 & .785 & \textbf{1.024} & \underline{1.007} & 1.000 & .977 & .950 & .961 \\
    \texttt{fico} & \textbf{1.000} & \textbf{1.000} & \textbf{1.000} &  \underline{.832} & .657 & .680 & \textbf{1.006} & \underline{1.003} & 1.000 & .984 & .964 & .990 \\
    \texttt{german}&  \underline{.849} & .832 &  \textbf{.945} & .765 & .589 & .686 & .994 & .963 & \underline{1.054} & \textbf{1.067} & 1.034 & .992 \\
    \texttt{home}  & \textbf{1.000} & \textbf{1.000} & \textbf{1.000} &  \underline{.897} & .679 & .826 & \textbf{1.033} & 1.000 & \underline{1.011} & \underline{1.011} & .956 & .966 \\
    \texttt{ionos} &  \underline{.994} & .992 & \textbf{1.000} & .868 & .713 & .749 & \textbf{1.053} & 1.045 & \underline{1.052} & 1.006 & 1.004 & 1.006 \\
    \texttt{sonar} &  \underline{.986} & .967 &  \textbf{.995} & .871 & .671 & .800 & \underline{1.003} & .977 & \textbf{1.027} & .890 & .998 & .946 \\
    \texttt{spam} & .921 &  \underline{.930} &  \textbf{.953} & .816 & .585 & .673 & \textbf{1.016} & \underline{1.015} & .994 & .975 & .924 & .977 \\
    \texttt{titanic} & .868 & .869 & \textbf{1.000} &  \underline{.896} & .603 & .797 & \textbf{1.047} & \textbf{1.047} & 1.000 & \underline{1.014} & .977 & 1.002 \\
    \texttt{wine} & \textbf{1.000} & \textbf{1.000} & \textbf{1.000} &  \underline{.868} & .670 & .789 & \textbf{1.063} & \underline{1.000} & \underline{1.000} & .974 & .890 & .933 \\
    \texttt{yeast}&  \underline{.999} & \textbf{1.000} & \textbf{1.000} & .887 & .652 & .799 & .995 & \textbf{1.000} & \textbf{1.000} & .965 & .746 & .870 \\
    \midrule
    \end{tabular}
\label{tab:quality_all}
\end{table*}

\newpage

\subsection{Detailed Experimental Results}
\label{app:more}
Here, we report in details all the results that are shown in aggregated form in the main paper. 

\smallskip
\noindent\textbf{RQ1 - Goodness of Latent Space}

\begin{figure*}[h]
\centering
\includegraphics[width=0.35\linewidth]{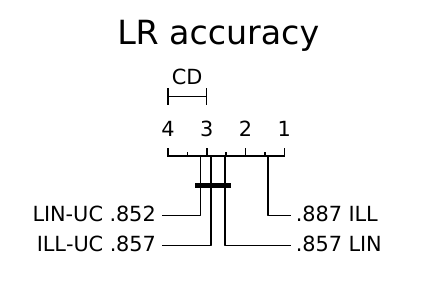}
\includegraphics[width=0.35\linewidth]{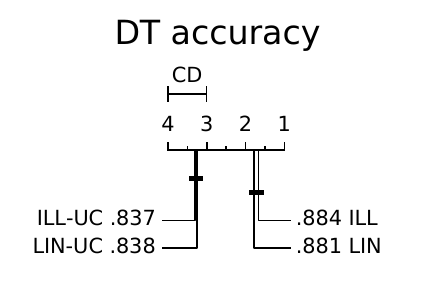}
\caption{Critical Difference plots with Nemenyi test at 90\%
confidence level for prediction accuracy metrics in real-world datasets, comparing \textsc{lr} and \textsc{dt} performances over different training feature spaces.}
\label{fig:surr_suppl}
\end{figure*}

In Tables~\ref{tab:quality_all}, \ref{tab:quality_lgbm} and \ref{tab:quality_xgb} are reported extensive results for all metrics, datasets and black-boxes that have been summarized
in the main paper to answer \textbf{RQ1}. In particular, central and right columns of Tables \ref{tab:quality_lgbm} and \ref{tab:quality_xgb} report Macro-F1 accuracies for \textsc{lr} and \textsc{dt} trained on latent spaces both with and without black-box decision conditioning. The same results are aggregated and ranked in Figure~\ref{fig:surr_suppl}. Firstly, we observe that models without conditioning consistently achieve lower accuracies in surrogate classification tasks. Secondly, particularly for the \textsc{lr} surrogate, \approach{} demonstrates significant improvements compared to \textsc{lin} encodings.
This underscores the importance of label conditioning, combined with using more expressive latent features, in enhancing the accuracy of surrogate predictions.

\smallskip
\noindent\textbf{RQ2 - Correctness for Synthetic Black-Box Explanations}

In Figure~\ref{fig:synth_suppl} we report additional results from synthetic black-box models. 
We also present preliminary results on generating \textbf{counterfactual explanations} using the approach described in \ref{app:counterfactuals}. 

In the top pictures, we display with line plots the variation of explanation accuracy metrics with respect to the number of input dimensions, comparing the best-regularized setup of \approach{} with the un-regularized one.
For feature importance explanations, we observe that the non-collinear ($co$) regularizer slightly enhances the correctness of the explanations across most scenarios. However, in cases involving $16$+$48$ input features, the explanation accuracy experiences a minor decline. In contrast, for both factual and counterfactual rules, the combination of sparsity ($\alpha$=$2$) and soft-orthogonality ($so$) significantly improves the correctness of the explanations. These findings underscore the importance of consistently applying sparsity regularizers within \approach{} when generating rule-based explanations.

In the CD plots, corresponding to the results presented for \textbf{RQ2} in the main paper, 
we display aggregated rankings for each metric across all datasets. We observe the same trends as reported in the main paper: \approach{} and its variants perform similarly in terms of feature importance correctness, while \approach{}\textsc{-dt} shows comparable performance to \textsc{inp-dt} in factual rules correctness. Regarding counterfactual  rules, there is no statistically significant difference between \approach{}\textsc{-dt} and \textsc{lore}. However, \textsc{inp-dt} outperforms the others, consistent with its superior performance on factual rules.

\smallskip
\noindent\textbf{RQ3 - Faithfulness and Robustness in Real-World Datasets}

Tables~\ref{tab:fimp_lgbm} and~\ref{tab:fimp_xgb} present comprehensive results for feature importance metrics, while Tables~\ref{tab:drule_lgbm} and~\ref{tab:drule_xgb} display detailed results for decision rules metrics. 
These tables include all datasets and black-box models that were summarized for answering \textbf{RQ3} in
the main paper. 

\begin{figure*}[h]
\centering
\includegraphics[width=0.49\linewidth]{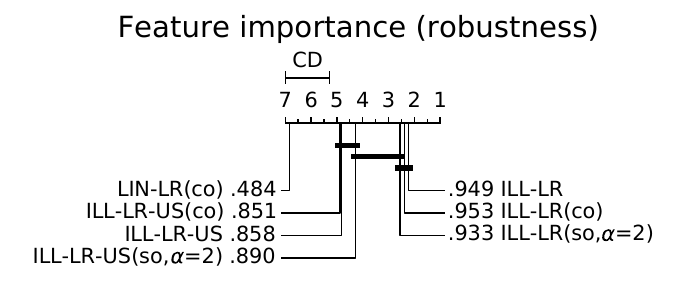}
\includegraphics[width=0.49\linewidth]{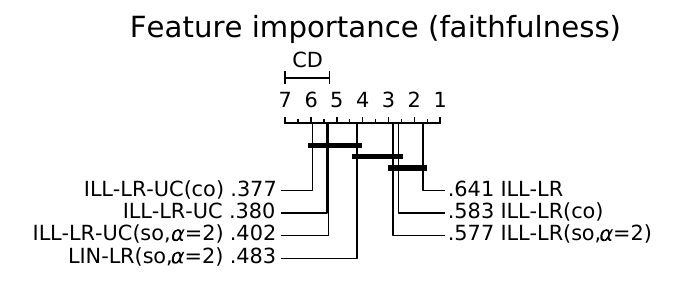}
\caption{Critical Difference plots with Nemenyi test at 90\%
confidence level for robustness and faithfulness metrics in real-world datasets, comparing various regularized versions of \approach{}-\textsc{lr}.}
\label{fig:fimp_suppl}
\end{figure*}

\begin{figure*}[h]
\includegraphics[width=0.49\linewidth]{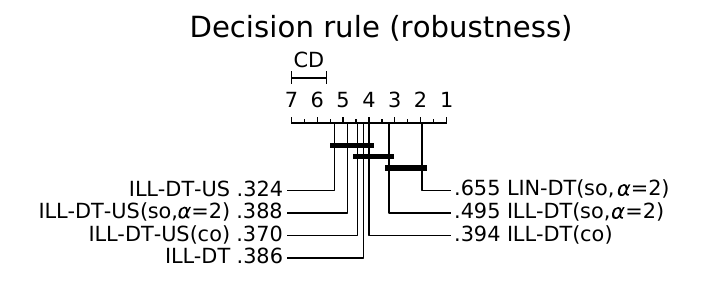}
\includegraphics[width=0.49\linewidth]{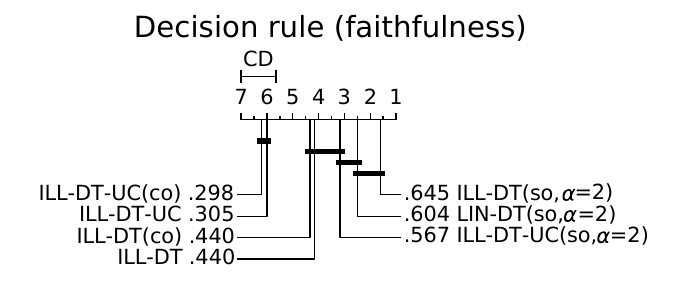}
\caption{Critical Difference plots with Nemenyi test at 90\%
confidence level for robustness and faithfulness metrics in real-world datasets, comparing various regularized versions of \approach{}-\textsc{dt}.}
\label{fig:drule_suppl}
\end{figure*}

Figure~\ref{fig:fimp_suppl} presents additional findings that compare the effects of different regularizations on the robustness and faithfulness metrics for feature importance explanations. We observe that sparse regularizations ($so$, $\alpha$=$2$) do not enhance either robustness or faithfulness for feature importance. Additionally, models regularized for non-collinearity ($co$) perform similarly to unregularized models, although they achieve the highest average robustness scores.
In contrast, Figures~\ref{fig:drule_suppl} show additional results comparing the impact of various regularizations on the robustness and faithfulness metrics for decision rules. Here, sparse regularizations ($so$, $\alpha$=$2$) significantly improve the quality of decision rules for both metrics, thereby confirming our earlier observations regarding sparsity and decision rule correctness.
Overall, for both types of explanations, models optimized for stability and label conditioning perform significantly worse in terms of robustness and faithfulness, respectively.

\begin{figure*}[h]
\centering
\includegraphics[height=0.25\linewidth]{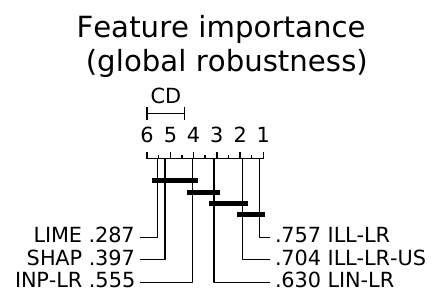}
\includegraphics[height=0.25\linewidth]{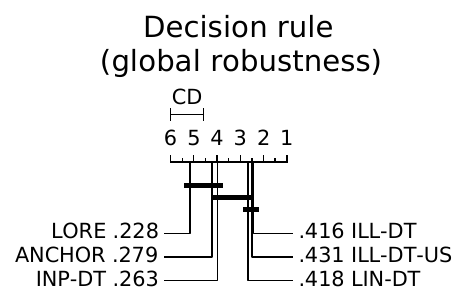}
\caption{Critical Difference plots with Nemenyi test at 90\%
confidence level for global robustness  metrics in real-world datasets.}
\label{fig:grob_suppl}
\end{figure*}

Tables~\ref{tab:global_lgbm} and Tables~\ref{tab:global_xgb} present additional results on explainers' \textbf{robustness} with the evaluation of a \textbf{global metric} rather than  local metrics 
based on sensitivity to perturbations. Intuitively, a --globally--robust explainer should produce similar explanations for closely located data points -- with the same back-box predicted label -- and distinct explanations for those farther apart.
Thus, robustness is assessed globally by calculating the Spearman's rank correlation coefficient between the pairwise distances of explanations, $\{d_{\mathcal{E}}(i, j)\}_{i<j}$, and the corresponding pairwise distances of input records, $\{\rvert \rvert\mathbf{x}_i - \mathbf{x}_j\rvert \rvert_2)\}_{i<j}$, for those pairs of points with accordant predicted labels.
Like in \textsc{seneca}, 
we use similarity metrics $\mathit{cs}\text{-}\mathit{score}(\cdot,\cdot)$ and $\mathit{cplt}\text{-}\mathit{score}(\cdot,\cdot)$ instead of distance metrics. 
Figure~\ref{fig:grob_suppl} also presents the aggregate performance of feature importance explainers. In this analysis, \approach{}-\textsc{lr} achieves the highest performance, while \approach{}-\textsc{lr-us} performs comparably. This is in contrast to the local robustness results reported in the main paper, where \approach{}-\textsc{lr-us} was significantly worse. Additionally, \textsc{lime} demonstrates the poorest performance in this context, despite showing adequate local robustness in the main paper.
Regarding decision rule explainers, \textsc{lin-lr} experiences a decline in performance compared to the local robustness results and ranks similarly to both \approach{}-\textsc{dt} and \approach{}-\textsc{dt-us}.

\begin{table}[t]
    \centering
    \setlength{\tabcolsep}{0.8mm}
    \caption{Ranking correctness of feature importance. Prediction accuracy of surrogate models inside parentheses.
    }
    \label{tab:synth1}

    \begin{tabular}{l@{\hspace{5pt}}c@{\hspace{5pt}}c@{\hspace{5pt}}c@{\hspace{5pt}}}
        &\multicolumn{3}{c}{\textbf{Feature Ranking Correctness} (\textit{spearman})}\\
        \cline{2-4}
        \textbf{LGB} & \texttt{ssin-2c} & \texttt{int2-3c} & \texttt{int2-8p} \\
        \midrule
        \textsc{ill-lr}{\scriptsize(co)} & \textbf{1.000}$\pm$.000 \scriptsize{(98.6)} &
        .184$\pm$.038 \scriptsize{(92.3)} &
        .204$\pm$.039 \scriptsize{(81.7)} \\
        \textsc{ill-lr-uc}{\scriptsize(co)} & \textbf{1.000}$\pm$.000 \scriptsize{(98.5)} &
        .185$\pm$.054 \scriptsize{(91.8)} &
        .202$\pm$.043 \scriptsize{(81.7)} \\
        \textsc{ill-lr-us}{\scriptsize(co)} & .904$\pm$.041 \scriptsize{(98.7)} &
        \textbf{.473}$\pm$.107 \scriptsize{(96.3)} &
        \textbf{.235}$\pm$.079 \scriptsize{(82.1)} \\
        \midrule
        \textsc{lin-lr}{\scriptsize(co)} & .836$\pm$.009 \scriptsize{(97.2)} &
        .033$\pm$.001 \scriptsize{(89.1)} &
        .117$\pm$.013 \scriptsize{(80.4)} \\
        \textsc{inp-lr} & .806$\pm$.006 \scriptsize{(96.6)} &
        .032$\pm$.001 \scriptsize{(89.0)} &
        .110$\pm$.012 \scriptsize{(80.1)} \\
        \textsc{lime} & \underline{.993}$\pm$.001 &
        .009$\pm$.001  &
        .082$\pm$.003  \\
        \textsc{shap} & .257$\pm$.001 &
        \underline{.197}$\pm$.010  &
        \underline{.163}$\pm$.020  \\
        \midrule
    \end{tabular}
    
    \begin{tabular}{l@{\hspace{5pt}}c@{\hspace{5pt}}c@{\hspace{5pt}}c@{\hspace{5pt}}}
        \textbf{XGB}& \texttt{ssin-2c} & \texttt{int2-3c} & \texttt{int2-8p} \\
        \midrule
        \textsc{ill-lr}{\scriptsize(co)} & \textbf{1.000}$\pm$.000 \scriptsize{(98.0)} &
        \underline{.264}$\pm$.043 \scriptsize{(91.6)} &
        .197$\pm$.019 \scriptsize{(82.7)} \\
        \textsc{ill-lr-uc}{\scriptsize(co)} & \textbf{1.000}$\pm$.000 \scriptsize{(98.3)} &
        .186$\pm$.055 \scriptsize{(91.8)} &
        \underline{.199}$\pm$.039 \scriptsize{(82.7)} \\
        \textsc{ill-lr-us}{\scriptsize(co)} & .947$\pm$.028 \scriptsize{(98.1)} &
        \textbf{.676}$\pm$.095 \scriptsize{(96.0)} &
        \textbf{.458}$\pm$.109 \scriptsize{(83.3)} \\
        \midrule
        \textsc{lin-lr}{\scriptsize(co)} & .847$\pm$.008 \scriptsize{(97.0)} &
        .036$\pm$.004 \scriptsize{(89.0)} &
        .117$\pm$.012 \scriptsize{(81.4)} \\
        \textsc{inp-lr} & .812$\pm$.009 \scriptsize{(96.1)} &
        .033$\pm$.001 \scriptsize{(88.7)} &
        .108$\pm$.013 \scriptsize{(81.2)} \\
        \textsc{lime} & \underline{.995}$\pm$.002 &
        .010$\pm$.001  &
        .081$\pm$.005  \\
        \textsc{shap} & .276$\pm$.008 &
        .211$\pm$.007  &
        .155$\pm$.018  \\
        \midrule
    \end{tabular}

\begin{tabular}{l@{\hspace{5pt}}c@{\hspace{5pt}}c@{\hspace{5pt}}c@{\hspace{5pt}}}
        \textbf{CTB}& \texttt{ssin-2c} & \texttt{int2-3c} & \texttt{int2-8p} \\
        \midrule
        \textsc{ill-lr}{\scriptsize(co)} & \textbf{1.000}$\pm$.000 \scriptsize{(98.6)} &
        .165$\pm$.048 \scriptsize{(90.6)} &
        \textbf{.278}$\pm$.021 \scriptsize{(82.7)} \\
        \textsc{ill-lr-uc}{\scriptsize(co)} & \textbf{1.000}$\pm$.000 \scriptsize{(98.4)} &
        .168$\pm$.055 \scriptsize{(91.1)} &
        \underline{.203}$\pm$.042 \scriptsize{(82.7)} \\
        \textsc{ill-lr-us}{\scriptsize(co)} & .965$\pm$.015 \scriptsize{(98.6)} &
        \textbf{.542}$\pm$.050 \scriptsize{(97.2)} &
        .147$\pm$.025 \scriptsize{(82.9)} \\
        \midrule
        \textsc{lin-lr}{\scriptsize(co)} & .824$\pm$.006 \scriptsize{(97.3)} &
        .038$\pm$.006 \scriptsize{(89.0)} &
        .116$\pm$.011 \scriptsize{(81.6)} \\
        \textsc{inp-lr} & .806$\pm$.006 \scriptsize{(96.9)} &
        .033$\pm$.002 \scriptsize{(88.9)} &
        .111$\pm$.011 \scriptsize{(81.4)} \\
        \textsc{lime} & \underline{.993}$\pm$.001 &
        .008$\pm$.001  &
        .084$\pm$.002  \\
        \textsc{shap} & .259$\pm$.012 &
        \underline{.218}$\pm$.009  &
        .122$\pm$.007  \\
        \midrule
    \end{tabular}

\end{table}

\begin{table}[h!]
    \centering
    \setlength{\tabcolsep}{0.8mm}
    \caption{Correctness of synthetic explanations. Prediction accuracy of surrogate models inside parentheses.
    }
    \label{tab:synth2}

    \begin{tabular}{l@{\hspace{5pt}}c@{\hspace{5pt}}c@{\hspace{5pt}}c@{\hspace{5pt}}}
        &\multicolumn{3}{c}{\textbf{Feature Importance Correctness} (\textit{cs-score})}\\
        \cline{2-4}
        \textbf{LGB}& \texttt{ssin-2c} & \texttt{int2-3c} & \texttt{int2-8p} \\
        \midrule
        \textsc{ill-lr}{\scriptsize(co)} & \textbf{.998}$\pm$.000 \scriptsize{(98.6)} &
        .139$\pm$.042 \scriptsize{(92.3)} &
        .210$\pm$.012 \scriptsize{(81.7)} \\
        \textsc{ill-lr-uc}{\scriptsize(co)} & \textbf{.998}$\pm$.000 \scriptsize{(98.5)} &
        .123$\pm$.050 \scriptsize{(91.8)} &
        .211$\pm$.017 \scriptsize{(81.7)} \\
        \textsc{ill-lr-us}{\scriptsize(co)} & .963$\pm$.011 \scriptsize{(98.7)} &
        \textbf{.332}$\pm$.027 \scriptsize{(96.3)} &
        \textbf{.305}$\pm$.074 \scriptsize{(82.1)} \\
        \midrule
        \textsc{lin-lr}{\scriptsize(co)} & 
        .899$\pm$.004 \scriptsize{(97.2)} &
        .050$\pm$.003 \scriptsize{(89.1)} &
        .249$\pm$.012 \scriptsize{(80.4)} \\
        \textsc{inp-lr} & 
        .891$\pm$.004 \scriptsize{(96.6)} &
        .047$\pm$.003 \scriptsize{(89.0)} &
        .248$\pm$.012 \scriptsize{(80.1)} \\
        \textsc{lime} & 
        \underline{.993}$\pm$.001 &
        .000$\pm$.000  &
        \underline{.259}$\pm$.017  \\
        \textsc{shap} & 
        .223$\pm$.011 &
        \underline{.315}$\pm$.015  &
        .142$\pm$.021  \\
        \midrule
    \end{tabular}
    
    \begin{tabular}{l@{\hspace{5pt}}c@{\hspace{5pt}}c@{\hspace{5pt}}c@{\hspace{5pt}}}
        \textbf{XGB}& \texttt{ssin-2c} & \texttt{int2-3c} & \texttt{int2-8p} \\
        \midrule
        \textsc{ill-lr}{\scriptsize(co)} & \textbf{.996}$\pm$.001 \scriptsize{(98.0)} &
        .176$\pm$.050 \scriptsize{(91.6)} &
        .219$\pm$.021 \scriptsize{(82.7)} \\
        \textsc{ill-lr-uc}{\scriptsize(co)} & \textbf{.997}$\pm$.001 \scriptsize{(98.3)} &
        .124$\pm$.054 \scriptsize{(91.8)} &
        .213$\pm$.020 \scriptsize{(82.7)} \\
        \textsc{ill-lr-us}{\scriptsize(co)} & .967$\pm$.010 \scriptsize{(98.1)} &
        \textbf{.411}$\pm$.028 \scriptsize{(96.0)} &
        \textbf{.305}$\pm$.104 \scriptsize{(83.3)} \\
        \midrule
        \textsc{lin-lr}{\scriptsize(co)} &
        .903$\pm$.005 \scriptsize{(97.0)} &
        .053$\pm$.005 \scriptsize{(89.0)} &
        .254$\pm$.014 \scriptsize{(81.4)} \\
        \textsc{inp-lr} & 
        .893$\pm$.005 \scriptsize{(96.1)} &
        .049$\pm$.002 \scriptsize{(88.7)} &
        .254$\pm$.013 \scriptsize{(81.2)} \\
        \textsc{lime} & 
        \underline{.992}$\pm$.001 &
        .000$\pm$.000  &
        \underline{.269}$\pm$.017  \\
        \textsc{shap} & 
        .236$\pm$.006 &
        \underline{.321}$\pm$.008  &
        .137$\pm$.023  \\
        \midrule
    \end{tabular}

\begin{tabular}{l@{\hspace{5pt}}c@{\hspace{5pt}}c@{\hspace{5pt}}c@{\hspace{5pt}}}

        \textbf{CTB}& \texttt{ssin-2c} & \texttt{int2-3c} & \texttt{int2-8p} \\
        \midrule
        \textsc{ill-lr}{\scriptsize(co)} & \textbf{.998}$\pm$.001 \scriptsize{(98.6)} &
        .100$\pm$.050 \scriptsize{(90.6)} &
        .221$\pm$.012 \scriptsize{(82.7)} \\
        \textsc{ill-lr-uc}{\scriptsize(co)} & \textbf{.998}$\pm$.001 \scriptsize{(98.4)} &
        .117$\pm$.052 \scriptsize{(91.1)} &
        .211$\pm$.017 \scriptsize{(82.7)} \\
        \textsc{ill-lr-us}{\scriptsize(co)} & .989$\pm$.003 \scriptsize{(98.6)} &
        \textbf{.357}$\pm$.015 \scriptsize{(97.2)} &
        .230$\pm$.058 \scriptsize{(82.9)} \\
        \midrule
        \textsc{lin-lr}{\scriptsize(co)} & 
        .899$\pm$.005 \scriptsize{(97.3)} &
        .054$\pm$.007 \scriptsize{(89.0)} &
        \textbf{.251}$\pm$.013 \scriptsize{(81.6)} \\
        \textsc{inp-lr} & 
        .891$\pm$.004 \scriptsize{(96.9)} &
        .044$\pm$.001 \scriptsize{(88.9)} &
        \textbf{.250}$\pm$.013 \scriptsize{(81.4)} \\
        \textsc{lime} & 
        \underline{.993}$\pm$.001 &
        .000$\pm$.000  &
        \underline{.244}$\pm$.012  \\
        \textsc{shap} & 
        .231$\pm$.014 &
        \underline{.308}$\pm$.004  &
        .082$\pm$.011  \\
        \midrule
    \end{tabular}

\end{table}

\subsection{Additional Experimental Results on Synthetic Data}

In this section, we further extend the evaluation of explanation methods on synthetic benchmarks beyond \textsc{seneca} framework. In line with~\cite{cortez2013using}, we adopt synthetic nonlinear functions derived from the Friedman model~\cite{friedman1991multivariate} and adapted for classification tasks. Concretely, we employ publicly available datasets\footnote{\url{http://www3.dsi.uminho.pt/pcortez/data}} \texttt{ssin-2c}, \texttt{int2-3c} and \texttt{int2-8c} respectively with two, three and eight classes. They consists of $n=1000$ synthetic instances described by 4 continuous features $\{x_1, x_2, x_3, x_4\}$.  
Instead of providing ground-truth importance values $\{\hat\psi_1, \hat\psi_2, \hat\psi_3, \hat\psi_4\}$, in~\cite{cortez2013using} the authors provides importance rankings: in \texttt{ssin-2c}, the ranking is $\hat\psi_1\!>\!\hat\psi_2\!>\!\hat\psi_3\!>\!\hat\psi_4$; in \texttt{int2-3c} and \texttt{int2-8c}, the ranking is $(\hat\psi_1, \hat\psi_2)\!>\!\hat\psi_3\!>\!\hat\psi_4$. 
Using the same experimental setting adopted for the \textsc{seneca} benchmarks, we evaluate local explanation methods with respect to ranking quality (in Table~\ref{tab:synth1} via Spearman correlation) and importance proximity (in Table~\ref{tab:synth2} via the already used \textit{cs-score}) across all test instances and five independent train–test splits in every dataset. To evaluate correctness scores, we set ground-truth importance values aligned with reference rankings as following: in \texttt{ssin-2c}, we define $\hat\psi=\{4., 2., 1., 0.\}$; in \texttt{int2-3c} and \texttt{int2-8c}, we define $\hat\psi=\{3., 3., 1., 0.\}$. 
Because these synthetic datasets do not provide white-box predictive functions like in \textsc{seneca}, we fit ensemble-tree black-boxes to mimic the underlying decision process, and then explain their local predictions. In multi-class datasets, explanations are generated for the majority class. The reported results reveal that \approach{} consistently outperforms competing explainers. On \texttt{ssin-2c}, \textsc{ill-lr} and \textsc{ill-lr-uc} achieve perfect rankings and similarities, with \textsc{lime} as the second best approach. 
For the \texttt{int2} datasets, \textsc{ill-lr} and \textsc{ill-lr-us} mostly attains the best results with XGB and LGB, consistently outperforming \textsc{lime} and tree\textsc{shap}. When CTB is explained, however, the global linear baselines \textsc{lin-lr} and \textsc{inp-lr} yield more aligned explanations on \texttt{int2-8p}. 
Overall, these findings underscore the value of carefully designing synthetic benchmarks for explanation methods evaluation.

\clearpage

\begin{table*}[h]
\centering
\caption{Decision-based latent space quality metrics with LGB as  black-box for all datasets and methods.}
\footnotesize
\begin{tabular}{lcccc|cccc|cccc}
\midrule
\textbf{LGB} & \multicolumn{4}{c}{\textbf{Decision Preservation}} & \multicolumn{4}{c}{\textbf{\textsc{lr} Accuracy} (Macro-F1)} & \multicolumn{4}{c}{\textbf{\textsc{dt} Accuracy} (Macro-F1)} \\
\cmidrule(lr){2-5} \cmidrule(lr){6-9} \cmidrule(lr){10-13}
 & \textsc{ill} & \textsc{lin} & \textsc{ill-uc} & \textsc{lin-uc} & \textsc{ill} & \textsc{lin} & \textsc{ill-uc} & \textsc{lin-uc} & \textsc{ill} & \textsc{lin} & \textsc{ill-uc} & \textsc{lin-uc} \\
\midrule
\texttt{aids}        &  \textbf{.618} &  \underline{.596} & .549  & .555 &  \textbf{.952} &  \underline{.902} & .899        &  \underline{.902} &  \textbf{.930} &  \underline{.874} & .768 & .780 \\
\texttt{austr}  &  \textbf{.743} &  \underline{.729} & .613  & .623 & .920          &  \textbf{.927}      & .919        &  \underline{.920} &  \underline{.927} &  \textbf{.942} & .869 & .913 \\
\texttt{bank}        &  \textbf{.659} &  \underline{.656} & .595  & .588 &  \textbf{.901} &  \underline{.866} & .839        & .840               &  \textbf{.881} &  \textbf{.881} &  \underline{.759} & .736 \\
\texttt{breast}      &  \textbf{.779} &  \underline{.739} & .728  & .718 & .981          &  \underline{.990}               & .981        &  \textbf{.991}      & .972 & .963 &  \underline{.981} &  \textbf{.991} \\
\texttt{churn}       &  \textbf{.610} &  \underline{.583} & .542  & .548 &  \textbf{.875} & .637               &  \underline{.650} & .623          &  \textbf{.852} &  \underline{.827} & .738 & .760 \\
\texttt{compas}      &  \textbf{.667} &  \underline{.665} & .644  & .640 &  \textbf{.926} &  \underline{.914} & .902        &  \underline{.914} &  \textbf{.913} &  \underline{.893} & .891 & .891 \\
\texttt{ctg}         &  \textbf{.693} &  \underline{.689} & .624  & .611 &  \underline{.993}          &  \textbf{.996}      &  \textbf{.996} &  \underline{.993}          &  \underline{.993} &  \textbf{.996} & .969 & .949 \\
\texttt{diabetes}    &  \textbf{.642} &  \underline{.640} & .621  & .627 &  \textbf{.906} & .871               &  \underline{.881} & .871          &  \textbf{.874} & .832 & .836 &  \underline{.864} \\
\texttt{ecoli}       &  \textbf{.726} & .700              &  \underline{.706} &  \underline{.706} &  \textbf{.970} & .936               &  \underline{.947} & .936          &  \textbf{.939} &  \textbf{.939} & .910 &  \underline{.921} \\
\texttt{fico}        &  \textbf{.642} & .578              &  \underline{.586} & .580 &  \textbf{.908} &  \underline{.903} & .894        &  \underline{.903} & .862 &  \textbf{.869} & .843 & .854 \\
\texttt{german}      &  \textbf{.669} &  \underline{.627} & .566  & .557 &  \underline{.814}          & .807               &  \textbf{.816} & .806          &  \textbf{.805} &  \underline{.791} & .699 & .707 \\
\texttt{home}        &  \textbf{.677} &  \underline{.626} & .614  & .614 &  \textbf{.949} &  \textbf{.949}      &  \textbf{.949} &  \textbf{.949}    &  \textbf{.970} & .939 & .929 &  \underline{.949} \\
\texttt{ionos}  &  \underline{.631} &  \textbf{.665} & .586  & .597 & .886          &  \underline{.906} &  \textbf{.934} & .854          & .919 &  \underline{.952} &  \textbf{.968} & .907 \\
\texttt{sonar}       &  \textbf{.671} &  \underline{.629} & .576  & .610 &  \textbf{.904} &  \underline{.857} & .810        & .833               &  \textbf{.881} &  \underline{.833} & .785 & .786 \\
\texttt{spam}        &  \textbf{.704} & .647              &  \underline{.674} & .623 &  \textbf{.963} &  \underline{.942} &  \underline{.942} & .936          &  \textbf{.955} &  \underline{.946} & .913 & .903 \\
\texttt{titanic}     &  \underline{.743} &  \textbf{.744} & .650  & .672 &  \underline{.885}          &  \textbf{.893}      &  \textbf{.893} &  \underline{.885}          &  \textbf{.937} & .919 & .922 &  \underline{.931} \\
\texttt{wine}        &  \textbf{.550} &  \underline{.546} & .536  & .540 &  \textbf{.305} & .266               &  \underline{.292} & .286          &  \underline{.516} &  \textbf{.537} & .497 & .440 \\
\texttt{yeast}       &  \textbf{.630} & .615              & .609  &  \underline{.623} &  \textbf{.629} & .565               &  \underline{.612} & .572          &  \textbf{.556} &  \underline{.534} & .525 & .512 \\
\midrule
\end{tabular}
\label{tab:quality_lgbm}
\end{table*}

\begin{table*}[h]
\centering
\caption{Decision-based latent space quality metrics with XGB as  black-box for all datasets and methods.}
\footnotesize
\begin{tabular}{lcccc|cccc|cccc}
\midrule
\textbf{XGB} & \multicolumn{4}{c}{\textbf{Decision Preservation}} & \multicolumn{4}{c}{\textbf{\textsc{lr} Accuracy} (Macro-F1)} & \multicolumn{4}{c}{\textbf{\textsc{dt} Accuracy} (Macro-F1)} \\
\cmidrule(lr){2-5} \cmidrule(lr){6-9} \cmidrule(lr){10-13}
 & \textsc{ill} & \textsc{lin} & \textsc{ill-uc} & \textsc{lin-uc} & \textsc{ill} & \textsc{lin} & \textsc{ill-uc} & \textsc{lin-uc} & \textsc{ill} & \textsc{lin} & \textsc{ill-uc} & \textsc{lin-uc} \\
\midrule
\texttt{aids} &  \textbf{.661} &  \underline{.610} & .563 & .557 &  \textbf{.941} & .871 &  \underline{.880} & .868 &  \textbf{.938} &  \underline{.877} & .730 & .774 \\
\texttt{austr} &  \underline{.739} &  \textbf{.751} & .657 & .652 &  \textbf{.934} &  \underline{.913} & .912 & .912 &  \textbf{.941} &  \underline{.911} & .898 & .898 \\
\texttt{bank} &  \underline{.653} &  \textbf{.662} & .593 & .582 &  \textbf{.893} &  \underline{.868} & .832 & .853 &  \textbf{.886} &  \underline{.873} & .750 & .744 \\
\texttt{breast} &  \textbf{.821} & .737 &  \underline{.775} & .730 &  \underline{.981} &  \textbf{.990} &  \underline{.981} &  \textbf{.990} &  \underline{.972} &  \underline{.972} & .961 & \textbf{1.000} \\
\texttt{churn} &  \textbf{.602} &  \underline{.587} & .536 & .541 &  \textbf{.872} &  \underline{.660} & .649 & .644 &  \textbf{.852} &  \underline{.786} & .728 & .728 \\
\texttt{compas} &  \underline{.671} &  \textbf{.673} & .640 & .638 &  \textbf{.933} &  \underline{.915} & .911 &  \underline{.915} &  \textbf{.915} &  \underline{.912} & .891 & .888 \\
\texttt{ctg} &  \textbf{.743} &  \underline{.739} & .653 & .629 & \textbf{1.000} &  \underline{.996} &  \underline{.996} &  \underline{.996} &  \underline{.957} & .945 & \textbf{1.000} & .945 \\
\texttt{diabetes} &  \textbf{.684} &  \underline{.649} & .625 & .626 &  \textbf{.910} & .871 &  \underline{.879} & .871 &  \underline{.857} &  \textbf{.895} & .849 &  \underline{.857} \\
\texttt{ecoli} &  \underline{.721} & .703 &  \textbf{.726} & .709 & \textbf{1.000} & .969 &  \underline{.988} & .980 &  \underline{.980} &  \textbf{.992} & .947 &  \underline{.980} \\
\texttt{fico} &  \textbf{.650} & .588 &  \underline{.592} & .588 &  \textbf{.937} &  \underline{.929} & .922 &  \underline{.929} &  \underline{.886} &  \textbf{.897} & .882 & .874 \\
\texttt{german} &  \textbf{.655} &  \underline{.617} & .535 & .516 &  \underline{.789} &  \textbf{.810} & .761 & .778 &  \underline{.780} &  \textbf{.791} & .714 & .697 \\
\texttt{home} &  \textbf{.663} &  \underline{.642} & .630 & .626 &  \underline{.939} &  \textbf{.949} &  \underline{.939} &  \underline{.939} & .929 &  \textbf{.970} &  \underline{.939} & .929 \\
\texttt{ionos} & .566 &  \textbf{.603} &  \underline{.575} & .572 &  \underline{.891} & .868 &  \textbf{.904} & .883 & .874 &  \textbf{.954} &  \underline{.936} & .880 \\
\texttt{sonar} &  \underline{.619} &  \textbf{.629} & .595 &  \textbf{.629} & .786 & .785 &  \underline{.810} &  \textbf{.833} &  \textbf{.857} & .762 &  \underline{.786} & .762 \\
\texttt{spam} &  \textbf{.721} & .667 &  \underline{.675} & .631 &  \textbf{.968} &  \underline{.947} & .942 & .940 &  \textbf{.966} &  \underline{.952} & .919 & .901 \\
\texttt{titanic} &  \textbf{.725} & .687 & .655 &  \underline{.694} & .877 &  \textbf{.890} &  \underline{.883} & .877 &  \textbf{.923} &  \underline{.921} &  \underline{.921} & .910 \\
\texttt{wine} &  \textbf{.587} &  \underline{.574} & .562 & .566 &  \textbf{.334} & .276 &  \underline{.319} & .287 & .422 &  \textbf{.544} & .441 &  \underline{.494} \\
\texttt{yeast} &  \textbf{.700} &  \underline{.695} & .690 & .694 &  \textbf{.790} & .755 &  \underline{.776} & .752 &  \underline{.676} &  \textbf{.698} & .644 & .656 \\
\midrule
\end{tabular}
\label{tab:quality_xgb}
\end{table*}

\begin{table*}[h]
    \centering
    \caption{Robustness and faithfulness metrics for feature importance methods in all datasets with LGB as black-box. Prediction accuracy of surrogate classifiers is reported inside parentheses.}
    \footnotesize
\makebox[\linewidth]{
    \begin{tabular}{lcccccc|cccccc}
        \midrule
        \textbf{LGB}&\multicolumn{6}{c}{\textbf{Feature Importance Robustness}}&\multicolumn{6}{c}{\textbf{Feature Importance Faithfulness}}\\
        \cmidrule(lr){2-7}  \cmidrule(lr){8-13}
        & \textsc{ill}-\textsc{lr} & \textsc{ill}-\textsc{lr-us} & \textsc{lin-lr} & \textsc{inp-lr} & \textsc{lime} & \textsc{shap} & \textsc{ill}-\textsc{lr} &
        \textsc{ill}-\textsc{lr-uc} &
        \textsc{lin-lr} & \textsc{inp-lr} & \textsc{lime} & \textsc{shap} \\
        \midrule
        \texttt{aids} &  \textbf{.994} (\scriptsize{95.3}) & .797 (\scriptsize{97.0}) & .582 (\scriptsize{93.7}) & .244 (\scriptsize{93.7}) &  \underline{.978} & .445 &  \textbf{.908} (\scriptsize{97.4}) & .234 (\scriptsize{93.5}) & .540 (\scriptsize{93.5}) & .610 (\scriptsize{93.7}) & .054 &  \underline{.806} \\
\texttt{austr} &  \textbf{.980} (\scriptsize{93.5}) & .901 (\scriptsize{89.9}) & .659 (\scriptsize{92.0}) & .500 (\scriptsize{92.0}) &  \underline{.904} & .443 &  \underline{.785} (\scriptsize{92.0}) & .520 (\scriptsize{92.0}) & .561 (\scriptsize{89.9}) & .659 (\scriptsize{92.0}) & .033 &  \textbf{.798} \\
\texttt{bank} &  \textbf{.994} (\scriptsize{96.1}) &  \underline{.958} (\scriptsize{96.0}) & .536 (\scriptsize{94.9}) & .271 (\scriptsize{95.0}) & .916 & .127 &  \textbf{.790} (\scriptsize{96.8}) & .304 (\scriptsize{95.4}) & .537 (\scriptsize{95.0}) &  \underline{.540} (\scriptsize{95.0}) & .122 & .410 \\
\texttt{breast} & .777 (\scriptsize{98.2}) &  \underline{.811} (\scriptsize{97.4}) & .388 (\scriptsize{99.1}) & .219 (\scriptsize{98.2}) &  \textbf{.964} & .328 & .846 (\scriptsize{99.1}) &  \underline{.859} (\scriptsize{98.2}) & .819 (\scriptsize{97.4}) & .833 (\scriptsize{98.2}) & .113 &  \textbf{.906} \\
\texttt{churn} &  \textbf{.981} (\scriptsize{95.8}) &  \underline{.932} (\scriptsize{94.8}) & .719 (\scriptsize{87.3}) & .672 (\scriptsize{87.6}) & .618 & .160 &  \textbf{.679} (\scriptsize{96.6}) & .142 (\scriptsize{89.1}) & .055 (\scriptsize{87.4}) & .247 (\scriptsize{87.6}) & .091 &  \underline{.644} \\
\texttt{compas} &  \textbf{.993} (\scriptsize{94.7}) & .711 (\scriptsize{95.1}) & .711 (\scriptsize{94.4}) & .635 (\scriptsize{94.4}) &  \underline{.971} & .538 &  \textbf{.798} (\scriptsize{95.8}) & .317 (\scriptsize{94.0}) & .319 (\scriptsize{94.4}) & .313 (\scriptsize{94.4}) & .015 &  \underline{.448} \\
\texttt{ctg} &  \textbf{.980} (\scriptsize{99.8}) &  \underline{.840} (\scriptsize{99.8}) & .758 (\scriptsize{99.5}) & .617 (\scriptsize{99.3}) & .781 & .656 &  \textbf{.741} (\scriptsize{99.8}) & .357 (\scriptsize{99.5}) & .624 (\scriptsize{99.5}) & .519 (\scriptsize{99.3}) & .487 &  \underline{.682} \\
\texttt{diabetes} &  \underline{.966} (\scriptsize{92.9}) & .797 (\scriptsize{94.2}) & .180 (\scriptsize{89.6}) & .176 (\scriptsize{89.6}) &  \textbf{.985} & .163 & .396 (\scriptsize{92.9}) & .298 (\scriptsize{90.9}) &  \underline{.454} (\scriptsize{89.6}) & .450 (\scriptsize{89.6}) & .021 &  \textbf{.580} \\
\texttt{ecoli} & .958 (\scriptsize{100.}) &  \underline{.961} (\scriptsize{100.}) & .735 (\scriptsize{100.}) & .673 (\scriptsize{100.}) &  \textbf{.981} & .585 &  \textbf{.640} (\scriptsize{100.}) & .623 (\scriptsize{100.}) & .575 (\scriptsize{100.}) & .586 (\scriptsize{100.}) & .260 &  \underline{.503} \\
\texttt{fico} &  \textbf{.981} (\scriptsize{90.3}) & .899 (\scriptsize{90.8}) & .400 (\scriptsize{90.3}) & .324 (\scriptsize{90.3}) &  \underline{.918} & .268 &  \underline{.443} (\scriptsize{91.4}) & .338 (\scriptsize{89.5}) & .379 (\scriptsize{90.3}) & .377 (\scriptsize{90.3}) & .000 &  \textbf{.600} \\
\texttt{german} &  \textbf{.996} (\scriptsize{89.0}) &  \underline{.930} (\scriptsize{85.5}) & .394 (\scriptsize{85.5}) & .338 (\scriptsize{88.5}) & .840 & .069 &  \textbf{.531} (\scriptsize{87.5}) & .198 (\scriptsize{88.0}) & .174 (\scriptsize{85.5}) & .190 (\scriptsize{88.5}) & .124 &  \underline{.238} \\
\texttt{home} &  \underline{.878} (\scriptsize{96.0}) & .756 (\scriptsize{97.0}) & .204 (\scriptsize{94.9}) & .122 (\scriptsize{94.9}) &  \textbf{.976} & .178 &  \underline{.741} (\scriptsize{97.0}) & .337 (\scriptsize{94.9}) & .495 (\scriptsize{94.9}) & .486 (\scriptsize{94.9}) & .077 &  \textbf{.752} \\
\texttt{ionos} &  \underline{.845} (\scriptsize{94.4}) & .688 (\scriptsize{94.4}) & .371 (\scriptsize{88.7}) & .353 (\scriptsize{88.7}) &  \textbf{.932} & .366 & .385 (\scriptsize{93.0}) & .153 (\scriptsize{94.4}) & .568 (\scriptsize{91.5}) &  \underline{.576} (\scriptsize{88.7}) & .193 &  \textbf{.841} \\
\texttt{sonar} & .621 (\scriptsize{85.7}) &  \underline{.698} (\scriptsize{92.9}) & .104 (\scriptsize{83.3}) & .085 (\scriptsize{83.3}) &  \textbf{.914} & .078 &  \underline{.211} (\scriptsize{90.5}) & .055 (\scriptsize{88.1}) & .201 (\scriptsize{88.1}) & .077 (\scriptsize{83.3}) & .032 &  \textbf{.469} \\
\texttt{spam} &  \textbf{.923} (\scriptsize{96.0}) &  \underline{.859} (\scriptsize{96.2}) & .396 (\scriptsize{94.0}) & .311 (\scriptsize{94.4}) & .808 & .166 &  \textbf{.848} (\scriptsize{96.4}) & .441 (\scriptsize{94.5}) & .522 (\scriptsize{94.6}) & .037 (\scriptsize{94.4}) & .000 &  \underline{.662} \\
\texttt{titanic} &  \textbf{.991} (\scriptsize{90.5}) &  \underline{.919} (\scriptsize{90.5}) & .616 (\scriptsize{89.9}) & .646 (\scriptsize{89.4}) & .971 & .536 &  \textbf{.769} (\scriptsize{91.6}) & .640 (\scriptsize{90.5}) & .639 (\scriptsize{89.9}) & .641 (\scriptsize{89.4}) & .389 &  \underline{.733} \\
\texttt{wine} &  \textbf{.943} (\scriptsize{64.2}) &  \underline{.640} (\scriptsize{67.2}) & .192 (\scriptsize{59.1}) & .176 (\scriptsize{58.6}) & .779 & .166 &  \textbf{.152} (\scriptsize{72.3}) & .105 (\scriptsize{62.8}) & .135 (\scriptsize{58.8}) &  \underline{.146} (\scriptsize{58.6}) & .022 & .104 \\
\texttt{yeast} &  \textbf{.934} (\scriptsize{76.4}) &  \underline{.705} (\scriptsize{75.1}) & .323 (\scriptsize{69.4}) & .305 (\scriptsize{69.4}) & .289 & .128 &  \textbf{.280} (\scriptsize{78.5}) & .143 (\scriptsize{74.1}) & .216 (\scriptsize{70.7}) &  \underline{.217} (\scriptsize{69.4}) & .018 & .177 \\
        
    \midrule
    \end{tabular}
    }
\label{tab:fimp_lgbm}
\end{table*}

\begin{table*}[h]
    \centering
    \caption{Robustness and faithfulness metrics for feature importance methods in all datasets with XGB as black-box. Prediction accuracy of surrogate classifiers is reported inside parentheses.}
    \footnotesize
\makebox[\linewidth]{
    \begin{tabular}{lcccccc|cccccc}
        \midrule
        \textbf{XGB}&\multicolumn{6}{c}{\textbf{Feature Importance Robustness}}&\multicolumn{6}{c}{\textbf{Feature Importance Faithfulness}}\\
        \cmidrule(lr){2-7}  \cmidrule(lr){8-13}
        & \textsc{ill}-\textsc{lr} & \textsc{ill}-\textsc{lr-us} & \textsc{lin-lr} & \textsc{inp-lr} & \textsc{lime} & \textsc{shap} & \textsc{ill}-\textsc{lr} &
        \textsc{ill}-\textsc{lr-uc} &
        \textsc{lin-lr} & \textsc{inp-lr} & \textsc{lime} & \textsc{shap} \\
        \midrule

\texttt{aids} &  \textbf{.994} \scriptsize{(95.3)} & .901 \scriptsize{(95.1)} & .505 \scriptsize{(91.1)} & .328 \scriptsize{(91.1)} &  \underline{.965} & .393 &  \textbf{.838} \scriptsize{(95.8)} & .252 \scriptsize{(91.8)} & .654 \scriptsize{(91.1)} & .574 \scriptsize{(91.1)} & .083 &  \underline{.799} \\
\texttt{austr} &  \textbf{.987} \scriptsize{(92.0)} & .872 \scriptsize{(92.8)} & .614 \scriptsize{(90.6)} & .514 \scriptsize{(90.6)} &  \underline{.929} & .341 &  \textbf{.783} \scriptsize{(94.2)} & .560 \scriptsize{(91.3)} & .704 \scriptsize{(90.6)} & .698 \scriptsize{(90.6)} & .000 &  \underline{.737} \\
\texttt{bank} &  \textbf{.989} \scriptsize{(96.2)} &  \underline{.929} \scriptsize{(96.0)} & .503 \scriptsize{(95.3)} & .260 \scriptsize{(95.5)} & .910 & .213 &  \textbf{.826} \scriptsize{(97.1)} & .397 \scriptsize{(94.9)} & .352 \scriptsize{(93.8)} & .473 \scriptsize{(95.5)} & .269 &  \underline{.765} \\
\texttt{breast} &  \underline{.875} \scriptsize{(99.1)} & .803 \scriptsize{(99.1)} & .451 \scriptsize{(100.)} & .502 \scriptsize{(97.4)} &  \textbf{.970} & .302 &  \underline{.888} \scriptsize{(98.2)} & .883 \scriptsize{(98.2)} & .850 \scriptsize{(98.2)} & .864 \scriptsize{(97.4)} & .128 &  \textbf{.927} \\
\texttt{churn} &  \textbf{.974} \scriptsize{(95.1)} &  \underline{.840} \scriptsize{(95.1)} & .776 \scriptsize{(88.0)} & .680 \scriptsize{(87.9)} & .549 & .163 &  \textbf{.612} \scriptsize{(95.8)} & .128 \scriptsize{(89.1)} & .154 \scriptsize{(88.5)} & .197 \scriptsize{(87.9)} & .078 &  \underline{.611} \\
\texttt{compas} &  \textbf{.989} \scriptsize{(95.4)} & .742 \scriptsize{(95.2)} & .713 \scriptsize{(94.5)} & .597 \scriptsize{(94.5)} &  \underline{.971} & .532 &  \textbf{.731} \scriptsize{(96.1)} & .320 \scriptsize{(94.4)} & .330 \scriptsize{(94.5)} & .321 \scriptsize{(94.5)} & .004 &  \underline{.525} \\
\texttt{ctg} &  \textbf{.992} \scriptsize{(100.)} &  \underline{.859} \scriptsize{(99.8)} & .670 \scriptsize{(99.8)} & .477 \scriptsize{(99.8)} & .824 & .502 &  \textbf{.776} \scriptsize{(100.)} & .323 \scriptsize{(99.8)} & .688 \scriptsize{(99.8)} & .434 \scriptsize{(99.8)} & .066 &  \underline{.696} \\
\texttt{diabetes} &  \underline{.971} \scriptsize{(92.2)} & .780 \scriptsize{(92.9)} & .225 \scriptsize{(90.3)} & .208 \scriptsize{(87.0)} &  \textbf{.984} & .171 & .390 \scriptsize{(92.9)} & .273 \scriptsize{(90.3)} &  \underline{.517} \scriptsize{(89.0)} & .507 \scriptsize{(87.0)} & .048 &  \textbf{.671} \\
\texttt{ecoli} &  \underline{.973} \scriptsize{(100.)} & .945 \scriptsize{(100.)} & .722 \scriptsize{(100.)} & .659 \scriptsize{(98.5)} &  \textbf{.990} & .726 &  \textbf{.720} \scriptsize{(100.)} & .711 \scriptsize{(100.)} & .683 \scriptsize{(100.)} & .691 \scriptsize{(98.5)} & .358 &  \underline{.516} \\
\texttt{fico} &  \textbf{.973} \scriptsize{(93.4)} & .944 \scriptsize{(93.2)} & .433 \scriptsize{(92.9)} & .368 \scriptsize{(92.9)} &  \underline{.971} & .304 &  \underline{.516} \scriptsize{(95.3)} & .409 \scriptsize{(92.3)} & .453 \scriptsize{(92.9)} & .450 \scriptsize{(92.9)} & .000 &  \textbf{.672} \\
\texttt{german} &  \textbf{.996} \scriptsize{(87.0)} &  \underline{.953} \scriptsize{(87.0)} & .512 \scriptsize{(86.0)} & .320 \scriptsize{(86.0)} & .782 & .071 &  \textbf{.492} \scriptsize{(89.0)} & .131 \scriptsize{(86.5)} & .167 \scriptsize{(82.5)} & .200 \scriptsize{(86.0)} & .005 &  \underline{.358} \\
\texttt{home} &  \underline{.956} \scriptsize{(93.9)} & .912 \scriptsize{(96.0)} & .162 \scriptsize{(93.9)} & .141 \scriptsize{(93.9)} &  \textbf{.974} & .148 &  \underline{.636} \scriptsize{(97.0)} & .277 \scriptsize{(94.9)} & .474 \scriptsize{(93.9)} & .470 \scriptsize{(93.9)} & .027 &  \textbf{.712} \\
\texttt{ionos} &  \underline{.772} \scriptsize{(91.5)} & .734 \scriptsize{(91.5)} & .382 \scriptsize{(87.3)} & .351 \scriptsize{(90.1)} &  \textbf{.950} & .322 & .244 \scriptsize{(97.2)} & .189 \scriptsize{(91.5)} & .533 \scriptsize{(88.7)} &  \underline{.551} \scriptsize{(90.1)} & .162 &  \textbf{.914} \\
\texttt{sonar} &  \underline{.784} \scriptsize{(81.0)} & .686 \scriptsize{(81.0)} & .090 \scriptsize{(76.2)} & .069 \scriptsize{(78.6)} &  \textbf{.935} & .168 &  \underline{.183} \scriptsize{(83.3)} & .137 \scriptsize{(83.3)} & .124 \scriptsize{(81.0)} & .084 \scriptsize{(78.6)} & .042 &  \textbf{.633} \\
\texttt{spam} &  \textbf{.912} \scriptsize{(96.5)} & .819 \scriptsize{(96.9)} & .291 \scriptsize{(94.5)} & .141 \scriptsize{(94.6)} &  \underline{.878} & .140 &  \textbf{.823} \scriptsize{(97.0)} & .462 \scriptsize{(94.5)} & .421 \scriptsize{(94.6)} & .265 \scriptsize{(94.6)} & .013 &  \underline{.721} \\
\texttt{titanic} &  \textbf{.993} \scriptsize{(91.1)} & .921 \scriptsize{(91.1)} & .662 \scriptsize{(89.4)} & .608 \scriptsize{(89.9)} &  \underline{.973} & .633 & .571 \scriptsize{(90.5)} &  \underline{.671} \scriptsize{(89.9)} & .638 \scriptsize{(88.8)} & .634 \scriptsize{(89.9)} & .155 &  \textbf{.809} \\
\texttt{wine} &  \textbf{.952} \scriptsize{(66.3)} & .581 \scriptsize{(66.6)} & .183 \scriptsize{(60.4)} & .174 \scriptsize{(59.3)} &  \underline{.856} & .278 &  \textbf{.223} \scriptsize{(72.9)} & .165 \scriptsize{(64.3)} & .206 \scriptsize{(60.1)} &  \underline{.209} \scriptsize{(59.3)} & .042 & .107 \\
\texttt{yeast} &  \textbf{.964} \scriptsize{(86.5)} & .795 \scriptsize{(88.2)} & .391 \scriptsize{(81.5)} & .394 \scriptsize{(81.8)} &  \underline{.941} & .813 &  \textbf{.528} \scriptsize{(88.2)} & .353 \scriptsize{(84.2)} & .325 \scriptsize{(81.5)} & .311 \scriptsize{(81.8)} & .152 &  \underline{.290} \\

        \midrule
    \end{tabular}
}
\label{tab:fimp_xgb}
\end{table*}

\begin{table*}[h]
    \centering
    \caption{Robustness and faithfulness metrics for decision rules methods in all datasets with LGB as black-box. Prediction accuracy of surrogate classifiers is reported inside parentheses.}
    \footnotesize
\makebox[\linewidth]{
    \begin{tabular}{lcccccc|cccccc}
        \midrule
        \textbf{LGB}&\multicolumn{6}{c}{\textbf{Decision Rule Robustness}}&\multicolumn{6}{c}{\textbf{Decision Rule Faithfulness}}\\
        \cmidrule(lr){2-7}  \cmidrule(lr){8-13}
        & \textsc{ill}-\textsc{dt} & \textsc{ill}-\textsc{dt-us} & \textsc{lin-dt} & \textsc{inp-dt} & \textsc{lore} & \textsc{anchor} & \textsc{ill}-\textsc{dt} & \textsc{ill}-\textsc{dt-uc} & \textsc{lin-dt} & \textsc{inp-dt} & \textsc{lore} & \textsc{anchor} \\
        \midrule
        \texttt{aids} &  \underline{.676} (\scriptsize 97.4) & .301 (\scriptsize 96.7) &  \textbf{.911} (\scriptsize 97.9) & .486 (\scriptsize 98.8) & .060 & .180 & .739 (\scriptsize 97.4) & .644 (\scriptsize 96.5) &  \underline{.825} (\scriptsize 97.9) & .804 (\scriptsize 98.8) & .762 &  \textbf{.868} \\
\texttt{austr} & .404 (\scriptsize 92.0) &  \underline{.438} (\scriptsize 92.0) &  \textbf{.573} (\scriptsize 97.1) & .288 (\scriptsize 94.9) & .147 & .218 &  \textbf{.707} (\scriptsize 92.0) & .653 (\scriptsize 94.9) &  \underline{.694} (\scriptsize 97.1) & .661 (\scriptsize 94.9) & .666 & .672 \\
\texttt{bank} & .595 (\scriptsize 95.5) & .504 (\scriptsize 95.5) &  \textbf{.679} (\scriptsize 95.3) &  \underline{.600} (\scriptsize 96.1) & .038 & .053 & .642 (\scriptsize 95.5) & .566 (\scriptsize 93.8) &  \textbf{.734} (\scriptsize 95.3) &  \underline{.675} (\scriptsize 96.1) & .510 & .457 \\
\texttt{breast} & .352 (\scriptsize 98.2) & .114 (\scriptsize 96.5) &  \underline{.452} (\scriptsize 98.2) &  \textbf{.536} (\scriptsize 96.5) & .077 & .025 &  \textbf{.863} (\scriptsize 98.2) & .788 (\scriptsize 98.2) &  \underline{.817} (\scriptsize 96.5) & .736 (\scriptsize 96.5) & .735 & .381 \\
\texttt{churn} &  \underline{.618} (\scriptsize 95.7) & .471 (\scriptsize 95.2) &  \textbf{.890} (\scriptsize 97.3) & .358 (\scriptsize 98.2) & .215 & .099 &  \textbf{.609} (\scriptsize 95.7) & .399 (\scriptsize 91.6) & .560 (\scriptsize 97.3) &  \underline{.605} (\scriptsize 98.2) & .605 & .413 \\
\texttt{compas} &  \underline{.603} (\scriptsize 95.1) & .351 (\scriptsize 94.3) &  \textbf{.961} (\scriptsize 95.1) & .361 (\scriptsize 95.8) & .115 & .168 & .584 (\scriptsize 95.1) & .564 (\scriptsize 94.5) &  \textbf{.619} (\scriptsize 95.1) & .536 (\scriptsize 95.8) & .503 &  \underline{.592} \\
\texttt{ctg} & .791 (\scriptsize 99.3) & .677 (\scriptsize 99.3) &  \textbf{.975} (\scriptsize 99.3) &  \underline{.866} (\scriptsize 99.3) & .317 & .031 & .777 (\scriptsize 99.3) &  \underline{.779} (\scriptsize 99.1) & .743 (\scriptsize 99.3) &  \textbf{.839} (\scriptsize 99.3) & .666 & .335 \\
\texttt{diabetes} &  \underline{.116} (\scriptsize 89.6) & .057 (\scriptsize 87.7) &  \textbf{.166} (\scriptsize 89.6) & .098 (\scriptsize 88.3) & .047 & .097 & .449 (\scriptsize 89.6) &  \underline{.545} (\scriptsize 90.9) & .344 (\scriptsize 89.6) & .295 (\scriptsize 88.3) & .276 &  \textbf{.646} \\
\texttt{ecoli} &  \underline{.632} (\scriptsize 100.) & .569 (\scriptsize 100.) &  \textbf{.784} (\scriptsize 100.) & .529 (\scriptsize 100.) & .478 & .462 & .570 (\scriptsize 100.) &  \underline{.587} (\scriptsize 100.) & .531 (\scriptsize 100.) & .513 (\scriptsize 100.) & .541 &  \textbf{.600} \\
\texttt{fico} &  \underline{.374} (\scriptsize 90.1) & .262 (\scriptsize 89.5) &  \textbf{.390} (\scriptsize 89.5) & .200 (\scriptsize 90.5) & .038 & .167 &  \underline{.512} (\scriptsize 90.1) & .425 (\scriptsize 89.4) &  \underline{.512} (\scriptsize 89.5) & .481 (\scriptsize 90.5) & .457 &  \textbf{.637} \\
\texttt{german} &  \textbf{.618} (\scriptsize 86.0) & .287 (\scriptsize 85.5) &  \underline{.562} (\scriptsize 83.0) & .154 (\scriptsize 81.0) & .058 & .085 &  \underline{.327} (\scriptsize 86.0) & .142 (\scriptsize 87.0) & .112 (\scriptsize 83.0) & .141 (\scriptsize 81.0) & .118 &  \textbf{.369} \\
\texttt{home} & .064 (\scriptsize 96.0) & .058 (\scriptsize 96.0) & .051 (\scriptsize 98.0) &  \textbf{.150} (\scriptsize 94.9) & .014 &  \underline{.073} &  \underline{.511} (\scriptsize 96.0) & .420 (\scriptsize 97.0) & .469 (\scriptsize 98.0) & .485 (\scriptsize 94.9) & .407 &  \textbf{.680} \\
\texttt{ionos} &  \underline{.438} (\scriptsize 98.6) & .297 (\scriptsize 97.2) &  \textbf{.570} (\scriptsize 95.8) & .376 (\scriptsize 95.8) & .204 & .155 &  \textbf{.793} (\scriptsize 98.6) & .722 (\scriptsize 97.2) &  \underline{.751} (\scriptsize 95.8) & .697 (\scriptsize 95.8) & .748 & .645 \\
\texttt{sonar} & .194 (\scriptsize 83.3) & .147 (\scriptsize 85.7) &  \textbf{.515} (\scriptsize 81.0) &  \underline{.398} (\scriptsize 83.3) & .108 & .043 &  \textbf{.458} (\scriptsize 85.7) & .444 (\scriptsize 85.7) & .422 (\scriptsize 81.0) &  \underline{.453} (\scriptsize 83.3) & .214 & .441 \\
\texttt{spam} & .402 (\scriptsize 96.1) &  \underline{.459} (\scriptsize 95.1) &  \textbf{.652} (\scriptsize 94.5) & .249 (\scriptsize 93.7) & .120 & .113 &  \textbf{.767} (\scriptsize 95.1) & .622 (\scriptsize 92.8) &  \underline{.701} (\scriptsize 94.5) & .642 (\scriptsize 93.7) & .253 & .495 \\
\texttt{titanic} & .628 (\scriptsize 93.9) & .476 (\scriptsize 95.0) &  \textbf{.948} (\scriptsize 96.1) &  \underline{.687} (\scriptsize 94.4) & .288 & .420 & .762 (\scriptsize 95.0) & .748 (\scriptsize 96.1) &  \underline{.792} (\scriptsize 96.1) & .702 (\scriptsize 94.4) & .697 &  \textbf{.795} \\
\texttt{wine} &  \underline{.296} (\scriptsize 72.2) & .230 (\scriptsize 70.1) &  \textbf{.525} (\scriptsize 70.6) & .154 (\scriptsize 70.2) & .134 & .224 & .097 (\scriptsize 72.2) & .087 (\scriptsize 71.4) & .099 (\scriptsize 70.6) & .108 (\scriptsize 70.2) &  \underline{.214} &  \textbf{.388} \\
\texttt{yeast} &  \underline{.568} (\scriptsize 78.5) & .471 (\scriptsize 77.8) &  \textbf{.713} (\scriptsize 76.4) & .362 (\scriptsize 78.5) & .467 & .418 & .412 (\scriptsize 78.5) & .367 (\scriptsize 77.8) & .274 (\scriptsize 75.4) & .319 (\scriptsize 78.5) &  \textbf{.427} & .358 \\
        \midrule
    \end{tabular}
}
\label{tab:drule_lgbm}
\end{table*}

\begin{table*}[h]
    \centering
    \caption{Robustness and faithfulness metrics for decision rules methods in all datasets with XGB as black-box. Prediction accuracy of surrogate classifiers is reported inside parentheses.}
    \footnotesize
\makebox[\linewidth]{
    \begin{tabular}{lcccccc|cccccc}
        \midrule
        \textbf{XGB}&\multicolumn{6}{c}{\textbf{Decision Rule Robustness}}&\multicolumn{6}{c}{\textbf{Decision Rule Faithfulness}}\\
        \cmidrule(lr){2-7}  \cmidrule(lr){8-13}
        & \textsc{ill}-\textsc{dt} & \textsc{ill}-\textsc{dt-us} & \textsc{lin-dt} & \textsc{inp-dt} & \textsc{lore} & \textsc{anchor} & \textsc{ill}-\textsc{dt} & \textsc{ill}-\textsc{dt-uc} & \textsc{lin-dt} & \textsc{inp-dt} & \textsc{lore} & \textsc{anchor} \\
        \midrule

\texttt{aids} &  \underline{.666} (\scriptsize{95.6}) & .378 (\scriptsize{93.0}) &  \textbf{.798} (\scriptsize{94.2}) & .098 (\scriptsize{95.1}) & .023 & .153 & .642 (\scriptsize{95.6}) & .518 (\scriptsize{93.7}) &  \underline{.657} (\scriptsize{94.2}) & .645 (\scriptsize{95.1}) & .639 &  \textbf{.821} \\
\texttt{austr} &  \underline{.513} (\scriptsize{92.0}) & .387 (\scriptsize{92.8}) &  \textbf{.713} (\scriptsize{92.8}) & .262 (\scriptsize{89.9}) & .123 & .256 &  \textbf{.780} (\scriptsize{92.8}) & .691 (\scriptsize{90.6}) & .587 (\scriptsize{92.8}) & .591 (\scriptsize{89.9}) & .697 &  \underline{.751} \\
\texttt{bank} &  \underline{.620} (\scriptsize{95.9}) & .528 (\scriptsize{95.5}) &  \textbf{.767} (\scriptsize{94.9}) & .549 (\scriptsize{95.1}) & .022 & .047 &  \underline{.728} (\scriptsize{95.9}) & .539 (\scriptsize{95.6}) & .551 (\scriptsize{94.9}) &  \textbf{.743} (\scriptsize{95.1}) & .360 & .465 \\
\texttt{breast} & .243 (\scriptsize{96.5}) & .184 (\scriptsize{96.5}) &  \underline{.465} (\scriptsize{97.4}) &  \textbf{.563} (\scriptsize{96.5}) & .068 & .032 &  \textbf{.813} (\scriptsize{96.5}) &  \underline{.772} (\scriptsize{97.4}) & .769 (\scriptsize{97.4}) & .795 (\scriptsize{96.5}) & .703 & .374 \\
\texttt{churn} & .291 (\scriptsize{96.1}) &  \underline{.625} (\scriptsize{94.5}) &  \textbf{.895} (\scriptsize{95.8}) & .355 (\scriptsize{99.0}) & .229 & .112 & .517 (\scriptsize{96.1}) & .365 (\scriptsize{93.4}) & .504 (\scriptsize{95.8}) &  \textbf{.571} (\scriptsize{99.0}) &  \underline{.533} & .394 \\
\texttt{compas} &  \underline{.811} (\scriptsize{94.0}) & .448 (\scriptsize{94.2}) &  \textbf{.965} (\scriptsize{94.5}) & .385 (\scriptsize{94.6}) & .101 & .154 &  \textbf{.606} (\scriptsize{94.2}) & .523 (\scriptsize{93.4}) & .576 (\scriptsize{94.5}) & .506 (\scriptsize{94.6}) & .489 &  \underline{.592} \\
\texttt{ctg} & .869 (\scriptsize{99.8}) & .852 (\scriptsize{100.}) &  \textbf{.976} (\scriptsize{99.5}) &  \underline{.882} (\scriptsize{99.5}) & .340 & .033 &  \underline{.787} (\scriptsize{100.}) & .727 (\scriptsize{99.3}) & .722 (\scriptsize{99.5}) &  \textbf{.795} (\scriptsize{99.5}) & .654 & .347 \\
\texttt{diabetes} & .124 (\scriptsize{91.6}) & .062 (\scriptsize{90.3}) & .124 (\scriptsize{90.9}) &  \underline{.176} (\scriptsize{92.2}) & .044 &  \textbf{.177} & .532 (\scriptsize{91.6}) &  \underline{.563} (\scriptsize{92.2}) & .503 (\scriptsize{90.9}) & .411 (\scriptsize{92.2}) & .293 &  \textbf{.688} \\
\texttt{ecoli} &  \underline{.680} (\scriptsize{100.}) & .533 (\scriptsize{100.}) &  \textbf{.800} (\scriptsize{100.}) & .588 (\scriptsize{100.}) & .581 & .497 &  \underline{.748} (\scriptsize{100.}) & .673 (\scriptsize{100.}) &  \textbf{.753} (\scriptsize{100.}) & .666 (\scriptsize{100.}) & .623 & .562 \\
\texttt{fico} & .340 (\scriptsize{93.8}) & .279 (\scriptsize{92.1}) &  \textbf{.413} (\scriptsize{92.8}) & .221 (\scriptsize{93.8}) & .053 & .187 & .617 (\scriptsize{93.8}) & .632 (\scriptsize{93.0}) & .614 (\scriptsize{92.8}) & .628 (\scriptsize{93.8}) &  \underline{.643} &  \textbf{.679} \\
\texttt{german} &  \underline{.337} (\scriptsize{87.0}) & .129 (\scriptsize{82.5}) &  \textbf{.565} (\scriptsize{81.5}) & .107 (\scriptsize{83.5}) & .033 & .081 &  \underline{.226} (\scriptsize{87.0}) & .169 (\scriptsize{81.0}) & .052 (\scriptsize{81.5}) & .041 (\scriptsize{83.5}) & .098 &  \textbf{.388} \\
\texttt{home} &  \underline{.095} (\scriptsize{96.0}) & .037 (\scriptsize{96.0}) & .082 (\scriptsize{97.0}) &  \textbf{.132} (\scriptsize{94.9}) & .010 & .062 &  \underline{.578} (\scriptsize{96.0}) & .414 (\scriptsize{97.0}) & .555 (\scriptsize{97.0}) & .488 (\scriptsize{94.9}) & .455 &  \textbf{.730} \\
\texttt{ionos} & .385 (\scriptsize{95.8}) & .374 (\scriptsize{94.4}) &  \underline{.514} (\scriptsize{97.2}) &  \textbf{.615} (\scriptsize{97.2}) & .212 & .232 & .701 (\scriptsize{95.8}) & .683 (\scriptsize{95.8}) &  \underline{.790} (\scriptsize{97.2}) &  \textbf{.829} (\scriptsize{97.2}) & .710 & .730 \\
\texttt{sonar} & .211 (\scriptsize{85.7}) & .221 (\scriptsize{85.7}) &  \textbf{.533} (\scriptsize{88.1}) &  \underline{.254} (\scriptsize{92.9}) & .141 & .072 &  \underline{.546} (\scriptsize{85.7}) &  \textbf{.597} (\scriptsize{85.7}) & .324 (\scriptsize{88.1}) & .535 (\scriptsize{92.9}) & .427 & .516 \\
\texttt{spam} & .419 (\scriptsize{96.1}) &  \underline{.483} (\scriptsize{97.1}) &  \textbf{.699} (\scriptsize{94.6}) & .403 (\scriptsize{94.7}) & .130 & .090 &  \textbf{.797} (\scriptsize{97.1}) & .643 (\scriptsize{93.1}) & .663 (\scriptsize{94.6}) &  \underline{.675} (\scriptsize{94.7}) & .227 & .478 \\
\texttt{titanic} & .620 (\scriptsize{95.5}) & .469 (\scriptsize{95.0}) &  \textbf{.955} (\scriptsize{97.2}) &  \underline{.693} (\scriptsize{96.1}) & .387 & .463 &  \textbf{.819} (\scriptsize{95.5}) & .690 (\scriptsize{97.2}) & .781 (\scriptsize{97.2}) & .737 (\scriptsize{96.1}) & .738 &  \underline{.807} \\
\texttt{wine} &  \underline{.303} (\scriptsize{71.0}) & .244 (\scriptsize{71.0}) &  \textbf{.458} (\scriptsize{70.6}) & .162 (\scriptsize{68.2}) & .150 & .229 & .160 (\scriptsize{71.0}) & .148 (\scriptsize{71.5}) & .164 (\scriptsize{70.6}) & .145 (\scriptsize{68.2}) &  \underline{.298} &  \textbf{.403} \\
\texttt{yeast} &  \underline{.533} (\scriptsize{89.2}) & .437 (\scriptsize{89.9}) &  \textbf{.788} (\scriptsize{88.2}) & .427 (\scriptsize{89.9}) & .469 & .387 &  \underline{.590} (\scriptsize{89.9}) & .469 (\scriptsize{87.5}) & .470 (\scriptsize{88.2}) & .538 (\scriptsize{89.9}) &  \textbf{.613} & .361 \\       
\midrule
\end{tabular}
}
\label{tab:drule_xgb}
\end{table*}

\begin{table*}[h]
    \centering
    \caption{Global robustness metrics in all datasets with LGB as black-box. Prediction accuracy of surrogate classifiers is reported inside parentheses.}
    \footnotesize
\makebox[\linewidth]{
    \begin{tabular}{lcccccc|cccccc}
        \toprule
        \textbf{LGB}&\multicolumn{6}{c}{\textbf{Feature Importance Robustness}}&\multicolumn{6}{c}{\textbf{Decision Rule Robustness}}\\
        \cmidrule(lr){2-7}  \cmidrule(lr){8-13}
        & \textsc{ill}-\textsc{lr} & \textsc{ill}-\textsc{lr-us} & \textsc{lin-lr} & \textsc{inp-lr} & \textsc{lime} & \textsc{shap} & \textsc{ill}-\textsc{dt} & \textsc{ill}-\textsc{dt-uc} & \textsc{lin-dt} & \textsc{inp-dt} & \textsc{lore} & \textsc{anchor} \\
        \midrule
\texttt{aids} &  \textbf{.743} (\scriptsize{95.3}) &  \underline{.653} (\scriptsize{97.0}) & .563 (\scriptsize{93.7}) & .352 (\scriptsize{93.7}) & .163 & .254 &  \underline{.267} (\scriptsize{97.4}) & .252 (\scriptsize{96.7}) &  \textbf{.595} (\scriptsize{97.9}) & .192 (\scriptsize{98.8}) & .170 & .191 \\
\texttt{australian} &  \textbf{.801} (\scriptsize{93.5}) &  \underline{.651} (\scriptsize{89.9}) & .488 (\scriptsize{92.0}) & .396 (\scriptsize{92.0}) & .041 & .181 & .297 (\scriptsize{91.3}) &  \textbf{.383} (\scriptsize{92.0}) &  \underline{.361} (\scriptsize{97.1}) & .274 (\scriptsize{94.9}) & .197 & .141 \\
\texttt{bank} &  \textbf{.897} (\scriptsize{96.1}) &  \underline{.839} (\scriptsize{96.0}) & .798 (\scriptsize{94.9}) & .671 (\scriptsize{95.0}) & .147 & .298 &  \textbf{.540} (\scriptsize{95.5}) & .387 (\scriptsize{95.5}) &  \underline{.425} (\scriptsize{95.3}) & .192 (\scriptsize{96.1}) & .055 & .066 \\
\texttt{breast} & .532 (\scriptsize{97.4}) &  \textbf{.638} (\scriptsize{96.5}) &  \underline{.568} (\scriptsize{99.1}) & .553 (\scriptsize{98.2}) & .280 & .226 &  \textbf{.285} (\scriptsize{98.2}) &  \underline{.266} (\scriptsize{96.5}) & .227 (\scriptsize{96.5}) & .123 (\scriptsize{96.5}) & .024 & .122 \\
\texttt{churn} &  \textbf{.664} (\scriptsize{95.8}) &  \underline{.655} (\scriptsize{94.8}) & .614 (\scriptsize{87.0}) & .476 (\scriptsize{87.6}) & .000 & .299 &  \textbf{.350} (\scriptsize{95.7}) &  \underline{.256} (\scriptsize{95.2}) & .210 (\scriptsize{97.3}) & .167 (\scriptsize{98.2}) & .138 & .100 \\
\texttt{compas} &  \textbf{.875} (\scriptsize{94.7}) &  \underline{.685} (\scriptsize{95.1}) & .527 (\scriptsize{94.4}) & .421 (\scriptsize{94.4}) & .142 & .352 & .432 (\scriptsize{95.1}) &  \textbf{.547} (\scriptsize{94.3}) &  \underline{.493} (\scriptsize{95.1}) & .205 (\scriptsize{95.8}) & .193 & .152 \\
\texttt{ctg} &  \textbf{.758} (\scriptsize{99.8}) & .729 (\scriptsize{99.8}) &  \underline{.732} (\scriptsize{99.5}) & .647 (\scriptsize{99.3}) & .117 & .405 & .293 (\scriptsize{99.3}) &  \underline{.445} (\scriptsize{99.3}) &  \textbf{.511} (\scriptsize{99.3}) & .062 (\scriptsize{99.3}) & .011 & .275 \\
\texttt{diabetes} &  \textbf{.700} (\scriptsize{92.9}) &  \underline{.684} (\scriptsize{94.2}) & .499 (\scriptsize{89.6}) & .498 (\scriptsize{89.6}) & .486 & .335 &  \underline{.305} (\scriptsize{89.6}) & .304 (\scriptsize{87.7}) & .255 (\scriptsize{89.6}) &  \textbf{.344} (\scriptsize{88.3}) & .198 & .231 \\
\texttt{ecoli} & .825 (\scriptsize{100.}) &  \textbf{.871} (\scriptsize{100.}) & .834 (\scriptsize{100.}) &  \underline{.838} (\scriptsize{100.}) & .447 & .749 & .524 (\scriptsize{100.}) &  \textbf{.653} (\scriptsize{100.}) & .451 (\scriptsize{100.}) & .277 (\scriptsize{100.}) & .456 &  \underline{.633} \\
\texttt{fico} &  \textbf{.744} (\scriptsize{90.3}) &  \underline{.678} (\scriptsize{90.8}) & .501 (\scriptsize{90.3}) & .426 (\scriptsize{90.3}) & .415 & .411 &  \textbf{.318} (\scriptsize{90.1}) &  \underline{.300} (\scriptsize{89.5}) & .274 (\scriptsize{89.5}) & .224 (\scriptsize{90.5}) & .131 & .247 \\
\texttt{german} &  \textbf{.736} (\scriptsize{89.0}) &  \underline{.653} (\scriptsize{85.5}) & .559 (\scriptsize{85.5}) & .523 (\scriptsize{88.5}) & .072 & .242 & .243 (\scriptsize{86.0}) &  \underline{.252} (\scriptsize{85.5}) &  \textbf{.354} (\scriptsize{83.0}) & .214 (\scriptsize{81.0}) & .193 & .251 \\
\texttt{home} &  \textbf{.720} (\scriptsize{96.0}) &  \underline{.612} (\scriptsize{97.0}) & .606 (\scriptsize{94.9}) & .590 (\scriptsize{94.9}) & .437 & .311 & .478 (\scriptsize{96.0}) & .341 (\scriptsize{96.0}) & .430 (\scriptsize{98.0}) &  \textbf{.510} (\scriptsize{94.9}) & .434 &  \underline{.449} \\
\texttt{ionosphere} &  \textbf{.741} (\scriptsize{94.4}) &  \underline{.740} (\scriptsize{94.4}) & .723 (\scriptsize{88.7}) & .519 (\scriptsize{88.7}) & .259 & .586 &  \underline{.630} (\scriptsize{98.6}) &  \textbf{.663} (\scriptsize{90.1}) & .581 (\scriptsize{95.8}) & .144 (\scriptsize{95.8}) & .054 & .212 \\
\texttt{sonar} &  \textbf{.793} (\scriptsize{85.7}) &  \underline{.735} (\scriptsize{92.9}) & .725 (\scriptsize{83.3}) & .715 (\scriptsize{83.3}) & .288 & .388 &  \textbf{.581} (\scriptsize{83.3}) &  \underline{.382} (\scriptsize{85.7}) & .379 (\scriptsize{81.0}) & .228 (\scriptsize{83.3}) & .129 & .102 \\
\texttt{spam} &  \underline{.386} (\scriptsize{96.0}) & .178 (\scriptsize{96.2}) &  \textbf{.416} (\scriptsize{94.0}) & .357 (\scriptsize{94.4}) & .274 & .069 & .201 (\scriptsize{94.7}) &  \underline{.273} (\scriptsize{95.1}) & .152 (\scriptsize{94.5}) & .097 (\scriptsize{93.7}) & .147 &  \textbf{.289} \\
\texttt{titanic} &  \textbf{.864} (\scriptsize{90.5}) & .770 (\scriptsize{90.5}) &  \underline{.777} (\scriptsize{89.9}) & .689 (\scriptsize{89.4}) & .551 & .681 & .612 (\scriptsize{93.9}) &  \underline{.616} (\scriptsize{95.0}) &  \textbf{.658} (\scriptsize{96.1}) & .569 (\scriptsize{94.4}) & .470 & .548 \\
\texttt{wine} &  \textbf{.714} (\scriptsize{64.2}) &  \underline{.602} (\scriptsize{67.2}) & .577 (\scriptsize{59.1}) & .554 (\scriptsize{58.6}) & .243 & .484 & .269 (\scriptsize{72.2}) &  \underline{.271} (\scriptsize{70.1}) & .225 (\scriptsize{70.6}) & .253 (\scriptsize{70.2}) & .270 &  \textbf{.359} \\
\texttt{yeast} &  \textbf{.617} (\scriptsize{76.4}) &  \underline{.474} (\scriptsize{75.1}) & .436 (\scriptsize{69.4}) & .440 (\scriptsize{69.4}) & .096 & .299 & .331 (\scriptsize{78.5}) & .355 (\scriptsize{77.8}) &  \underline{.380} (\scriptsize{75.4}) & .376 (\scriptsize{78.5}) & .248 &  \textbf{.411} \\
        \bottomrule
    \end{tabular}
}
\label{tab:global_lgbm}
\end{table*}

\begin{table*}[h]
    \centering
    \caption{Global robustness metrics in all datasets with XGB as black-box. Prediction accuracy of surrogate classifiers is reported inside parentheses.}
    \footnotesize
\makebox[\linewidth]{
    \begin{tabular}{lcccccc|cccccc}
        \toprule
        \textbf{XGB}&\multicolumn{6}{c}{\textbf{Feature Importance Robustness}}&\multicolumn{6}{c}{\textbf{Decision Rule Robustness}}\\
        \cmidrule(lr){2-7}  \cmidrule(lr){8-13}
        & \textsc{ill}-\textsc{lr} & \textsc{ill}-\textsc{lr-us} & \textsc{lin-lr} & \textsc{inp-lr} & \textsc{lime} & \textsc{shap} & \textsc{ill}-\textsc{dt} & \textsc{ill}-\textsc{dt-uc} & \textsc{lin-dt} & \textsc{inp-dt} & \textsc{lore} & \textsc{anchor} \\
        \midrule
  \texttt{aids} &  \textbf{.720} \scriptsize{(95.3)} &  \underline{.492} \scriptsize{(95.1)} & .485 \scriptsize{(91.1)} & .400 \scriptsize{(91.1)} & .159 & .250 &  \underline{.298} \scriptsize{(95.6)} &  \textbf{.400} \scriptsize{(92.3)} & .249 \scriptsize{(94.2)} & .196 \scriptsize{(95.1)} & .194 & .186 \\
\texttt{australian} &  \textbf{.781} \scriptsize{(92.0)} & .482 \scriptsize{(92.8)} & .554 \scriptsize{(90.6)} &  \underline{.569} \scriptsize{(90.6)} & .022 & .269 &  \textbf{.378} \scriptsize{(92.0)} &  \underline{.377} \scriptsize{(92.8)} & .291 \scriptsize{(92.8)} & .193 \scriptsize{(89.9)} & .262 & .262 \\
\texttt{bank} &  \textbf{.872} \scriptsize{(96.2)} &  \underline{.810} \scriptsize{(96.0)} & .758 \scriptsize{(95.3)} & .668 \scriptsize{(95.5)} & .105 & .293 &  \textbf{.530} \scriptsize{(95.9)} &  \underline{.496} \scriptsize{(95.5)} & .436 \scriptsize{(94.9)} & .145 \scriptsize{(95.1)} & .012 & .055 \\
\texttt{breast} &  \underline{.698} \scriptsize{(99.1)} &  \textbf{.707} \scriptsize{(99.1)} & .506 \scriptsize{(100.0)} & .272 \scriptsize{(97.4)} & .262 & .251 &  \underline{.305} \scriptsize{(96.5)} &  \textbf{.356} \scriptsize{(95.6)} & .295 \scriptsize{(97.4)} & .183 \scriptsize{(96.5)} & .127 & .108 \\
\texttt{churn} & .585 \scriptsize{(95.1)} &  \textbf{.648} \scriptsize{(95.1)} &  \underline{.604} \scriptsize{(87.6)} & .473 \scriptsize{(87.9)} & .017 & .339 &  \underline{.324} \scriptsize{(96.1)} &  \textbf{.326} \scriptsize{(94.5)} & .315 \scriptsize{(95.8)} & .192 \scriptsize{(99.0)} & .132 & .175 \\
\texttt{compas} &  \textbf{.844} \scriptsize{(95.4)} & .713 \scriptsize{(95.2)} & .589 \scriptsize{(94.5)} & .407 \scriptsize{(94.5)} & .127 & .354 & .327 \scriptsize{(94.0)} &  \underline{.546} \scriptsize{(94.2)} &  \textbf{.552} \scriptsize{(94.5)} & .179 \scriptsize{(94.6)} & .162 & .137 \\
\texttt{ctg} &  \textbf{.811} \scriptsize{(100.)} &  \underline{.776} \scriptsize{(99.8)} & .760 \scriptsize{(99.8)} & .552 \scriptsize{(99.8)} & .000 & .545 &  \underline{.402} \scriptsize{(99.8)} & .390 \scriptsize{(100.)} &  \textbf{.405} \scriptsize{(99.5)} & .081 \scriptsize{(99.5)} & .000 & .271 \\
\texttt{diabetes} &  \textbf{.661} \scriptsize{(92.2)} &  \underline{.620} \scriptsize{(92.9)} & .487 \scriptsize{(90.3)} & .486 \scriptsize{(87.0)} & .600 & .380 & .278 \scriptsize{(91.6)} & .271 \scriptsize{(90.3)} &  \textbf{.307} \scriptsize{(90.9)} &  \underline{.281} \scriptsize{(92.2)} &  \textbf{.307} & .239 \\
\texttt{ecoli} &  \textbf{.892} \scriptsize{(100.)} &  \underline{.851} \scriptsize{(100.)} & .811 \scriptsize{(100.)} & .816 \scriptsize{(98.5)} & .381 & .662 & .466 \scriptsize{(100.)} &  \textbf{.723} \scriptsize{(100.)} &  \underline{.551} \scriptsize{(100.)} & .288 \scriptsize{(100.)} & .527 & .587 \\
\texttt{fico} &  \textbf{.718} \scriptsize{(93.4)} &  \underline{.674} \scriptsize{(93.2)} & .519 \scriptsize{(92.9)} & .342 \scriptsize{(92.9)} & .432 & .371 &  \textbf{.346} \scriptsize{(93.8)} &  \underline{.326} \scriptsize{(92.1)} & .234 \scriptsize{(92.8)} & .197 \scriptsize{(93.8)} & .108 & .236 \\
\texttt{german} &  \textbf{.714} \scriptsize{(87.0)} & .629 \scriptsize{(87.0)} &  \underline{.640} \scriptsize{(86.0)} & .541 \scriptsize{(86.0)} & .054 & .315 &  \underline{.270} \scriptsize{(87.0)} & .209 \scriptsize{(77.5)} &  \textbf{.403} \scriptsize{(81.5)} & .195 \scriptsize{(83.5)} & .160 & .261 \\
\texttt{home} &  \textbf{.729} \scriptsize{(93.9)} &  \underline{.671} \scriptsize{(96.0)} & .618 \scriptsize{(93.9)} & .613 \scriptsize{(93.9)} & .611 & .355 &  \textbf{.506} \scriptsize{(96.0)} & .408 \scriptsize{(96.0)} & .452 \scriptsize{(97.0)} &  \underline{.473} \scriptsize{(94.9)} & .407 & .457 \\
\texttt{ionosphere} & .678 \scriptsize{(91.5)} &  \textbf{.741} \scriptsize{(91.5)} &  \underline{.733} \scriptsize{(87.3)} & .479 \scriptsize{(90.1)} & .235 & .216 &  \textbf{.718} \scriptsize{(95.8)} &  \underline{.647} \scriptsize{(94.4)} & .456 \scriptsize{(97.2)} & .090 \scriptsize{(97.2)} & .086 & .120 \\
\texttt{sonar} &  \underline{.723} \scriptsize{(81.0)} &  \textbf{.756} \scriptsize{(81.0)} & .671 \scriptsize{(76.2)} & .673 \scriptsize{(78.6)} & .062 & .325 &  \underline{.391} \scriptsize{(85.7)} &  \textbf{.496} \scriptsize{(85.7)} & .259 \scriptsize{(88.1)} & .283 \scriptsize{(92.9)} & .138 & .027 \\
\texttt{spam} &  \underline{.577} \scriptsize{(96.5)} &  \textbf{.579} \scriptsize{(96.9)} & .425 \scriptsize{(94.5)} & .279 \scriptsize{(94.6)} & .351 & .069 &  \underline{.234} \scriptsize{(95.9)} &  \textbf{.238} \scriptsize{(97.1)} & .173 \scriptsize{(94.6)} & .046 \scriptsize{(94.7)} & .132 & .232 \\
\texttt{titanic} &  \textbf{.830} \scriptsize{(91.1)} & .704 \scriptsize{(91.1)} & .682 \scriptsize{(88.8)} &  \underline{.710} \scriptsize{(89.9)} & .433 & .588 & .621 \scriptsize{(95.5)} &  \underline{.642} \scriptsize{(95.0)} &  \textbf{.661} \scriptsize{(97.2)} & .594 \scriptsize{(96.1)} & .550 & .541 \\
\texttt{wine} &  \textbf{.667} \scriptsize{(66.3)} &  \underline{.629} \scriptsize{(66.6)} & .616 \scriptsize{(60.4)} & .563 \scriptsize{(59.3)} & .338 & .485 &  \underline{.286} \scriptsize{(71.0)} & .285 \scriptsize{(71.0)} & .253 \scriptsize{(70.6)} & .262 \scriptsize{(68.2)} & .273 &  \textbf{.347} \\
\texttt{yeast} &  \textbf{.644} \scriptsize{(86.5)} &  \underline{.637} \scriptsize{(88.2)} & .503 \scriptsize{(81.5)} & .472 \scriptsize{(81.8)} & .314 & .412 & .385 \scriptsize{(89.2)} & .425 \scriptsize{(89.9)} &  \underline{.434} \scriptsize{(88.2)} &  \textbf{.448} \scriptsize{(89.9)} & .329 & .436 \\
        \bottomrule
    \end{tabular}
}
\label{tab:global_xgb}
\end{table*}

\end{document}